\documentclass{cta-author}
\usepackage{amsmath,amssymb}
\usepackage{url}
\usepackage{diagbox}
\usepackage{xcolor}

{}
{}
{}

\def\x{{\mathbf{x}}}

\begin{document}

\supertitle{Review of wavelet-based unsupervised texture segmentation, advantage of adaptive wavelets}

\title{Review of wavelet-based unsupervised texture segmentation, advantage of adaptive wavelets}

\author{\au{Yuan Huang$^{1}$}, \au{Valentin De~Bortoli$^{2}$}, \au{Fugen Zhou$^1$}, \au{J\'{e}r\^{o}me Gilles$^{3}$}}

\address{\add{1}{Image Processing Center, Beihang University, Beijing 100191, China}
\add{2}{Centre pour les Math\'ematiques et Leurs Applications, Ecole Normale Sup\'erieure of Paris Saclay, 61 Av du Pr\'esident Wilson, 94230 Cachan, France}
\add{3}{Department of Mathematics and Statistics, San Diego State University, 5500 Campanile Dr, San Diego, CA, 92182, USA}
\email{jgilles@mail.sdsu.edu}}

\begin{abstract}
Wavelet-based segmentation approaches are widely used for texture segmentation purposes because of their ability to characterize different textures. In this paper, 
we assess the influence of the chosen wavelet and propose to use the recently introduced empirical wavelets. We show that the adaptability of the empirical wavelet 
permits to reach better results than classic wavelets. In order to focus only on the textural information, we also propose to perform a cartoon + texture decomposition step before applying the segmentation algorithm. The proposed method is tested on six classic benchmarks, based on several popular texture images.
\end{abstract}

\maketitle

\section{Introduction}\label{sec:introduction}
Image segmentation plays an important role in a wide variety of tasks including medical imaging, remote sensing, industrial automation and security. 
Although image segmentation has been studied for about thirty years by researchers in computer vision and mathematics, this problem is still considered as an open problem. In this 
paper, we focus on the case of texture segmentation, which aims to partition an image into regions corresponding to distinct textures. Texture segmentation is a 
particular case of image segmentation in the sense that the used descriptors are very specific. Textures are important for low-level image analysis and understanding, especially 
for applications like medical imaging (tumor detection), target detection (camouflaged targets), microscopy imaging (molecule alignments). Various supervised and 
unsupervised approaches try to address the long-standing problem of texture segmentation. Each algorithm specificity lies in the choice of the textural information descriptors.

\begin{figure}[!t]
\includegraphics[width=\columnwidth]{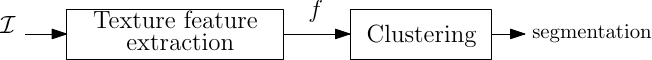}
\caption{General steps in unsupervised texture segmentation.}
\label{fig:scheme1}
\end{figure}

Among all segmentation methods (and in particular for texture segmentation), the unsupervised framework is the most difficult one since it does not consider learning 
in advance a dictionary of textures involved in the problem. Hence, the only information provided to the algorithm is the original image and a potential set of parameters. An unsupervised image segmentation algorithm thus relies on the simple idea that the image itself contains the necessary information (either in the image domain or in 
a transform domain) to perform the expected segmentation. In general, an unsupervised texture segmentation method can be split into two main steps (as depicted in 
Figure~\ref{fig:scheme1}): the extraction of texture feature vectors followed by a clustering step.\\
Although human vision is pretty efficient in distinguishing different textures, no clear formal mathematical definition of texture exists, which makes 
texture segmentation a particularly challenging problem. In the recent literature, the most common strategy to segment textures is to find an image model which supposedly 
to distinguish different textures and incorporate this model in a segmentation framework. Such a model is usually based on the extraction of texture features which represent 
characteristics of textures at different scales. Finally, some type of ``energy'' is computed and provided to a clustering or any other segmentation method. Several feature 
extraction algorithms based on different approaches were published in the literature.  Haralick and others \cite{cooccurence,Rampun} used statistical measurements to 
define 
co-occurence matrices in order to characterize textures. Probabilistic models, especially based on Markov Random Fields (MRF), were also proposed in \cite{tdefinition,MRFseg}, 
taking advantage of the Hammersley-Clifford theorem, the authors of \cite{HCTeorem} relate the local MRF information to the global distribution. Such approach allows 
for the definition of a global model through the local features which corresponds to characterize textures at different scales. Local regularity measured by either 
fractal dimension or local histogram on fuzzy regions (within a hierarchical framework to take into account different scales) was used to extract geometrical features in 
\cite{geofeature,Pustelnik,Zheng}. Because it corresponds to the human visual system, one of the most popular approach to extract texture features is based on Gabor 
wavelet filters \cite{Gabor,paragios,Sagiv04,Hammouda,Dunn,Nava}. Gabor filters had a lot of success for texture analysis because of their selectivity in both scales and 
orientations. However, the construction of a Gabor filter bank requires the choice of several parameters like spatial scales, carrier frequencies and orientations. Inspired by the 
scale selectivity of Gabor filters, the use of wavelets was proposed to extract texture features \cite{Wavelettexturesegmentation,wavelet1,wavelet2,Bashar,Hsin,Cimpoi,Storath,Gabortoolbox}. The opportunity to choose various basis functions for the wavelet transform permits some 
flexibility regarding the type of textures that can be analyzed.

Recently, adaptive decompositions like the Empirical Mode Decomposition (EMD) \cite{EMD} have received a lot of attention in the literature. Their purpose is to decompose a signal 
or an image in an ``optimized'' way, i.e extracting compact harmonic modes, thus providing more insight about the analyzed signal. Despite many great successes in applications in 
different fields, the EMD method is a purely algorithmic method and lacks mathematical foundations which restricts us from fully understanding and predicting its nonlinear 
behavior. To circumvent this issue, in \cite{EWT1D}, Gilles proposed to build an adaptive wavelet transform, called Empirical Wavelet Transform (EWT). It is based on the following 
idea: instead of using a set of prescribed scales like in the classic wavelet transform, the EWT automatically finds the Fourier supports of each wavelet filter by analyzing the 
magnitude of the Fourier spectrum of the signal under consideration. This approach was extended to 2D in \cite{EWT2D} where the classic tensor wavelet, ridgelet and curvelet 
transforms were revisited (as well as a new Littlewood-Paley wavelet transform) by adding the detection of the wavelet filters supports in the 2D Fourier space.

Let us introduce some general notations that will be used throughout the paper. Given an original gray scale image, $\mathcal{I}$ (normalized between $0$ and $1$) of size $\mathcal{N} \times \mathcal{N}$, in the following, we denote $\mathcal{I}(\x)$ the intensity of $\mathcal{I}$ at a pixel $\x\in \Omega=[1,\ldots,\mathcal{N}] \times [1,\ldots,\mathcal{N}]$. Extracting a texture feature vector consists in finding a vector $f(\x) \in \mathbb{R}^K$, for each pixel $\x\in \Omega$. The dimension $K$ represents the number of extracted features and generally depends on the chosen method.

In this paper, our goal is to review wavelet-based features for unsupervised segmentation. We focus on comparing several types of wavelets and we explore the opportunity to use 
the 
recently introduced 2D empirical wavelets. Since our main focus is the comparison of wavelets, we will use standard clustering algorithms for the last step and do not intend 
to develop ``optimized'' clustering approaches. The rest of the paper is organized as follows: Sections \ref{featproc} and \ref{clustalgo}
are remainders of general facts about the feature extraction step and standard clustering methods, respectively; in Section~\ref{sec:wavfeat}, we review existing wavelet methods, as well as the new 
empirical wavelets method, for texture representation; in Section~\ref{algo}, we describe the unsupervised texture segmentation strategy we used and suggest the introduction of a cartoon 
and texture decomposition intermediate step to focus on the textural information; Section~\ref{experiments} shows several experiments based on some well-known texture datasets and 
provides some benchmarks for easier comparisons; finally, we conclude and give some perspectives in Section~\ref{conclusion}.

%========================================================================================================================================
\section{Feature extraction}\label{featproc}
The first step in an unsupervised texture segmentation procedure is to extract a set of feature vectors $f(\x)\in\mathbb{R}^K$ which characterize the texture properties. It is 
expected that the texture information will be clustered making the classification easier to achieve. The construction of such feature vectors 
has generally two steps: the extraction of raw features and then some post-processing in order to homogenize the information.\\
In this paper, we focus on wavelet-based texture features. Wavelet-based features are very popular in the literature because they are easily obtained by linear filterings making 
them good candidates for real-time applications. In this particular case, the raw features correspond to the wavelet coefficients, denoted 
$\tilde{f}(\x):\Omega\to\mathbb{R}^K$, obtained by the corresponding filter bank where $K$ is the number of filters.\\
Wavelet coefficients are generally not usable by a classifier in their original form and need to be processed first. Let us denote $f(\x)\in\mathbb{R}^K$, $\x\in 
\Omega$, the final feature vector at pixel $\x$; and $f_k(\x)$ its $k-$th coordinate ($k=1,2,\ldots,K$). Let us denote $\mathcal{E}$ the post-processing operator, i.e 
$f(\x)=\mathcal{E}[\tilde{f}](\x)$. In the field of wavelet-based feature vectors, the processing mainly used is the computation of either local energy or local entropy 
\cite{Entropyenergy,Entropyenergy2}. Here, $\mathcal{E}[\tilde{f}]_k$ is defined by $\mathcal{E}[\tilde{f}]_k(\x)=W(E[\tilde{f}(\x)]_k)$ where $W$ denotes the local 
mean filter, i.e the mean filter processed over a window of a given size and centered at $\x$ (the influence of the window's size will be investigated in the experimental section). The 
operator $E$ is defined by $\forall \mathbf{u} = (u_1,...,u_K) \in \mathbb{R}^K$, $E(\mathbf{u})=(u_1^2,...,u_K^2)$ in the local energy case or 
$E(\mathbf{u})=(-u_1\log(u_1),...,-u_K \log(u_K))$ in the local entropy case. Note that in the second case we scale the feature vector between 0 and 1 so the entropy 
is well-defined. The size of the used window will be discussed in Section~\ref{winsize}. It is worth mentioning that in \cite{Entropyenergy3} the authors state that there is no 
clear 
and definitive answer on how to choose between the energy and the entropy.\\
Another type of processing that had some success in texture analysis is the Local Binary Pattern (LBP) transform, we refer the reader to \cite{LBPtoolbox,LBP} for full details. In this 
paper, we apply the LPB transform to the wavelet coefficients at each scale, i.e $\mathcal{E}[\tilde{f}]_k(\x)=LPB[\tilde{f}_k](\x)$.\\
%========================================================================================================================================
\section{Clustering algorithm}\label{clustalgo}
Once the final feature vectors $f(\x)$, $\x\in\Omega$, are obtained, the last step consists in performing a pixel-wise clustering in order to obtain the final segmentation. Many 
algorithms exist to address this problem. Most of them require a large number of parameters in order to run a correct clustering. One must take care of the \textit{curse of 
high-dimensionality} which states that the larger the number of features is, the more complicated it is to perform a relevant clustering \cite{cursehighdim}. In this study we will 
focus on two standard clustering algorithms, the \textit{k-means} algorithm \cite{kmeans} and the Nystr\"{o}m method which is based on spectral-clustering 
\cite{nystrom,spectralclustering}. The choice of \textit{k-means} was made because it is a classic algorithm and is widely used in the literature. For the reader convenience, we 
recall that it is a centroid-based algorithm which partition the data into Voronoi cells. The Nystr\"{o}m method is a recent clustering algorithm relying on the computation of 
eigenvectors of the Graph Laplacian of an affinity matrix see \cite{nystrom,spectralclustering} for details. In this paper, we aim at comparing the results obtained from these two 
algorithms in order to assess the influence of the clustering step. Both of these algorithms require the number of expected classes as an input parameter and we will provide this 
information to the algorithms. The standard procedure is used to initialize these algorithms: the first cluster center is chosen uniformly at random in the feature 
space, after that each subsequent cluster center is chosen randomly from the remaining feature vectors with probability proportional to its distance from the point's closest 
existing cluster center (see \cite{kmeansseeding} for details). For fair comparisons, the same initialization is used for both the \textit{k-means} and Nystr\"{o}m algorithms, as 
well as for all wavelets families.

%========================================================================================================================================
\section{Wavelet features}\label{sec:wavfeat}
This section reviews different existing wavelet methods which will be used to extract the texture feature vectors. We refer the reader to the two classic texts from Mallat \cite{Mallat} and Daubechies \cite{Daubechies} on wavelet analysis and its mathematical foundations. As in the previous section, we denote $\Omega$ the image domain, 
$\x=(i,j)\in\Omega$ a pixel location and we assume that all images $\mathcal{I}\in\mathrm{L}^2(\Omega)$. 
\subsection{Standard wavelets}
A wavelet decomposition of $\mathcal{I}$ consists in projecting $\mathcal{I}$ onto a family of functions (wavelets) $\{\Psi_{s,\x},\Phi_{L,\x}\}_{(s,\x)\in  \{ s \in \mathbb{Z}, \ s<L \} \times \Omega}$. In this paper, we will use the following convention: $\Psi_{L,\x}=\Phi_{L,\x}$ in order to simplify the notations (the wavelet family can now be written $\{\Psi_{s,\x}\}_{\{s\in\mathbb{Z},s\leq L\}\times \Omega}$). 
In practice, using a finite number of scales, the wavelet decomposition of $\mathcal{I}$ is given by standard inner products $\tilde{f}_k(\x)=\langle \mathcal{I},\Psi_{S_k,\x} \rangle,\x\in \Omega,k=1,2,\ldots,K$. It is well known \cite{Mallat, Daubechies} that such inner products are equivalent to perform a linear filtering with filters corresponding to each scale $S_k$. The most widely used scaling factor in the literature is the dyadic case, i.e. $S_k=2^k$ (with our previous convention, $L=2^K$). Each filters $\Psi_{S_k}$ provides the image details lying at scale $S_k$, for $k=1,2,\ldots,K-1$; while the filter $\Psi_{S_K}=\Phi_{S_K}$ provides a coarse approximation of the image at the scale $S_K$. 
\begin{figure}[!t]
\includegraphics[width=\columnwidth]{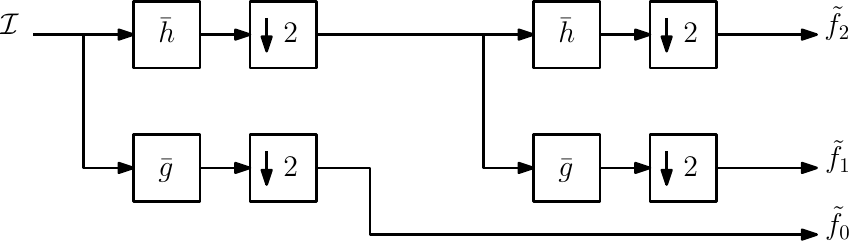}
\caption{Dyadic decimated wavelet transform for $k=2$.}
\label{fig:dydec}
\end{figure}
The easiest approach to numerically compute such 2D wavelet decomposition is to build tensor wavelets, i.e. to perform 1D filtering along the rows and columns, respectively. 
Moreover, two options exist depending on the expected type of wavelet families: with or without decimation. Hereafter we assume a dyadic decomposition. Let us denote $h$ and $g$ 
the digital filters corresponding respectively to $\Phi_{S_1}$ and $\Psi_{S_1}$. In the decimated wavelet transform, wavelet coefficients at a given scale are obtained by applying 
the same filters to a downsampled version of the output of $h$ from the previous scale (see Figure~\ref{fig:dydec}). This process is then repeated $K-1$ times to obtain the $K$ 
components of $\tilde{f}$.  
\begin{figure}[!t]
\centering\includegraphics[width=0.62\columnwidth]{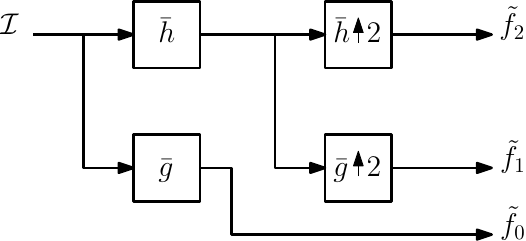}
\caption{Dyadic undecimated wavelet transform for $K=2$.}
\label{fig:dyundec}
\end{figure}
The undecimated wavelet transform follows the same process except that there is no image downsampling between each scale. The change of scale is obtained by dilating the filters 
$h$ and $g$ inserting zeros between each filter samples (this technique is also known as the \textit{A Trou} algorithm \cite{atrou}). These two approaches have radically 
different properties. The images obtained at each scale for the decimated transform have their size divided by two compared to the previous scale, hence in order to have all 
$\tilde{f}$ of the same size, it is necessary to perform an interpolation. The most common method, and the one used in our study, is a nearest-neighbour interpolation. Although 
this 
decimation is useful to build orthogonal wavelet families, it has a major drawback: the obtained transform is not translation invariant. This can be a serious issue, especially for 
recognition applications like texture analysis. The undecimated transform allows us to ignore this issue, however it does not allow to build orthogonal bases but wavelet frames. We 
recall that a family of functions $\{f_n\}$ defines a frame of $L^2(\Omega)$ if $\exists \ 0<A<B<\infty, \forall v \in L^2(\Omega)$, $$A\Vert v\Vert_{L^2(\Omega)}^2 
\le \sum_{n}\vert \langle f_n,v \rangle \vert^2 \le B\Vert v\Vert_{L^2(\Omega)}^2.$$ The frame is said to be tight if $A=B$ and adjusted if $A=B=1$ \cite{Christensen2008}. The 
choice of an undecimated transform over a decimated one when performing texture analysis was already suggested in \cite{Wavaujol}.
%========================================================================================================================================
\subsection{Packet extension}
The decimated transform presented in the previous section may not be the ``optimal'' representation, in the sense of information entropy, of the input image. A better 
representation can be obtained by also iterating the filtering process (with the same filters $h$ and $g$) on each subband and not only the coarsest one, in order to obtain a 
better adapted 
basis. This transform is called the wavelet packet transform, see \cite{Mallat,Daubechies}. The best basis algorithm \cite{Bestbasis} builds such optimal basis given a cost 
function based on the entropy. At a given scale, for each component, the algorithm checks whether decomposing again this component provides a representation with lower entropy or 
not. The final decomposition corresponds to the one with lower entropy at all scales. It is worth mentioning that the packet transform combined with the best 
basis algorithm is the first data-driven wavelet transform.
%========================================================================================================================================
\subsection{Gabor wavelets}
Other very popular wavelets, especially for texture analysis, are the 2D Gabor wavelets \cite{Mallat,Gabor,Gaborref1,Gaborref2}. A family of Gabor wavelets corresponds to an 
optimal filter bank in the sense that it optimizes the Gabor-Heisenberg uncertainty principle. The uncertainty principle states that a filter cannot measure accurate information 
simultaneously in space and Fourier domains. Let us denote $g$ a two-dimensional gaussian kernel, $(\alpha_1,...\alpha_n)$ a vector of angles 
and $\mathbf{\omega} = (\omega_1,\omega_2)$ a 2D frequency vector. Gabor wavelets can be easily defined in the Fourier domain by 
\begin{equation*}
\hat{\Psi}_{s,k}(\mathbf{\omega})=\sqrt{2^s}\hat{g}(2^s(\omega_1- \cos \alpha_k ),2^s(\omega_2- \sin \alpha_k))
\end{equation*}
Such a wavelet family does not correspond to a basis but to a frame. Besides the fact that physiological evidences show that the human visual system performs a 
Gabor wavelets 
decomposition, they are widely used in the texture analysis literature \cite{Gabor,Gaborref1,Gaborref2,paragios,Sagiv04,Hammouda,Dunn,Nava} because they can extract 
information at different orientations.
%========================================================================================================================================
\subsection{Empirical wavelets}
All wavelet families described in the previous sections correspond to build filters with supports in the Fourier domain which follow a prescribed scheme. More precisely, the positions and sizes of these Fourier supports depend on the choice of the scaling scheme, and therefore correspond to a prescribed partitioning of the Fourier domain. For instance, in 1D, the dyadic case is equivalent to divide the lowpass filter support by a factor two at each scale. The drawback of such fixed partitioning approach is that nothing guarantees that a harmonic mode will fall within a prescribed support, resulting in the separation of frequencies which are contributing to the same information. To circumvent this issue, in \cite{EWT1D}, the author proposes to build data-driven wavelets in order to capture harmonic modes with compact supports in the Fourier domain. Such an approach can be easily extended to 2D in different ways (tensor, Littlewood-Paley wavelets, curvelets) for image processing purposes. In this paper, we propose to use empirical wavelets features to better capture texture characteristics. In the following sections, we briefly sum up how empirical wavelets are constructed.
%========================================================================================================================================
\subsubsection{1D Case}
\label{EWTD}
\begin{figure}[!t]
\centering\includegraphics[width=\columnwidth]{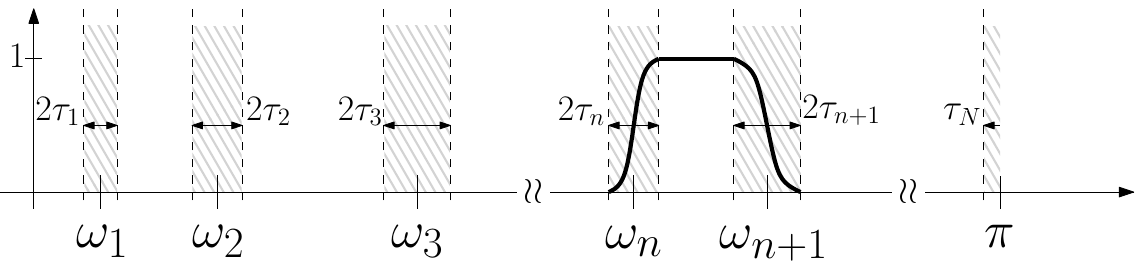}
\caption{Magnitude of the Fourier spectrum segmentation and empirical wavelet construction principle.}
\label{fig:ewt1d}
\end{figure}

For the reader's convenience, we start with a presentation of the 1D empirical wavelet transform (EWT). As described above, the aim of empirical wavelets is to separate harmonic 
modes of compact supports in the Fourier domain. The advantage is that the family of wavelets that results from it is driven by the information contained in the analyzed signal. To perform such 
decomposition, the first step consists in segmenting the magnitude of the Fourier spectrum of the input signal to find supports of meaningful harmonic modes. The 
Littlewood-Paley type wavelet filters are built upon the detected supports. Finally, the input signal is filtered by the obtained wavelet filter bank. Note that in 1D, the Hilbert 
transform can be applied to each component to extract the instantaneous amplitudes and frequencies providing an accurate time-frequency representation (see \cite{EWT1D}).\\
Assuming that the magnitude of the Fourier spectrum has $N$ modes and denoting $\omega$ the frequency (scaled between 0 and $\pi$), we define the set of (ordered) 
boundaries of each supports (see Figure~\ref{fig:ewt1d}) by $\left(\omega_n\right)_{n \in[0,\ldots,N]} \in \left[0,\pi\right]$ taking the convention $\omega_0=0$ and 
$\omega_N=\pi$. In \cite{EWT1D} and \cite{EWT2D}, the authors originally proposed different options to perform such boundary detection where the number of modes, $N$, was supposed 
to be known. More recently, a parameterless method (i.e. which automatically detects the number of meaningful modes) was proposed in \cite{scalespace} and we will use this approach 
throughout this paper.
Based on this set of boundaries and defining transition areas (illustrated by the shaded areas in Figure~\ref{fig:ewt1d}), the Littlewood-Paley filters are then easily defined in the Fourier domain by
\begin{equation}\label{eq:phi}
\hat{\Phi}_n(\omega)=
\begin{cases}
1 \quad \qquad \qquad \text{if}\;|\omega|\leq \omega_n-\tau_n\\
\cos\left[\frac{\pi}{2}\beta\left(\frac{1}{2\tau_n}(|\omega|-\omega_n+\tau_n)\right)\right] \\
\; \; \quad \qquad \qquad  \text{if}\;\omega_n-\tau_n\leq |\omega|\leq\omega_n+\tau_n \\
0 \quad \qquad \qquad \text{otherwise}
\end{cases}
\end{equation}
and
\begin{equation}\label{eq:psi}
\hat{\Psi}_n(\omega)=
\begin{cases}
1  \qquad\qquad \text{if}\; \omega_n+\tau_n\leq |\omega|\leq \omega_{n+1}-\tau_{n+1} \\
\cos\left[\frac{\pi}{2}\beta\left(\frac{1}{2\tau_{n+1}}(|\omega|-\omega_{n+1}+\tau_{n+1})\right)\right]  \\ 
\qquad\qquad\;\; \text{if}\; \omega_{n+1}-\tau_{n+1}\leq |\omega|\leq \omega_{n+1}+\tau_{n+1}\\
\sin\left[\frac{\pi}{2}\beta\left(\frac{1}{2\tau_n}(|\omega|-\omega_n+\tau_n)\right)\right] \\
\qquad\qquad\;\; \text{if}\; \omega_n-\tau_n\leq |\omega|\leq \omega_n+\tau_n\\
0 \qquad\qquad \text{otherwise.}
\end{cases}
\end{equation}
The function $\beta$ is a function observing the following conditions: $\forall x \in \left[ 0,1 \right], \beta(x)+\beta(1-x)=1$; $\forall x \in \mathbb{R}, \ x \ge 1, \beta(x)=1$ 
and $\forall x \in \mathbb{R}, \ x \le 0,\beta(x)=0$. A popular choice is given by $\beta(x)=x^4(35-84x+70x^2-20x^3)$ (see \cite{Daubechies}). It is proven in \cite{EWT1D} that, by 
properly choosing the coefficients $\tau_n$, the resulting wavelet family forms an adjusted tight frame.
%========================================================================================================================================
\subsubsection{2D Extension}
\begin{figure}[!t]
\begin{tabular}{cc}
\includegraphics[width=0.43\columnwidth]{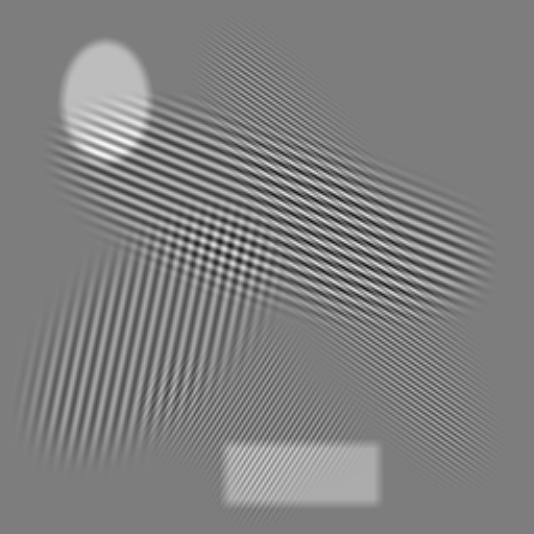} &
\includegraphics[width=0.43\columnwidth]{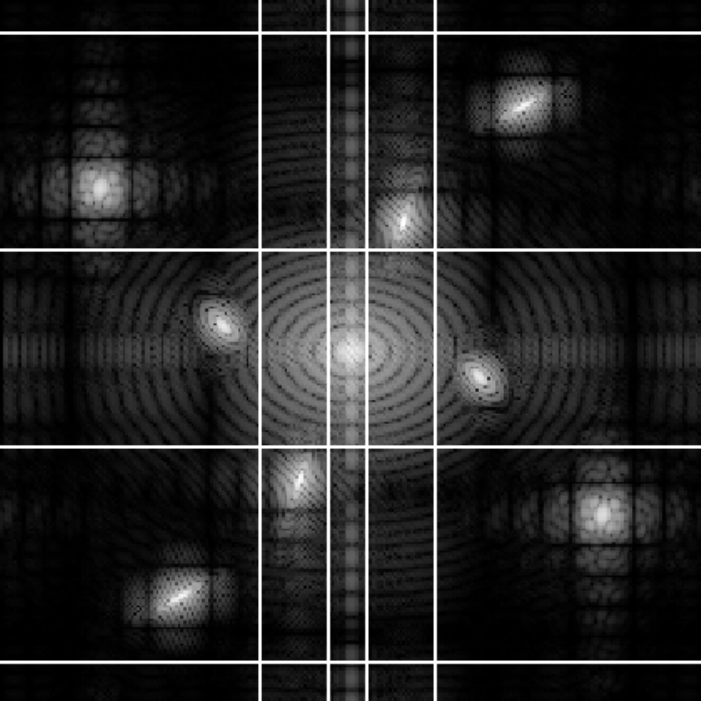} \\
simulated image & Tensor\\
\includegraphics[width=0.43\columnwidth]{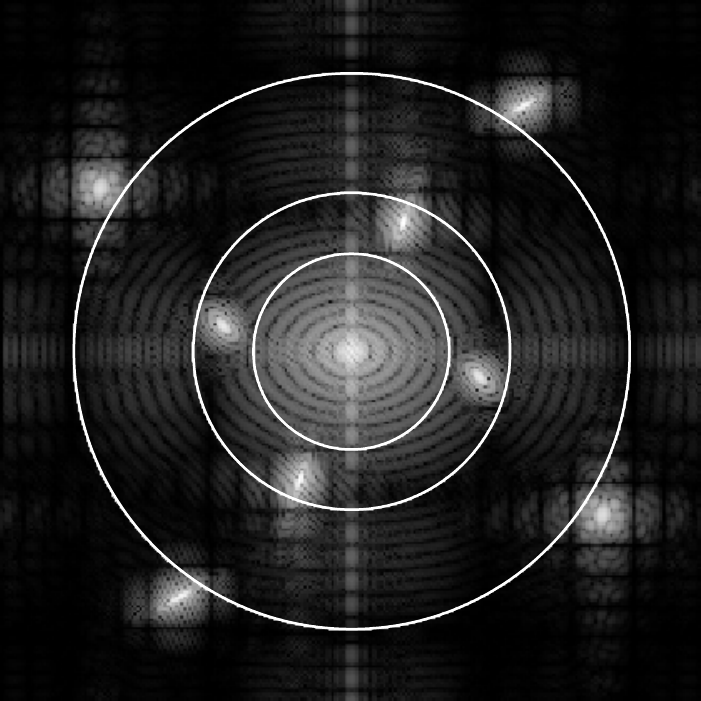} &
\includegraphics[width=0.43\columnwidth]{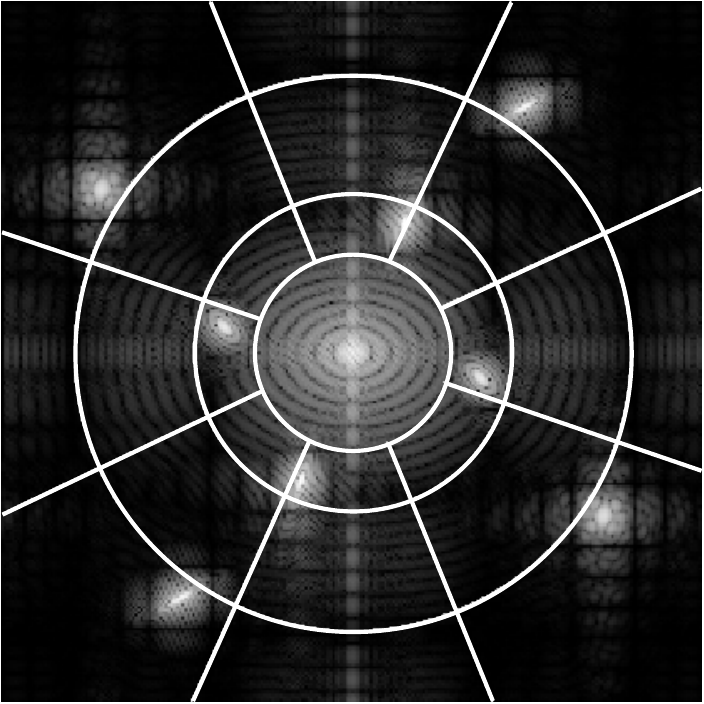} \\
Littlewood-Paley & Empirical Curvelet 1\\
\includegraphics[width=0.43\columnwidth]{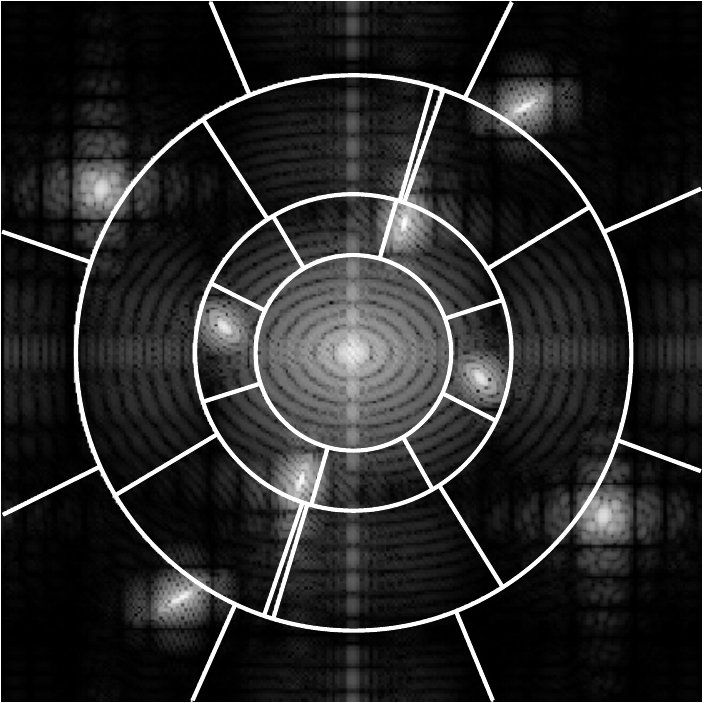} &
\includegraphics[width=0.43\columnwidth]{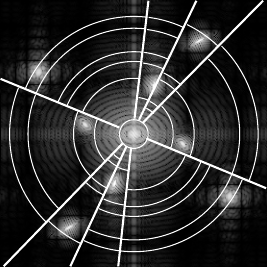} \\
Empirical Curvelet 2 & Empirical Curvelet 3\\
\end{tabular}
\caption{Fourier partitions corresponding to the different 2D empirical wavelets.}
\label{fig:part}
\end{figure}

In \cite{EWT2D}, the authors extend the EWT to several two-dimensional transforms for image processing purposes; in this paper, we will use them to extract texture features. In this section we recall four approaches which correspond to different partitioning of the Fourier domain. We refer the reader to \cite{EWT2D} for more details and examples. 
%Let still denote $\mathcal{I}$ an image and $\hat{\mathcal{I}}$ its Fourier transform.
%========================================================================================================================================
\paragraph{Tensor case (EWT2DT)} 
Like the classic 2D tensor wavelet transform, two sets of 1D empirical wavelets, $\{\Psi_n^r\}$ and $\{\Psi_n^c\}$ are used to process the rows and columns, respectively. The 
horizontal (vertical) set of filters is based on the boundaries detected on the spectrum corresponding to the average of all 1D row (column) spectra. Hence the set of 2D filters, 
$\{\Psi_{nm}^T(\x)\}$, is defined by the product of 1D filters (we denote $\x=(x,y)$): $\Psi_{nm}^T(\x)=\Psi_n^r(x)\Psi_m^c(y)$. An example of the obtained partition in the Fourier 
domain of a simulated image is given in the top-right image of Figure~\ref{fig:part}.
%========================================================================================================================================
\paragraph{Littlewood-Paley case (EWT2DLP)} 
\label{EWT2DLP}
In some image processing tasks, it can be more interesting to only focus on different scales and not pay attention to the orientations. The Littlewood-Paley filters aim at 
extracting components lying in specific scale ranges (with no preferred direction), such filters are supported over concentric rings (centered at the origin). Building empirical Littlewood-Paley wavelets is then equivalent to detect the radii of each rings. This can be efficiently achieved by using the polar Fourier domain \cite{Averbuch2006} since the same detection method used for tensor wavelets can be reused. If we denote $\omega=(\omega_x,\omega_y)$ the 2D frequency vector, then the Littlewood-Paley wavelets are defined in the Fourier domain by $\Psi_n^{LP}(\omega)=\Psi_n(|\omega|)$ where $\Psi_n$ are defined by (\ref{eq:phi}) and (\ref{eq:psi}). The left image in the second row of Figure~\ref{fig:part} illustrates the obtained partition in the Fourier domain on the synthetic image used in the previous section.
%========================================================================================================================================
\paragraph{Curvelet} 
The Littlewood-Paley construction described in the section above does not take into account orientations which are generally important information in images (especially for textures). In order to consider the orientation information, empirical wavelets can be built following the same philosophy as in the curvelet transform, providing an empirical curvelet transform (EWT2DC). The idea consists in partitioning the Fourier domain into angular sectors. Two options were initially proposed in \cite{EWT2D}: finding a set of global angular sectors which are scale independent (EWT2DC1) or finding a set of angular sectors per ring of scales (EWT2DC2). We also add a third option (EWT2DC3) which finds a set of scale radii per angular sectors. Again, the detection can be easily performed in the polar Fourier domain. The right image of the second row and images of the third row of Figure~\ref{fig:part} correspond to the EWT2DC1, EWT2DC2 and EWT2DC3 partitions on the same synthetic example, respectively.

%========================================================================================================================================
\section{Unsupervised texture segmentation algorithm}\label{algo}
\begin{figure}[!t]
\includegraphics[width=\columnwidth]{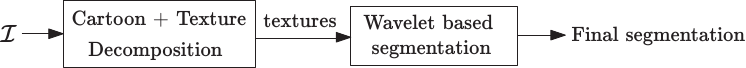}
\caption{Global segmentation strategy.}
\label{fig:seg}
\end{figure}

In most cases, images contain finite regions with or without several types of textures. The purpose of image segmentation is to extract those different regions, i.e if we denote 
$\Omega$ the image domain, we want to find a set $\{\Omega_i\}$ such that $\forall i,\Omega_i\subset\Omega$, $\forall i\neq j,\Omega_i\cap\Omega_j=\varnothing$ and 
$\Omega=\bigcup_i\Omega_i$. One challenging aspect of building feature vectors to characterize different regions is that those without textures are very different from 
regions with textures and the same type of feature vectors can fail to represent both types of regions. One solution to alleviate this problem relies on the fact that 
regions without textures do not contain details and thus should be mainly contained in the low frequency component of a wavelet decomposition. If at first glance 
such low frequency component seems to be sufficient to distinguish different non-textured regions, it is generally not accurate because such low frequency component is a very 
smooth function and does not contain regions with sharp boundaries. A better approach, initially suggested in \cite{aujolclassif,AujolMTS}, is to first decompose the input 
image into its cartoon and texture parts and then use specialized algorithms on each parts to eventually fusion the extracted information to get a unique final segmentation. In 
this paper, since we focus on texture segmentation, we first decompose the input image in its cartoon and texture parts and segment only the latter. We will not use the information 
contained in the cartoon part. The process we put forward is summarized in Figure~\ref{fig:seg}. The following sections describe the cartoon-texture decomposition model 
used in this paper and the final segmentation process used on the texture parts. 
%========================================================================================================================================
\subsection{Cartoon + Texture decomposition}
\begin{figure}[!t]
\begin{tabular}{ccc}
\includegraphics[width=0.28\columnwidth]{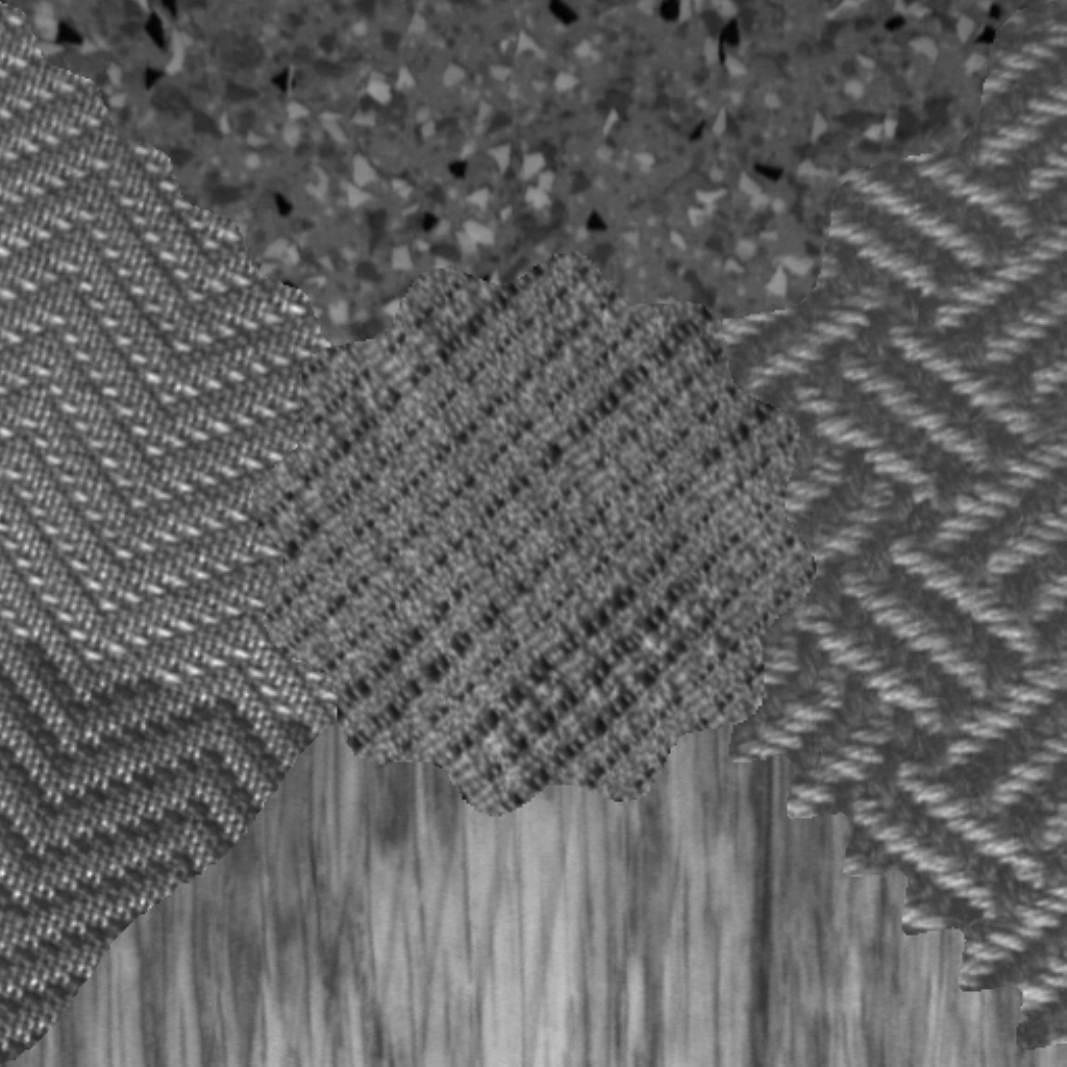} & \includegraphics[width=0.28\columnwidth]{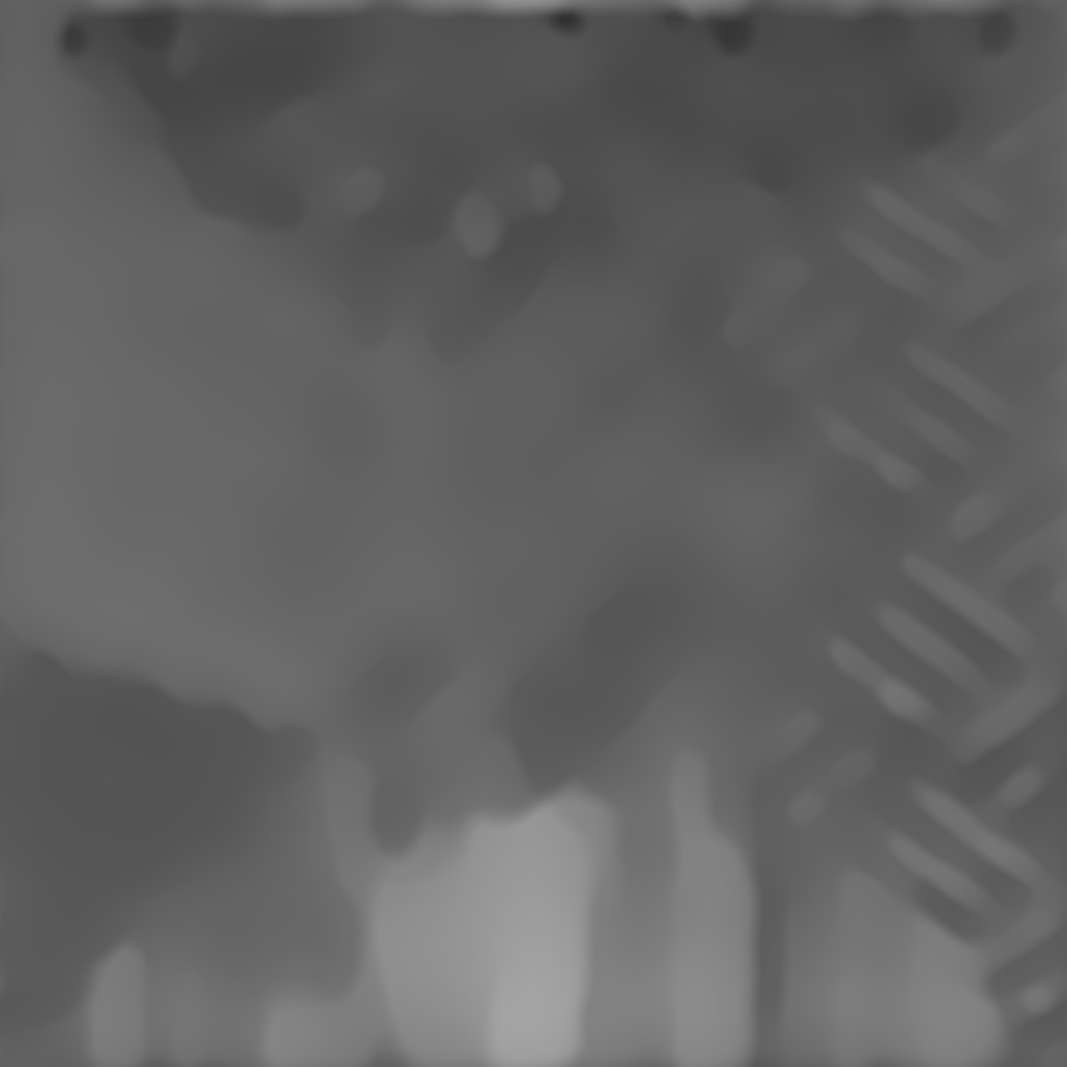} & \includegraphics[width=0.28\columnwidth]{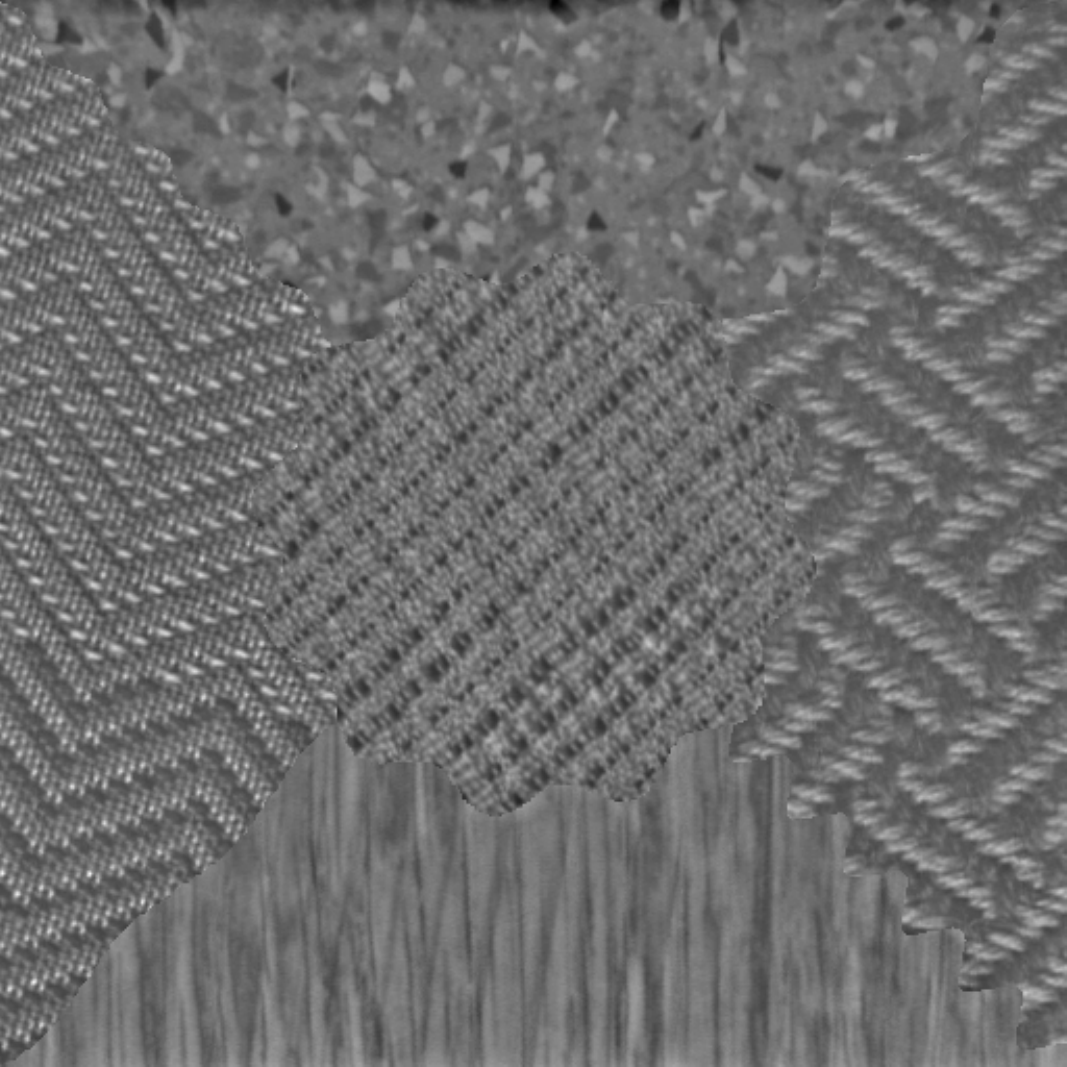} \\
\includegraphics[width=0.28\columnwidth]{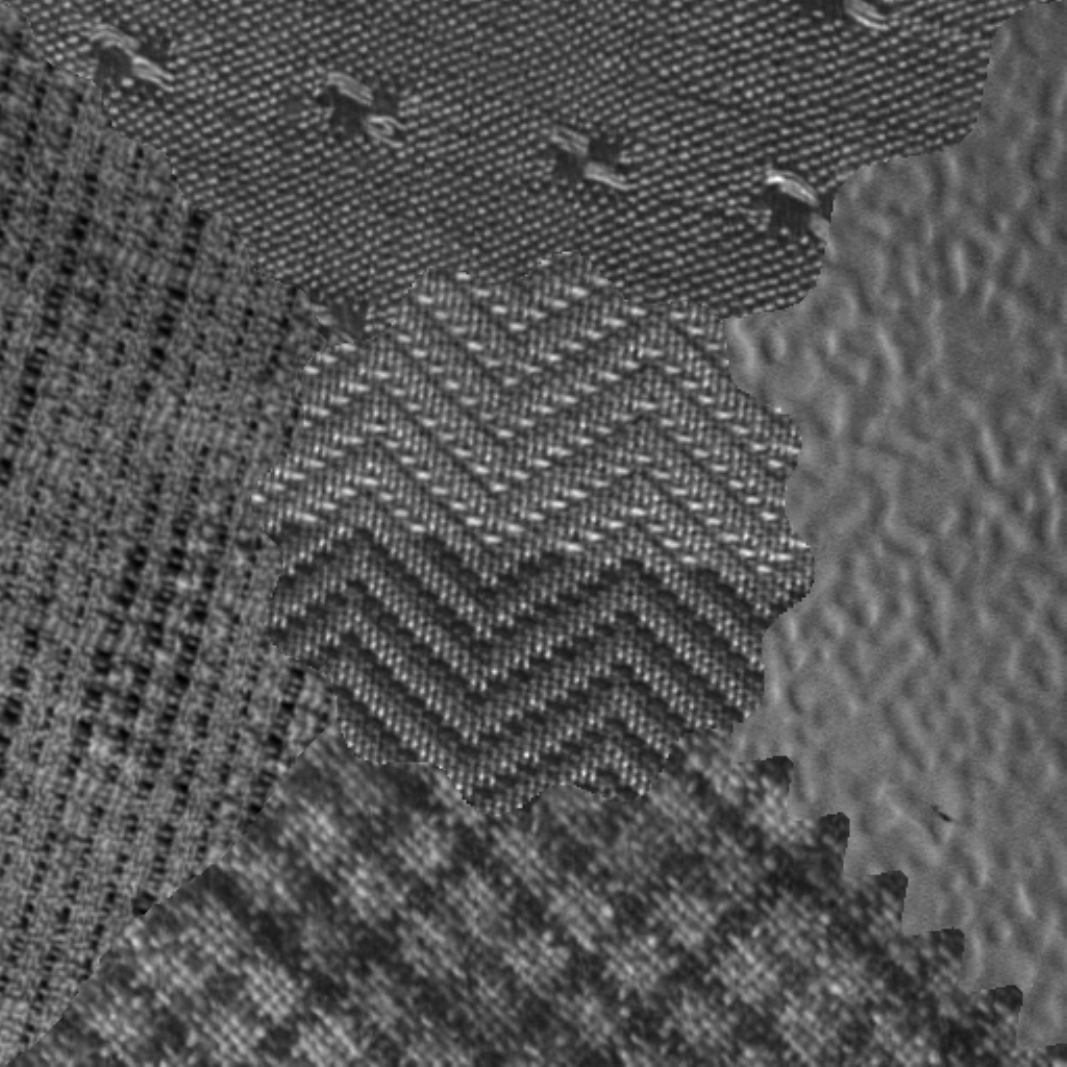} & \includegraphics[width=0.28\columnwidth]{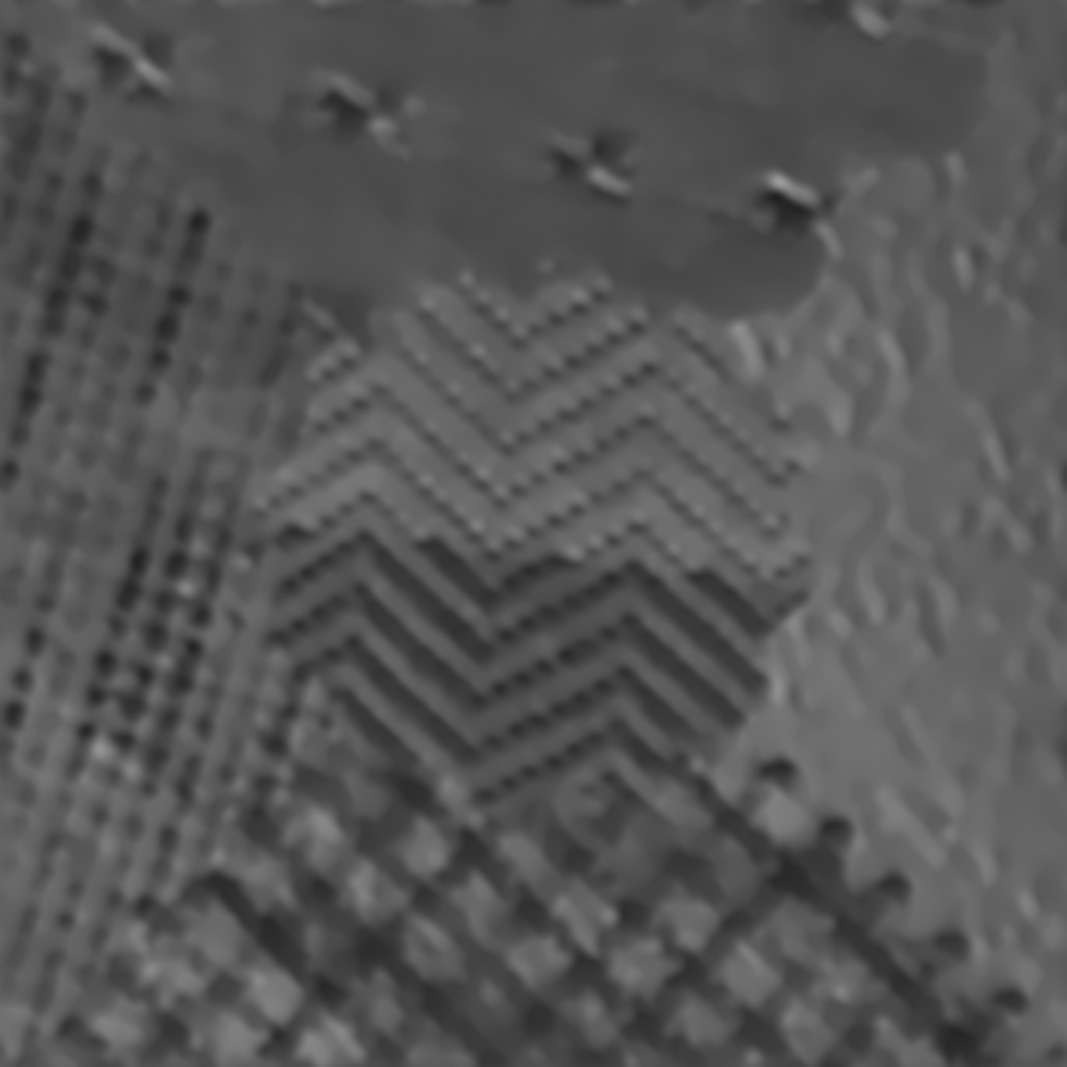} & \includegraphics[width=0.28\columnwidth]{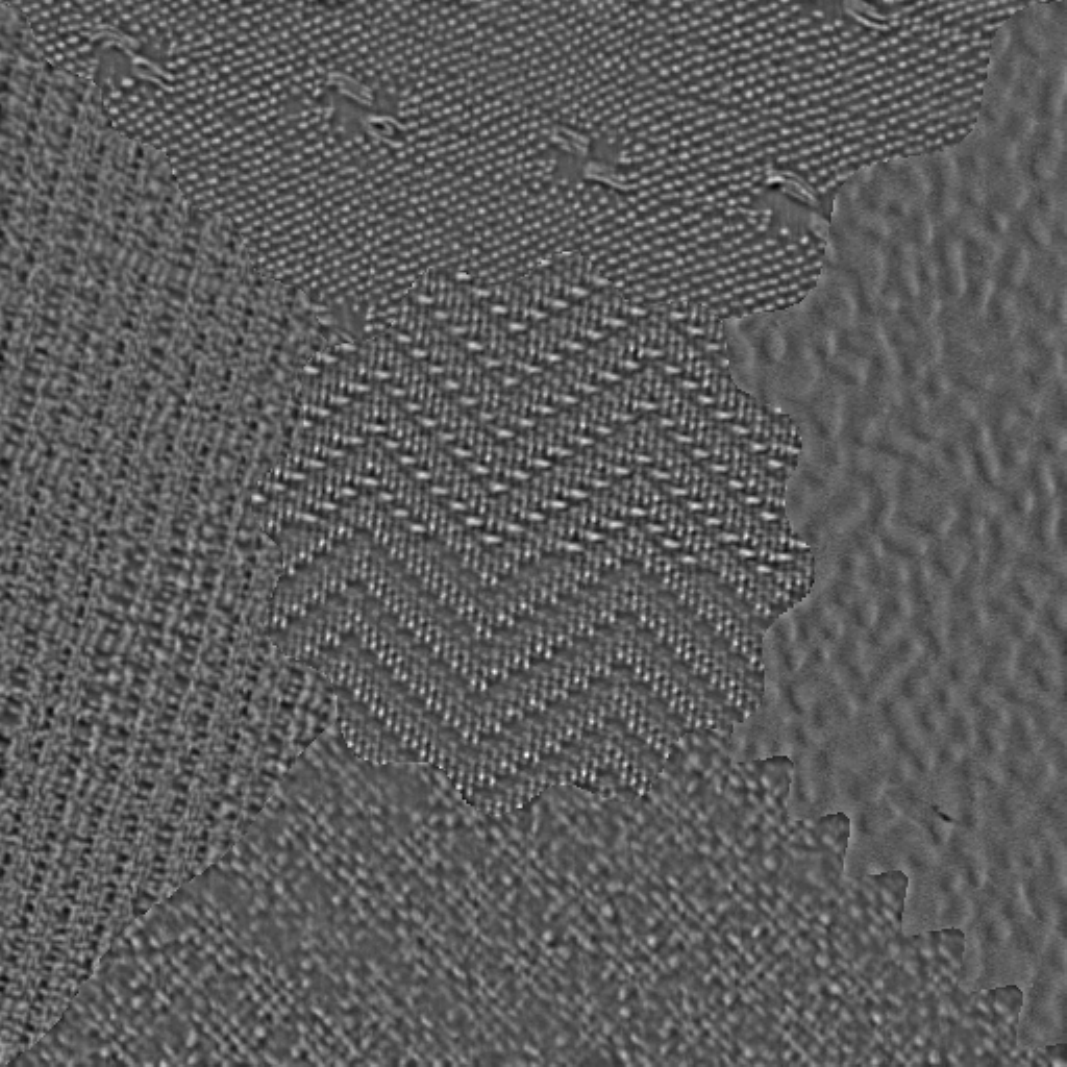} \\
Image & Cartoon & Texture\\
\end{tabular}
\caption{Cartoon and texture components.}
\label{fig:uv}
\end{figure}

As mentioned above, we focus only on the texture segmentation, thus we choose to first decompose the input image into its cartoon and texture parts and segment only the latter. 
Such decomposition models were widely studied in the last decade after the seminal work of Meyer \cite{Gspace}. This work has its roots in the famous Rudin-Osher-Fatemi (ROF) 
\cite{ROF,Chambolle,Goldstein} model. It minimizes a total variation (TV) based functional in order to remove the details in the image while keeping objects with sharp edges, i.e providing a cartoon 
version of the image. Meyer showed that the difference between the input image and its cartoon version does not necessarily contains all the expected details, more specifically some 
oscillatory (i.e textural) information can be lost. Thus he introduced a space of oscillating functions, denoted $G$ to model 
the textures, that he used in a modified version of the ROF model. If we denote $\mathcal{I}$ the input image, $u$ and $v$ the cartoon and texture components parts, 
respectively, the decomposition model consists in solving
\begin{align}
\label{rof}
(\hat{u},\hat{v})&=\underset{u\in BV,v\in G}{\arg\min} \|u\|_{TV}+\mu \|v\|_G, \\
&\text{subject to}\quad \mathcal{I}=u+v. \notag
\end{align}
The space $BV$ is the space of functions of bounded variations and is characterized by the semi-norm $\|u\|_{TV}=\|\nabla u\|_1$. The total variation will be small when $u$ is a 
cartoon component (i.e a piecewise like function) while the $G-$norm will be small when $v$ is a high oscillating function, the parameter $\mu$ controls the frontier between the 
two spaces (we refer the reader to \cite{Gspace} for more details). Such models and their properties were widely investigated in the literature 
\cite{Luminita2003a,Aujol2005a,Bregman,MTS,Wang}. In particular, the most handful formulation is given by 
\begin{align}
(\hat{u},\hat{v})&=\underset{u\in BV,v\in G_\mu}{\arg\min} \|u\|_{TV}+J^*\left(\frac{v}{\mu}\right)+\frac{\lambda}{2}\|\mathcal{I}-(u+v)\|_2^2,
\end{align}
where $G_\mu=\{v\in G\rvert \|v\|_G\leq \mu\}$ and $J^*$ is the characteristic function over $G_1$, i.e $J^*(v)=0$ if $v\in G_1$ and $J^*(v)=+\infty$ otherwise. The advantage of this formulation is that it can be easily solved numerically. In this paper, we used the algorithm described in \cite{Bregman} to perform all our experiments. 
Limited literature discusses the choice of the parameters $\mu$ and $\lambda$. Empirically, $\mu$ controls the sensibility to the frequential part, the smaller $\mu$ the more 
oscillating the textures are; while $\lambda$ controls the regularization of the cartoon part. Following the work of \cite{MTS}, we choose to set $\mu=\frac{\mathcal{N}}{2}$ (we 
recall that $\mathcal{N}$ is the image size). On the other hand, we fixed $\lambda = \frac{\omega_1^r}{2}$, where $(\omega_n^r)_{n 
\in [0,\ldots,N]}$ is the family of radii identified in Section~\ref{EWT2DLP}. Figure~\ref{fig:uv} provides an example of the obtained decomposition of two images from the Outex dataset.
%========================================================================================================================================
\subsection{Wavelet based segmentation of the texture part}
The last step consists in computing the final segmentation from the texture part. Here, we use wavelet features to characterize the textural information and follow the method 
described in Section~\ref{featproc}. Because the texture part contains mainly frequency information we will only use the bandpass wavelet filters to build the feature vectors 
associated with each pixels (including the post-processing). In other terms, we use only the filters $\{\Psi_n\}$ for $n\geq 1$ and get rid of $\Psi_0=\Phi_0$. As described 
earlier, whatever wavelet transform is chosen, it is always equivalent to building a feature vector $f(\x)\in\mathbb{R}^K$ (the ordering of the wavelet filters does not have any 
influence on the segmentation process since it only corresponds to reorder the feature space. The obtained feature vectors are finally used to feed one of the clustering algorithm 
described in Section~\ref{clustalgo} in order to obtain the segmentation of the texture part.
%========================================================================================================================================
\section{Experiments}\label{experiments}
\subsection{Algorithm options}\label{algooptions}
In this section, we present the obtained results when applying the algorithm described in the previous section. We test all possible combinations according to the following options:
\begin{itemize}
 \item the wavelet transform: 
 \begin{itemize}
    \item undecimated standard wavelet transform for Coiflet, Daubechies, Symmlet wavelets,
    \item decimated standard wavelet transform for Coiflet, Daubechies, Symmlet wavelets,
     \item Packet wavelet transform for Coiflet, Daubechies, Symmlet wavelets,
    \item Gabor wavelets,
     \item Meyer wavelets,
    \item Curvelets,
    \item EWT2DT,
    \item EWT2DLP,
    \item EWT2DC1, EWT2DC2, EWT2DC3,
 \end{itemize}
\item the feature extraction method: Energy, Entropy or LBP,
\item the final clustering method: \textit{k-means} or Nystr\"om,
\item the clustering associated distance: Euclidean ($\ell^2$), cityblock ($\ell^1$), correlation or cosine.
\end{itemize}

%========================================================================================================================================
\subsection{Datasets description}\label{testset}
\begin{figure}[!t]
\begin{center}
\begin{tabular}{cccc}
\includegraphics[width=0.215\columnwidth]{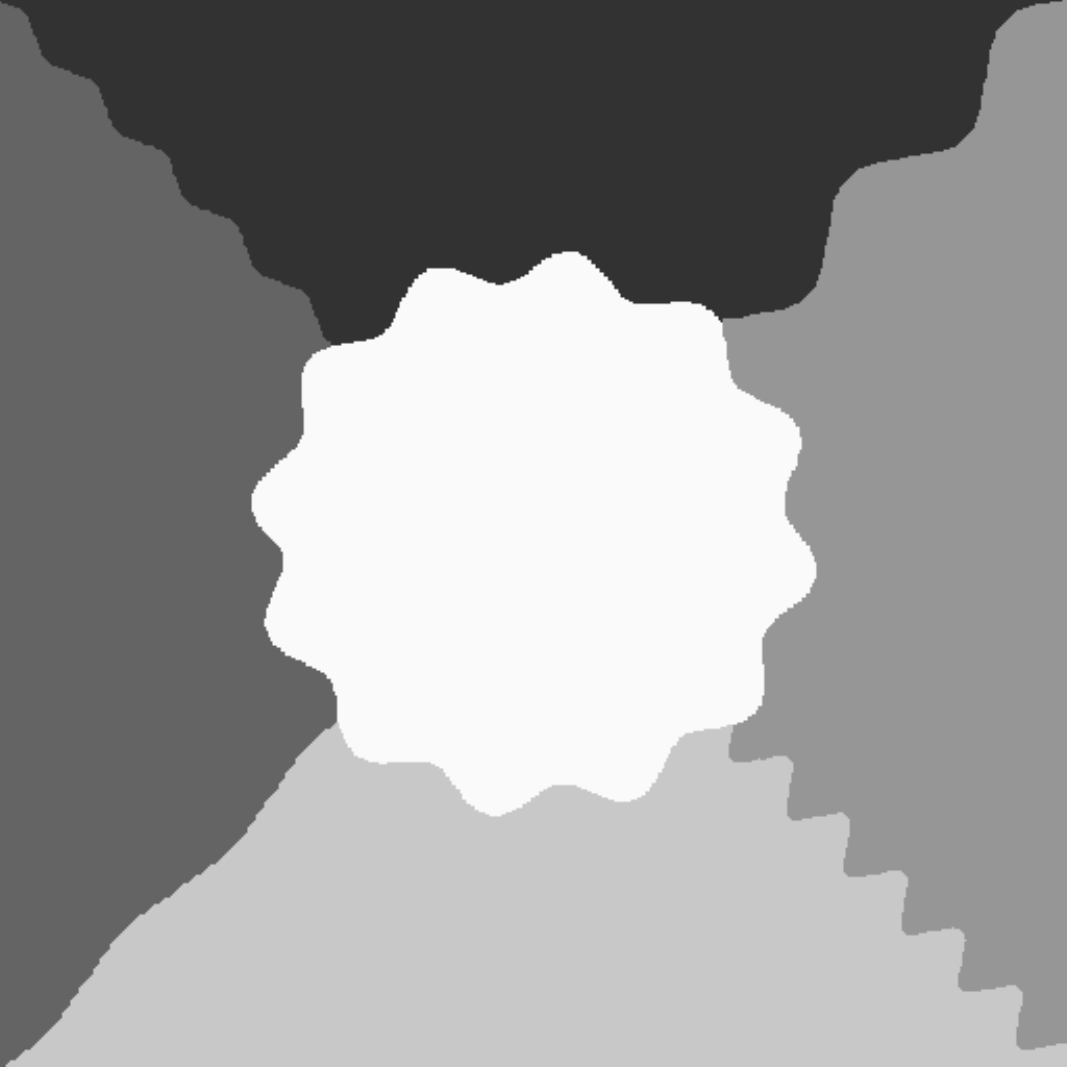}&
\includegraphics[width=0.215\columnwidth]{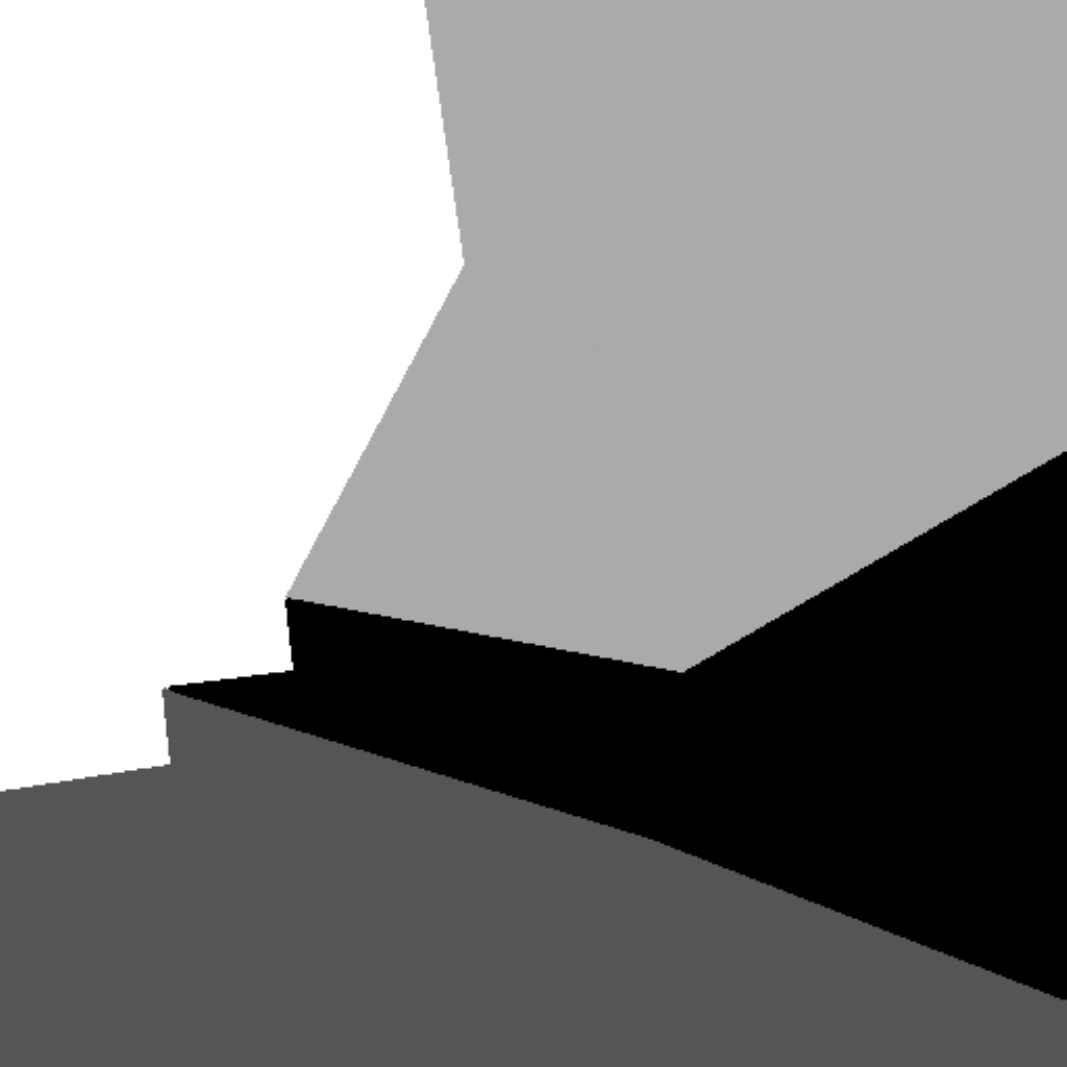}&
\includegraphics[width=0.215\columnwidth]{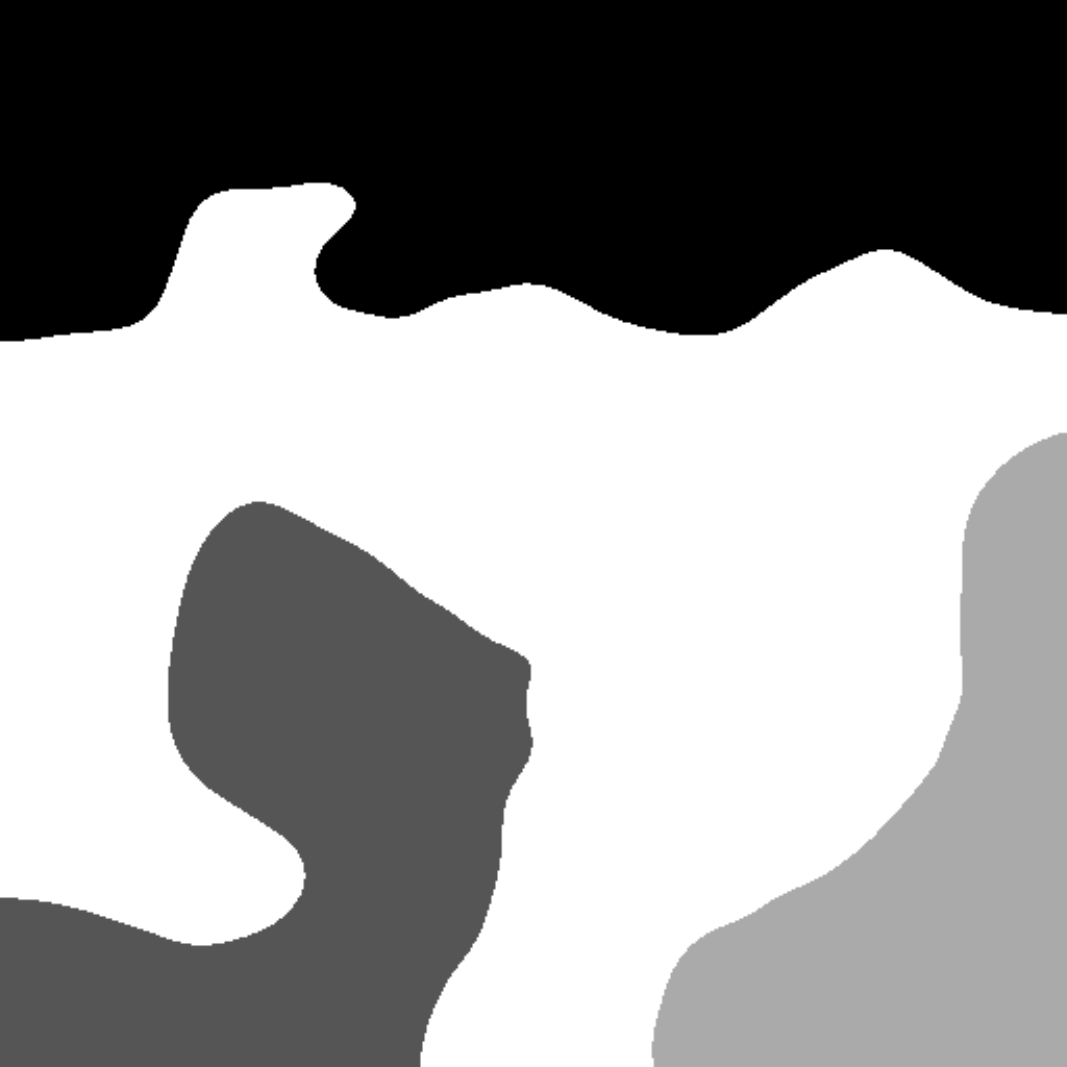}&
\includegraphics[width=0.215\columnwidth]{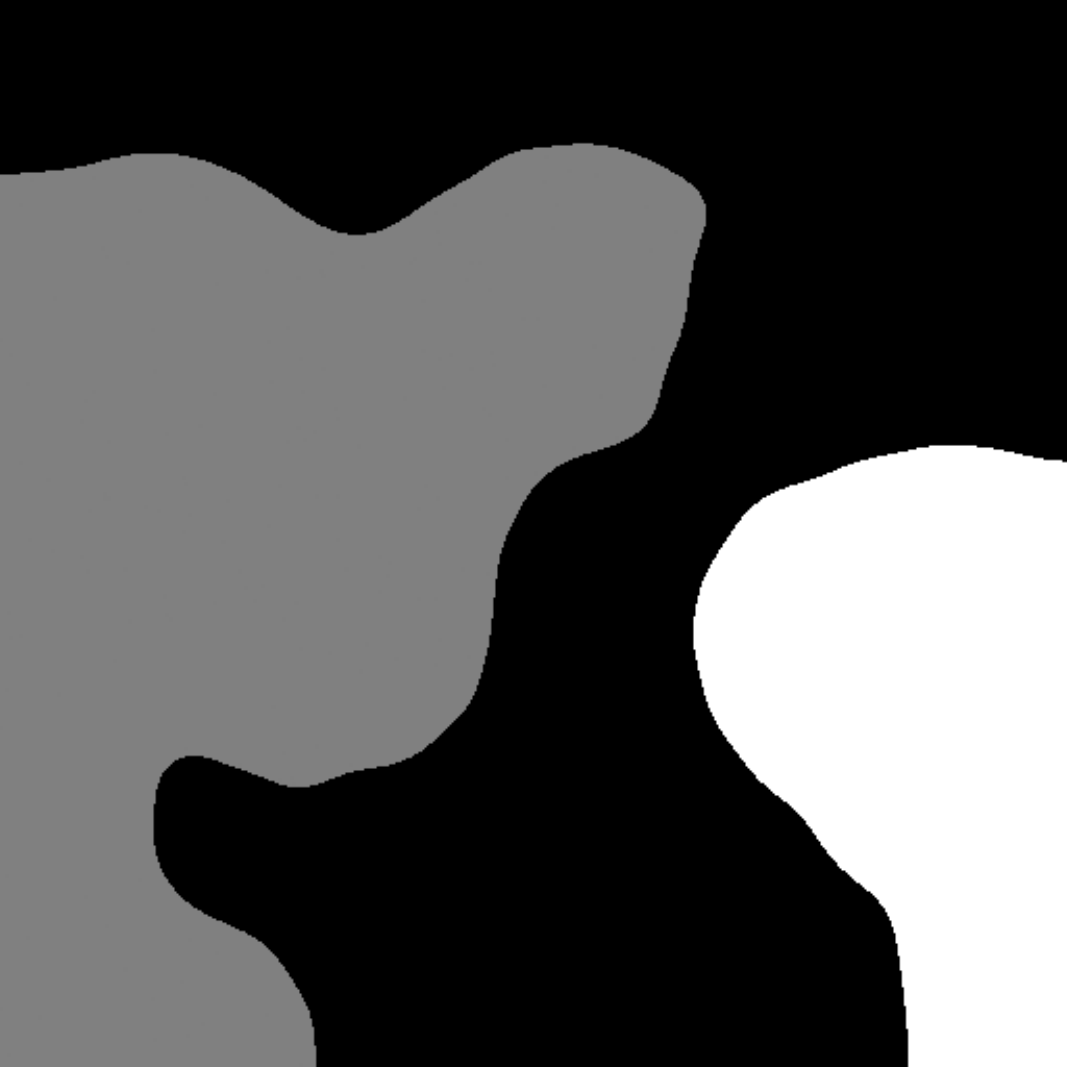} \\
\includegraphics[width=0.215\columnwidth]{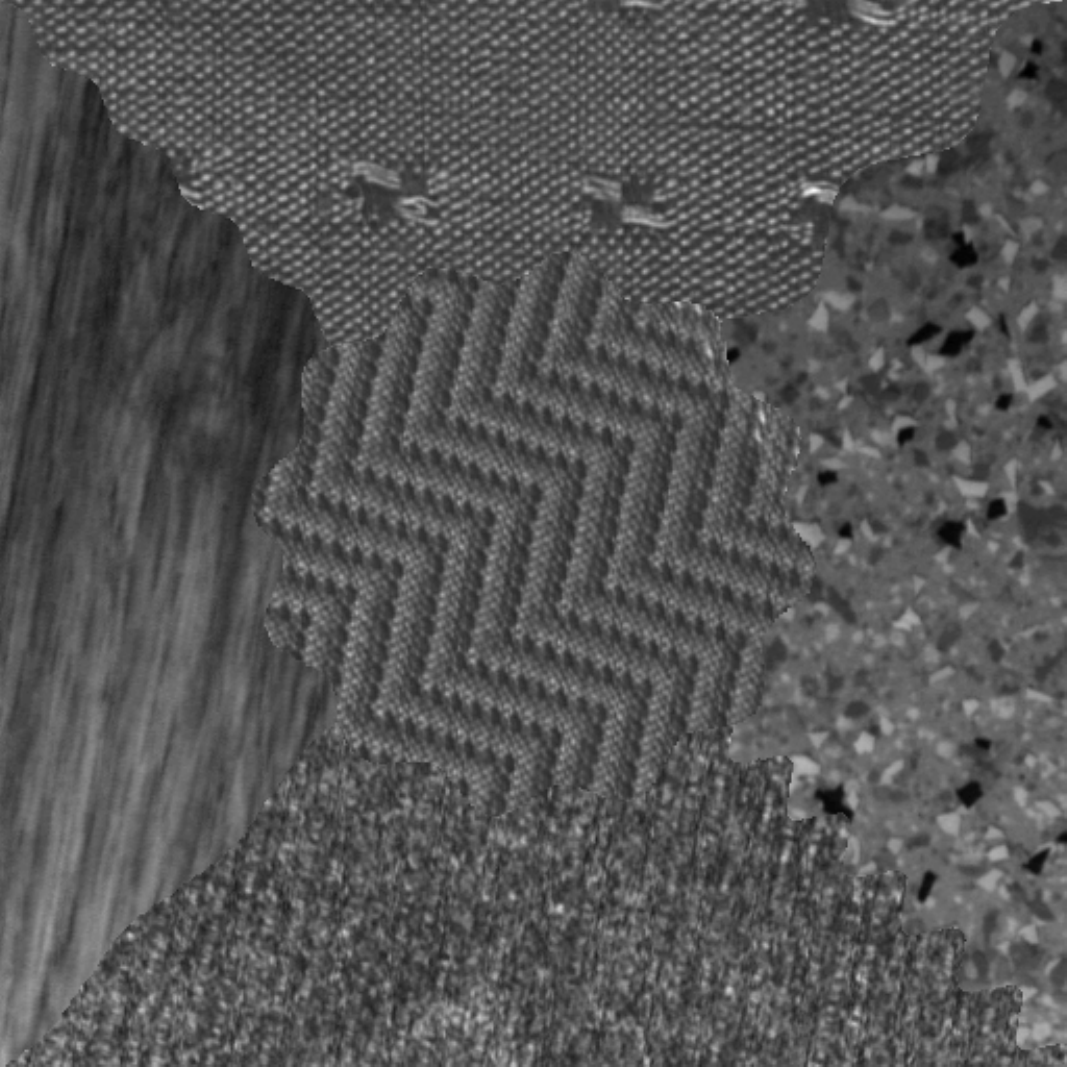} &
\includegraphics[width=0.215\columnwidth]{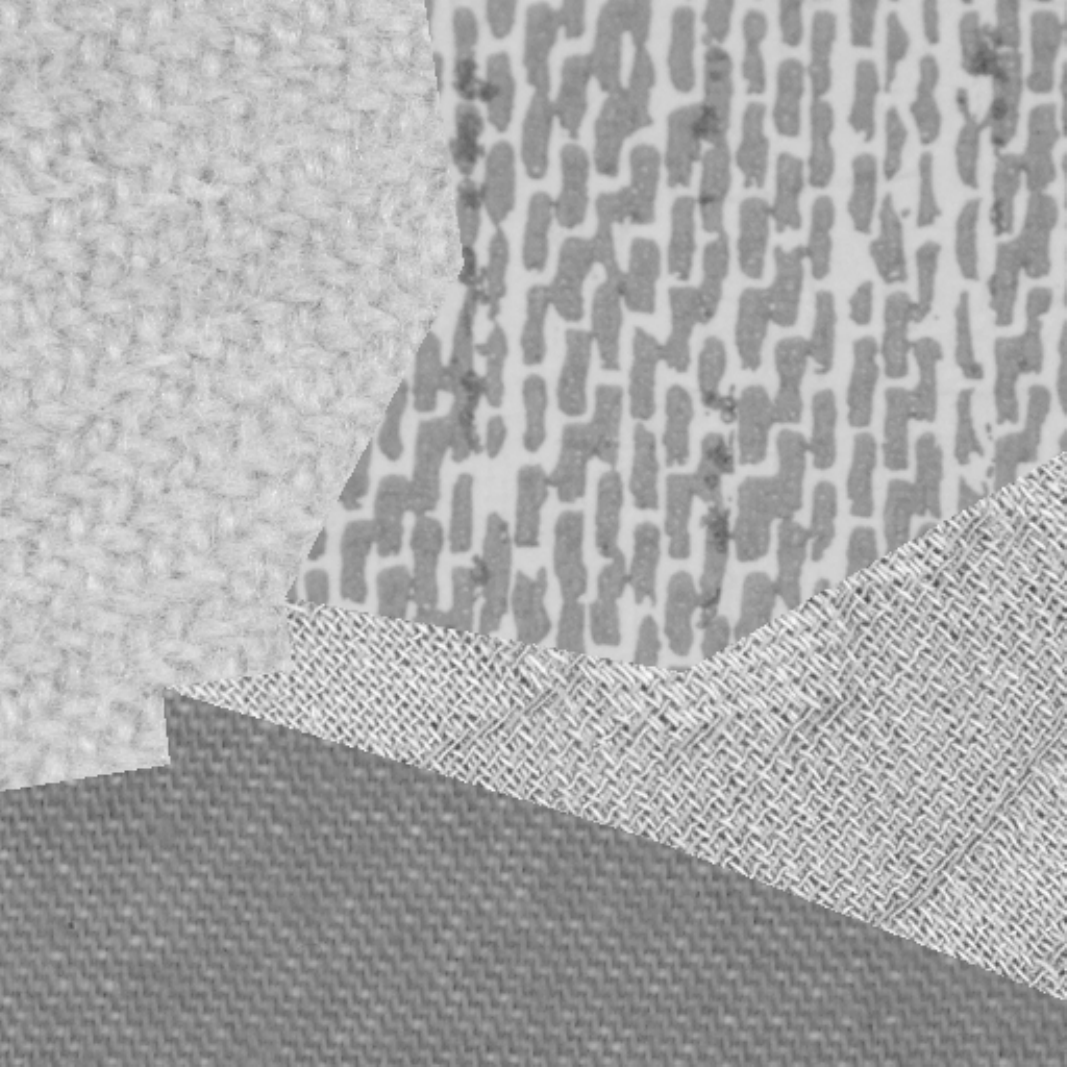} &
\includegraphics[width=0.215\columnwidth]{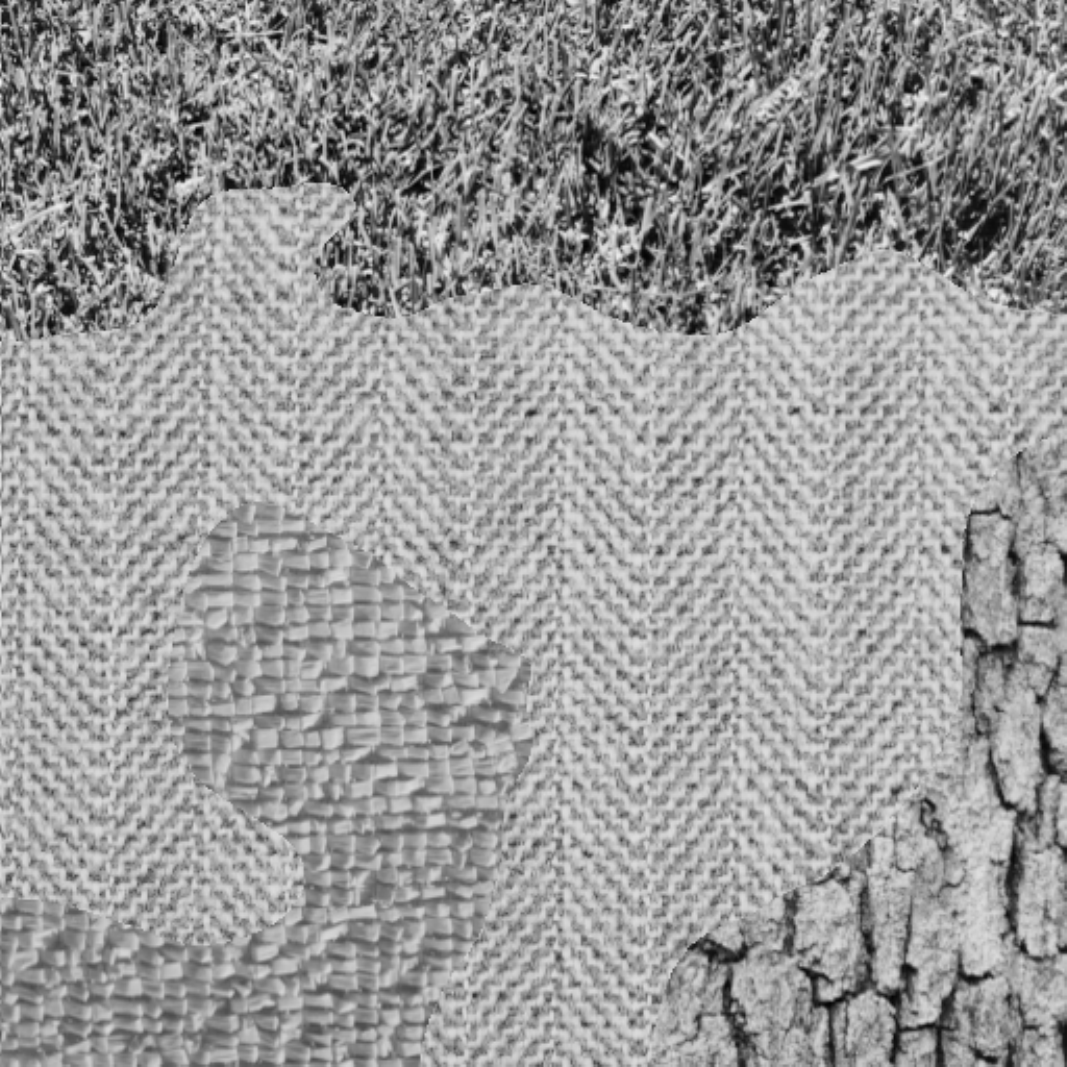} &
\includegraphics[width=0.215\columnwidth]{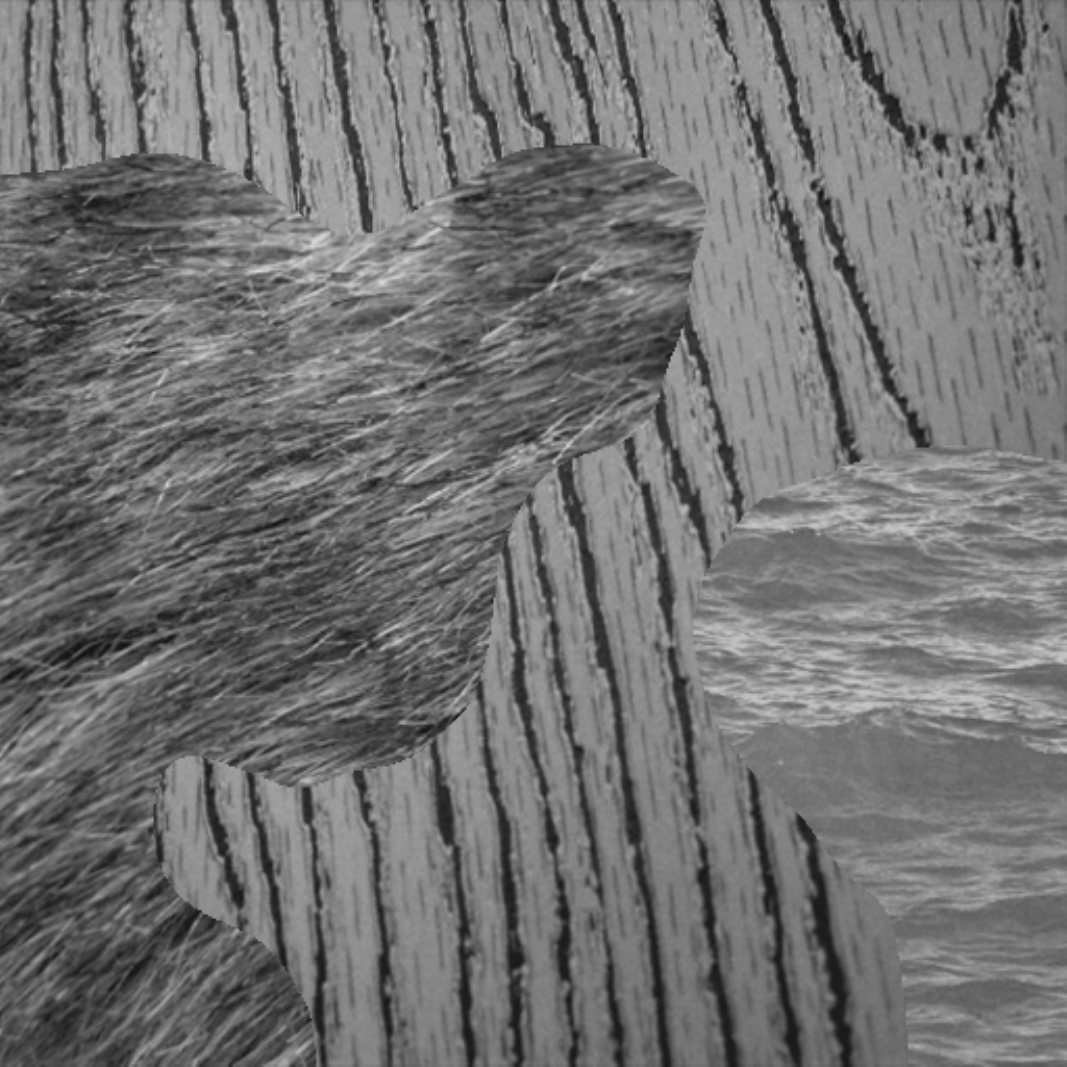}
\end{tabular}
\end{center}
\caption{The different used ground-truths (top row) and some corresponding test image samples (bottom row).}
\label{fig:dataset}
\end{figure}

In order to assess the efficiency of the different wavelet features, we built a large variety of test images by using textures from four popular texture datasets: 
Outex~\cite{Outex}, Brodatz~\cite{Brodatz}, ALOT~\cite{ALOT} and UIUC~\cite{UIUC}. The Outex dataset already has a hundred composite texture images. They are 
generated by 
mixing twelve different texture images according to the regions depicted by the ground-truth images shown in the top-left row of Figure~\ref{fig:dataset}. For each other 
datasets, 
we built similar sets of test images following the same procedure: we selected thirteen pristine textures and randomly composed a hundred test images using the different 
ground-truth images given on the top row of Figure~\ref{fig:dataset}. The size of all test images is $512\times 512$ pixels and are encoded on 256 gray levels. Moreover, the total 
number of regions is known and depends on the used ground-truth, this information is provided to the clustering algorithms.

%========================================================================================================================================   
\subsection{Benchmark Criteria}
We test the algorithm on the test sets described above in order to compare the performance of different wavelet features. Different approaches to compare segmentation results were 
proposed in the literature, each one corresponding to different interpretations of the notion of partition. A partition which evaluates the true boundaries of the image can be 
defined as a boundary-based interpretation. The pairs-of-pixels interpretation consists in classifying (into two classes) all pairs of pixels as belonging to the same cluster or 
not. The most used metric is the region-based interpretation which considers regions made of segmented pixels. We refer the reader to the review paper \cite{3Evaluation} for 
further insights.\\ 
In this paper, we choose to use the region-based interpretation. The following subsections describe different metrics associated with such approach and used in our experiments. 
The 
reader is referred to \cite{3Evaluation,benchmark,benchmark2} for more details and interpretations of these metrics. In what follows we consider gray scale textured 
images. We also assume that we have access to the ground-truth segmentation. We will denote $\mathcal{I}$ the original (test) image, $\mathcal{N}^2$ its number of pixels, 
$\mathcal{P}_G=\{R_1',...,R_N'\}$ the ground-truth segmentation and $\mathcal{P}_S=\{R_1,...,R_N\}$ the obtained segmentation where $N$ is the number of clusters. Given a region 
$R$, $|R|$ will denote its cardinality, i.e the number of pixels in the region $R$. All these metrics provide a number between 0\% and 100\%, the latter corresponding to perfect 
segmentation. The scores provided in the Results section, 
Section~\ref{results}, correspond to the sum of all these metrics divided by the number of metrics. All these metrics are computed via the \textit{eval\_segm} function available 
in the \textit{SEISM Toolbox}\footnote{\url{https://github.com/jponttuset/seism}}.

\subsubsection{Normalized Variation of Information}  \label{nvoi}
In \cite{VOI}, the variation of information was introduced to measure the information difference between two clusterings. The author uses the entropy, $H(\mathcal{P}_S)$, and mutual information, $I(\mathcal{P}_S,\mathcal{P}_G)$, of segmentations to define the so-called Variation of Information (see \cite{VOI} for all details) 
\begin{align}
\label{voi}
VoI(\mathcal{P}_S,\mathcal{P}_G)=H(\mathcal{P}_S)+H(\mathcal{P}_G)-2I(\mathcal{P}_S,\mathcal{P}_G).
\end{align}
This quantity can be normalized by $\log N$ to get the \textbf{Normalized Variation of Information} (NVOI) metric. Although this method gives some interesting theoretical 
properties, its perceptual meaning is still unclear.

\subsubsection{Swapped Directional Hamming Distance}
The basic idea of this metric is to evaluate the similarity between regions from two segmentations by finding the region $R$ with the maximum overlap for each region $R'$ from the ground-truth segmentation. The directional Hamming distance \cite{SDHD} from a partition $\mathcal{P}_S$ to a partition $\mathcal{P}_G$ is then defined as
\begin{align}
D_H(\mathcal{P}_S \Rightarrow  \mathcal{P}_G)=\mathcal{N}^2-\sum_{R'\in  \mathcal{P}_G}\max_{R \in \mathcal{P}_S } |R'\cap R|.
\end{align}
The \textbf{Swapped Directional Hamming Distance} (SDHD) corresponds to $D_H(\mathcal{P}_G \Rightarrow  \mathcal{P}_S)$.

\subsubsection{van Dongen Distance}
The \textbf{van Dogen} (VD) distance \cite{VD} is a symmetric extension of the directional Hamming distance:
\begin{align}
d_{vD}(\mathcal{P}_S,\mathcal{P}_G) = D_H(\mathcal{P}_S \Rightarrow  \mathcal{P}_G)+D_H(\mathcal{P}_G \Rightarrow  \mathcal{P}_S)
\end{align}

\subsubsection{Swapped Segmentation Covering}
The overlap between two regions can be used to assess the pixel-wise classification in a recognition task \cite{SSC1}. It is defined as
\begin{align}
\mathcal{O}(R,R')=\frac{|R\cap R'|}{|R \cup R'|}.
\end{align}
In \cite{SSC2}, the author defines the covering of a segmentation $\mathcal{P}_S$ by a segmentation $\mathcal{P}_G$ as
\begin{align}
\mathcal{C}(\mathcal{P}_G \Rightarrow  \mathcal{P}_S) = \frac{1}{\mathcal{N}^2}\sum_{R\in \mathcal{P}_S}|R|\cdot\max_{R'\in \mathcal{P}_G}\mathcal{O}(R,R').
\end{align}
The \textbf{swapped segmentation covering} (SSC) is defined by $\mathcal{C}(\mathcal{P}_S \Rightarrow  \mathcal{P}_G)$.
 
\subsubsection{Bipartite Graph Matching}
The \textbf{Bipartite Graph Matching} (BGM) metric aims to measure the maximum overlap between regions of two segmentations by using a bijective matching method. The two given partitions $\mathcal{P}_S$ and $\mathcal{P}_G$ are seen as one common set of nodes $\{R_1,...R_N\}\cup\{R_1',...R_N'\}$ in a graph where an edge is inserted between each pair of nodes with a certain weight. Given such graph, the sum of all weights, denoted $w$, can be used to define the maximum-weight bipartite graph matching 
measure
\begin{align}
\mathcal{BGM}(\mathcal{P}_S,\mathcal{P}_G)= 1-\frac{w}{\mathcal{N}^2}.
\end{align}
We refer the reader to \cite{BGM} for more details on this metric and \cite{Graph} for the optimization procedure used to obtain $w$.

\subsubsection{Bidirectional Consistency Error} \label{bce}
In \cite{BCE}, the author proposed a measure that does not tolerate refinement to oversegmentation. This measure, the \textbf{Bidirectional Consistency Error} (BCE) is defined by
\begin{align}
BCE(\mathcal{P}_S,\mathcal{P}_G) &= 1-\\ \notag
&\frac{1}{\mathcal{N}} \sum_{\substack{R\in \mathcal{P}_S \\ R' \in \mathcal{P}_G}} |R\cap R'| \min{ \left\{\frac{|R\cap R'|}{|R|},\frac{|R\cap R'|}{|R'|}\right\}}.
\end{align}

%========================================================================================================================================
\subsection{Results}\label{results}
In this section, we experimentally investigate the influence of the different options (window's kernel size, post-processing, clustering method and type of wavelet). 
Except in the last section, we only used a subset of possible wavelets in order to reduce the amount of computation. Curvelets, Gabor wavelets and Empirical wavelets are used to 
test the influence of the kernel's size while only Curvelets and Empirical Curvelets are used to test the post-processing and clustering method influence. We choose these wavelets 
because we expect them to be the best families based on their properties in terms of adaptability and how they characterize the geometrical information. The last section will 
present a broader comparison using other families of wavelets and will confirm this choice.

\subsubsection{Influence of the window's kernel size}\label{winsize}
We first investigate the choice of the window's kernel size used by the post-processing operation during the wavelet feature extraction step. No theoretical results exist 
regarding a 
potential optimal window size and it is usually empirically chosen. Therefore, we first ran the algorithm for the wavelet subset selected above using the \textit{k-means} 
clustering (the \textit{cityblock} distance was used) on the ALOT, UIUC, Outex and Brodatz datasets. We computed the average of all previously defined metrics. These experiments 
provided us the results given in Tables~\ref{tab:alots}, \ref{tab:uiucs}, \ref{tab:brodatzs} and \ref{tab:outexs}. If some variations appear between the different datasets, we can 
observe that a kernel size in the range of 19 to 25 gives the best results. Moreover, we see that the optimal size is almost constant within a given dataset. In the next 
experiments, we will use the optimal sizes we found for each dataset: 19 for ALOT and Outex, 25 for UIUC and Brodatz. 

\begin{table}[!t]
\processtable{ALOT: influence of the window's kernel size\label{tab:alots}}
{\begin{tabular*}{\columnwidth}{@{\extracolsep{\fill}}c|ccccc} \toprule
\diagbox{\textbf{Wavelet}}{\textbf{Size}} & \textbf{3} & \textbf{5} & \textbf{7} & \textbf{9} & \textbf{11} \\ \hline
Curvelet & 73.78  & 73.78  & 74.55  & 75.07  & 75.35   \\ \hline
EWTC1 	& 69.05  & 72.64  & 74.85  & 76.26  & 77.55   \\ \hline
EWTC2 	& 69.15  & 72.72  & 74.45  & 76.65  & 77.63  \\ \hline
EWTC3 	& 68.71  & 70.38  & 71.29  & 72.58  & 72.71  \\ \hline
EWT2DT 	& 70.01  & 74.43  & 76.42  & 76.58  & 77.39  \\ \hline
Gabor 	& 73.03  & 73.80  & 75.59  & 77.44  & 78.07  \\ \hline
\diagbox{\textbf{Wavelet}}{\textbf{Size}} & \textbf{13} & \textbf{15}& \textbf{17}& \textbf{19} & \textbf{21}  \\ \hline
Curvelet & 75.86  & 75.69  & 75.49 & 75.45  &  \textbf{76.38} \\ \hline
EWTC1 	& 78.43  & 78.40  & \textbf{79.03} & 78.68  & 78.26 \\ \hline
EWTC2 	& 78.02  & 78.46  & 78.05 & \textbf{78.70}  & 78.67 \\ \hline
EWTC3 	& 72.35  & \textbf{73.41} & 72.97 & 72.99  & 72.65 \\ \hline
EWT2DT 	& 78.55  & 78.06  & 78.09 &\textbf{78.60}  & 78.25 \\ \hline
Gabor 	& 78.60  & 78.26  & 78.62 & 78.49  & 78.67 \\ \hline
\diagbox{\textbf{Wavelet}}{\textbf{Size}} & \textbf{23}& \textbf{25} & \textbf{27}& \textbf{29}& \textbf{31} \\ \hline
Curvelet & 75.23  & 75.90  & 75.63  & 75.16  & 75.39 \\ \hline
EWTC1 & 78.53  & 78.25  & 77.75  & 77.05  & 77.79 \\ \hline
EWTC2 & 77.50  & 78.38  & 77.91  & 76.45  & 76.83 \\ \hline
EWTC3 & 72.31  & 72.25  & 71.42  & 71.30  & 71.33 \\ \hline
EWT2DT & 77.82  & 77.89  & 77.08  & 76.63  & 76.03 \\ \hline
Gabor & 77.68  & \textbf{78.89}  & 78.01  & 78.09  & 77.40 \\ \hline
\end{tabular*}}{}
\end{table}

\begin{table}[!t]
\processtable{UIUC: influence of the window's kernel size\label{tab:uiucs}}
{\begin{tabular*}{\columnwidth}{@{\extracolsep{\fill}}c|ccccc} \toprule
\diagbox{\textbf{Wavelet}}{\textbf{Size}} & \textbf{3} & \textbf{5} & \textbf{7} & \textbf{9} & \textbf{11} \\ \hline
Curvelet & 74.02  & 74.99  & 75.58  & 76.47  & 76.71   \\ \hline
EWTC1 	& 65.10  & 70.56  & 72.85  & 74.91  & 75.96   \\ \hline
EWTC2 	& 64.87  & 69.45  & 71.65  & 74.35  & 75.28  \\ \hline
EWTC3 	& 72.70  & 75.61  & 76.47  & 77.06  & 77.77  \\ \hline
EWT2DT 	& 52.67  & 59.42  & 63.19  & 65.93  & 67.94  \\ \hline
Gabor 	& 64.28  & 65.30  & 66.82  & 68.29  & 69.87  \\ \hline
\diagbox{\textbf{Wavelet}}{\textbf{Size}} & \textbf{13} & \textbf{15}& \textbf{17}& \textbf{19} & \textbf{21}  \\ \hline
Curvelet & 76.81  & 77.04  & 76.96 & 77.01  & \textbf{77.59} \\ \hline
EWTC1 	& 76.93  & 77.58  & 77.84 & 78.17  & 78.40 \\ \hline
EWTC2 	& 76.36  & 77.25  & 77.57 & 77.40  & 77.82 \\ \hline
EWTC3 	& 77.60  & 77.93  & 78.04 & 78.11  & 78.00 \\ \hline
EWT2DT 	& 69.47 & 70.28  & 71.17 & 71.65  & 72.09 \\ \hline
Gabor 	& 71.20  & 72.03  & 72.89 & 73.09  & 73.48 \\ \hline
\diagbox{\textbf{Wavelet}}{\textbf{Size}} & \textbf{23}& \textbf{25} & \textbf{27}& \textbf{29}& \textbf{31} \\ \hline
Curvelet & 77.23  & 77.52  & 77.35  & 77.55  & 77.46 \\ \hline
EWTC1 & \textbf{78.54} & 78.20  & 78.11  & 78.33  & 78.21 \\ \hline
EWTC2 & 77.76  & 77.89  & 77.97  & \textbf{78.03} & 77.79 \\ \hline
EWTC3 & 77.97  & \textbf{78.23} & 77.86  & 77.91  & 77.52 \\ \hline
EWT2DT & 72.31  & 72.54  & \textbf{72.82} & 72.68  & 72.56 \\ \hline
Gabor & 73.78  & \textbf{74.11} & 74.05  & 73.95  & 73.97 \\ \hline
\end{tabular*}}{}
\end{table}

\begin{table}[!ht]
\processtable{Brodatz: influence of the window's kernel size\label{tab:brodatzs}}
{\begin{tabular*}{\columnwidth}{@{\extracolsep{\fill}}c|ccccc} \toprule
\diagbox{\textbf{Wavelet}}{\textbf{Size}} & \textbf{3} & \textbf{5} & \textbf{7} & \textbf{9} & \textbf{11} \\ \hline
Curvelet & 68.88  & 70.48  & 71.65  & 72.15  & 72.41   \\ \hline
EWTC1 	& 66.79  & 70.88  & 73.05  & 75.16  & 76.33   \\ \hline
EWTC2 	& 67.10  & 70.83  & 73.56  & 75.22  & 76.42  \\ \hline
EWTC3 	& 69.57  & 73.34  & 73.70  & 75.11  & 75.33  \\ \hline
EWT2DT 	& 68.97  & 73.54  & 75.76  & 76.98  & 77.50  \\ \hline
Gabor 	& 68.80  & 70.03  & 71.63  & 73.13  & 74.07  \\ \hline
\diagbox{\textbf{Wavelet}}{\textbf{Size}} & \textbf{13} & \textbf{15}& \textbf{17}& \textbf{19} & \textbf{21}  \\ \hline
Curvelet & 72.94  & 72.88  & 73.93 &  74.32  & 74.67 \\ \hline
EWTC1 	& 76.28  & 75.96  & 76.45 & 77.14  & 76.98 \\ \hline
EWTC2 	& 76.91  & 76.37  & 77.25 & 76.93  & 77.89 \\ \hline
EWTC3 	& 74.37  & 73.31  & 75.81 &  \textbf{76.32} & 76.27 \\ \hline
EWT2DT 	& 78.21  & 78.52  & 78.13 & 78.47  & \textbf{78.83} \\ \hline
Gabor 	& 75.14  & 76.44  & 77.41 & 77.82  & 77.91 \\ \hline
\diagbox{\textbf{Wavelet}}{\textbf{Size}} & \textbf{23}& \textbf{25} & \textbf{27}& \textbf{29}& \textbf{31} \\ \hline
Curvelet & 75.10  & \textbf{75.39} & 75.29  & 75.38  & 75.40 \\ \hline
EWTC1 & 77.58  & \textbf{77.72} & 77.72  & 77.57  & 76.58 \\ \hline
EWTC2 & 77.66  & \textbf{77.95} & 77.63  & 77.15  & 76.24 \\ \hline
EWTC3 & 75.81  & 75.71  & 75.53  & 75.43  & 75.08 \\ \hline
EWT2DT & 78.58  & 78.41  & 78.24  & 77.60  & 77.12 \\ \hline
Gabor & 78.62  & 78.34  & 78.67  & \textbf{78.79} & 78.51 \\ \hline
\end{tabular*}}{}
\end{table}

\begin{table}[!ht]
\processtable{Outex: influence of the window's kernel size\label{tab:outexs}}
{\begin{tabular*}{\columnwidth}{@{\extracolsep{\fill}}c|ccccc} \toprule
\diagbox{\textbf{Wavelet}}{\textbf{Size}} & \textbf{3} & \textbf{5} & \textbf{7} & \textbf{9} & \textbf{11} \\ \hline
Curvelet & 81.63  & 82.58  & 82.78  & 83.02  & 83.51   \\ \hline
EWTC1 	& 76.84  & 80.12  & 81.69  & 82.03  & 83.44   \\ \hline
EWTC2 	& 77.31  & 80.02  & 82.01  & 82.70  & 83.48  \\ \hline
EWTC3 	& 77.37  & 79.62  & 80.62  & 80.88  & 81.53  \\ \hline
EWT2DT 	& 73.55  & 78.27  & 80.08  & 81.68  & 82.51  \\ \hline
Gabor 	& 77.70  & 78.75  & 80.36  & 81.49  & 82.77  \\ \hline
\diagbox{\textbf{Wavelet}}{\textbf{Size}} & \textbf{13} & \textbf{15}& \textbf{17}& \textbf{19} & \textbf{21}  \\ \hline
Curvelet & 83.37  & 83.57  & 83.95 & \textbf{85.02} & 84.00 \\ \hline
EWTC1 	& 83.92  & 84.06  & 86.30 & \textbf{87.24} & 86.61 \\ \hline
EWTC2 	& 83.97  & 84.34  & 86.13 & \textbf{86.98} & 86.34 \\ \hline
EWTC3 	& 82.47  & 82.83  & 83.38 & \textbf{83.66} & 83.30 \\ \hline
EWT2DT 	& 83.12  & 83.68  & \textbf{84.03} & 83.98  & 83.90 \\ \hline
Gabor 	& 83.60  & 84.44  & 84.31 & \textbf{85.58} & 84.87 \\ \hline
\diagbox{\textbf{Wavelet}}{\textbf{Size}} & \textbf{23}& \textbf{25} & \textbf{27}& \textbf{29}& \textbf{31} \\ \hline
Curvelet & 83.67  & 83.76  & 84.01  & 83.49  & 83.30 \\ \hline
EWTC1 & 84.33  & 84.51  & 83.92  & 83.04  & 83.11 \\ \hline
EWTC2 & 84.34  & 84.18  & 83.93  & 83.58  & 83.35 \\ \hline
EWTC3 & 83.39  & 82.84  & 82.80  & 82.80  & 82.09 \\ \hline
EWT2DT & 83.43  & 83.54  & 82.87  & 82.88  & 82.39 \\ \hline
Gabor & 84.76  & 85.02  & 84.76  & 84.80  & 84.68 \\ \hline
\end{tabular*}}{}
\end{table}

\subsubsection{Influence of the post-processing}
Next, we want to study the influence of the post-processing option: energy, entropy or LBP (see Section~\ref{featproc}). Using the appropriate kernel sizes  found in 
Section~\ref{winsize} and choosing 35 for the LBP (since LBPs are very long to compute, we run some computations on a smaller subset of images to find that this value gives the 
best results), we ran the segmentation algorithm with the \textit{k-means} clustering (the \textit{cityblock} distance was used) for the wavelets subset defined before and computed 
the average metric for each case. Table~\ref{tab:FA} presents the results obtained for the ALOT, UIUC, Brodatz and Outex datasets, 
respectively. We easily observe that the energy option provides the best results for all types of wavelet for all datasets. Therefore, in the next sections, we 
will only use the energy option in the extraction of feature vectors.

\begin{table}[!h]
\processtable{Influence of the post-processing\label{tab:FA}}
{\begin{tabular*}{\columnwidth}{@{\extracolsep{\fill}}c|ccc} \toprule
\diagbox{\textbf{Wavelet}}{\textbf{Post Proc.}} & \textbf{Entropy} & \textbf{Energy} & \textbf{LBP} \\ \hline
\multicolumn{4}{c}{ALOT Dataset} \\ \hline
Curvelet   &  74.43 & \textbf{75.45} & 48.74 \\ \hline
EWTC1    &  72.13 & \textbf{78.68}  & 76.84  \\ \hline
EWTC2   &   70.13  & \textbf{78.70}  & 76.06 \\ \hline
EWTC3   &   69.27  & \textbf{72.99}  & 71.98 \\ \midrule
\multicolumn{4}{c}{UIUC Dataset} \\ \hline
Curvelet   &  75.97 & \textbf{77.52} & 43.17 \\ \hline
EWTC1    &  75.71 & \textbf{78.20}  & 70.10  \\ \hline
EWTC2   &   75.75  & \textbf{77.89}  & 70.45 \\ \hline
EWTC3   &   72.12  & \textbf{78.23}  & 71.09 \\ \midrule
\multicolumn{4}{c}{Brodatz Dataset} \\ \hline
Curvelet   &  74.94 & \textbf{75.39} & 37.01 \\ \hline
EWTC1    &  74.52 & \textbf{77.72}  & 72.46  \\ \hline
EWTC2   &   74.51  & \textbf{77.95}  & 73.78 \\ \hline
EWTC3   &   69.93  & \textbf{75.71}  & 69.97 \\ \midrule
\multicolumn{4}{c}{Outex Dataset} \\ \hline
Curvelet   &  83.72 & \textbf{85.02} &56.69 \\ \hline
EWTC1    &  78.92 & \textbf{87.24}  & 82.38  \\ \hline
EWTC2   &   79.05  & \textbf{86.98}  & 81.74 \\ \hline
EWTC3   &   77.07  & \textbf{83.66}  & 80.74 \\ \hline
\end{tabular*}}{}
\end{table}

\subsubsection{Influence of the clustering method} 
We now investigate the influence of the chosen clustering algorithm in the final step. Thus, we assessed the obtained segmentations, using the energy post-processing 
and 
previously found kernel sizes, for the different clustering methods. The results are summarized in Table~\ref{tab:Acluster} for the ALOT, UIUC, Brodatz and Outex datasets, 
respectively. We observe that, except for the ALOT dataset where the \textit{Nystr\"om} clustering combined with the \textit{cityblock} distance gives the best results, the 
\textit{k-means} approach with the \textit{cityblock} distance provides the best performances among all options whatever the type of wavelets. This could be explained 
by the fact the \textit{cityblock} distance is based on the $L^1$ norm which is less sensitive to outliers than other distances. Therefore, we will restrict our last 
experiments by using only this clustering option.
\begin{table}[!h]
\processtable{Influence of the clustering method\label{tab:Acluster}}
{\begin{tabular*}{\columnwidth}{@{\extracolsep{\fill}}c|cccc} \toprule
\diagbox{\textbf{Clust.}}{\textbf{Wavelet}} & \textbf{Curvelet} & \textbf{EWTC1} & \textbf{EWTC2} & \textbf{EWTC3} \\ \hline
\multicolumn{5}{c}{ALOT Dataset} \\ \hline
kmeans\_euclidean & 71.57 & 71.98 & 71.90 & 67.89 \\ \hline
kmeans\_cityblock & \textbf{75.45}  & 78.68  & 78.70  & 72.99 \\ \hline
kmeans\_cosine & 71.09 & 70.32 & 70.49 & 67.94 \\ \hline
kmeans\_correlation & 72.66 & 69.87 & 70.16 & 67.98 \\ \hline
nystrom\_euclidean & 73.16 & 75.91 & 75.68 & 71.70 \\ \hline
nystrom\_cityblock & 68.69 & \textbf{79.29} & \textbf{79.09} & \textbf{74.61} \\ \hline
nystrom\_cosine & 67.91 & 65.90 & 66.94 & 63.53 \\ \hline
nystrom\_correlation & 69.11 & 66.11 & 66.85 & 63.82 \\ \midrule
\multicolumn{5}{c}{UIUC Dataset} \\ \hline
kmeans\_euclidean & 70.10  & 75.23  & 74.04  & 76.01 \\ \hline
kmeans\_cityblock & \textbf{77.52}  & \textbf{78.20}  & 77.89  & 78.23 \\ \hline
kmeans\_cosine & 74.71  & 77.56  & 77.45  & 78.04 \\ \hline
kmeans\_correlation & 75.01  & 77.68  & 77.19  & \textbf{78.71} \\ \hline
nystrom\_euclidean & 69.18  & 75.86  & 74.99  & 76.86 \\ \hline
nystrom\_cityblock & 64.99  & 77.92  & \textbf{78.16}  & 78.45 \\ \hline
nystrom\_cosine & 70.94  & 75.30  & 74.54  & 75.18 \\ \hline
nystrom\_correlation & 75.01  & 75.38  & 74.10  & 75.54 \\ \midrule
\multicolumn{5}{c}{Brodatz Dataset} \\ \hline
kmeans\_euclidean & 70.15  & 72.61  & 73.06  & 72.32 \\ \hline
kmeans\_cityblock & \textbf{75.39}  & \textbf{77.72}  & \textbf{77.95}  & \textbf{75.71} \\ \hline
kmeans\_cosine & 70.58  & 70.96  & 72.31  & 70.61 \\ \hline
kmeans\_correlation & 73.85  & 71.26  & 71.89  & 70.39 \\ \hline
nystrom\_euclidean & 68.69  & 72.98  & 73.49  & 72.33 \\ \hline
nystrom\_cityblock & 64.98  & 77.50  & 77.61  & 75.50 \\ \hline
nystrom\_cosine & 65.16  & 65.71  & 67.30  & 64.49 \\ \hline
nystrom\_correlation & 67.17  & 66.02  & 67.27  & 65.17 \\ \midrule
\multicolumn{5}{c}{Outex Dataset} \\ \hline
kmeans\_euclidean & 78.62 & 80.02 & 80.03 & 75.64 \\ \hline
kmeans\_cityblock & \textbf{85.02} & \textbf{87.24} & \textbf{86.98} & \textbf{83.66} \\ \hline
kmeans\_cosine   & 80.10 & 83.29 & 83.86 & 81.00  \\ \hline
kmeans\_correlation   &80.70 & 83.14 & 82.62 & 81.08 \\ \hline
nystrom\_euclidean   & 78.12 & 81.00 & 81.73 & 77.30   \\ \hline
nystrom\_cityblock   & 75.64 & 85.71 & 85.91 & 82.71 \\ \hline
nystrom\_cosine   & 76.23 & 77.74 & 77.59 & 76.05 \\ \hline
nystrom\_correlation   & 76.42 & 78.61 & 78.06 & 75.85 \\ \hline
\end{tabular*}}{}
\end{table}

\subsubsection{Influence of the type of wavelet} 
Finally, we evaluate the influence of the chosen wavelet using only the energy processing with the corresponding kernel sizes and the \textit{k-means} 
clustering with 
\textit{cityblock} distance. Table~\ref{tab:outex} shows the performances corresponding to each wavelet transforms for the Outex, Brodatz, ALOT and UIUC 
datasets, respectively. The metrics NVOI, SCC, SDHD, BGM, VD and BCE are averaged among the entire set of test images. Since in this paper we do not 
consider any specific application, we only provide the mean and standard deviation (within parenthesis) values computed among the six metrics. The first observation 
is that classic wavelet families like DWT, Undecimated and Packets give the worst performances. This is not surprising since these transforms do not take into account the presence 
of geometry which we know is an important information to characterize textures.\\
We can notice that, for the Outex dataset, the EWT2DC1 family gives the best performances, closely followed by EWT2DC2 and the classic Curvelets. In the case of the Brodatz 
dataset, the EWT2DT performs slightly better than Gabor, EWT2DC1 and EWT2DC2 (also note that Gabor wavelets performs better than the classic curvelets). Results from the 
EWT2DC1 for 
the ALOT dataset are slightly better than  EWT2DC2, followed by EWT2DT, Gabor and the classic curvelets. Finally, we observe that the EWT2DC1, followed by EWT2DC2 and the 
classic curvelets, give the best results for the UIUC dataset.

\begin{table}[!t]
\processtable{Datasets benchmark result\label{tab:outex}}
{\begin{tabular*}{\columnwidth}{@{\extracolsep{\fill}}c|cccc} \toprule
\textbf{Wavelet} & \textbf{Outex} & \textbf{Brodatz} & \textbf{ALOT} & \textbf{UIUC} \\ \hline
Curvelet   & 85.02(4.93)  & 80.11(7.83)  & 79.74(8.43)  & 78.58(7.18) \\
EWTC1      & \textbf{87.24(7.94)}  & 81.04(9.92)  & \textbf{81.51(9.94)}  & \textbf{79.32(9.81)}    \\
EWTC2      & 86.98(8.15)  & 81.09(9.88)  & 81.30(9.54)  & 78.64(9.82)    \\
EWTC3      & 83.66(11.31) & 76.63(14.31) & 74.97(13.41) & 74.36(15.81)   \\
EWTLP      & 61.55(11.19) & 65.00(13.64) & 73.01(12.41) & 55.71(13.00)   \\
EWT2DT     & 82.60(7.50)  & \textbf{84.01(10.20)} & 81.26(10.11) & 73.75(11.400   \\
Gabor      & 81.16(7.49)  & 82.23(10.58) & 80.30(10.53) & 73.69(9.80)    \\
Meyer\_2   & 72.68(7.08)  & 75.24(10.04) & 76.56(10.13) & 61.34(9.16)    \\
Meyer\_3   & 75.60(6.42)  & 81.33(8.73)  & 79.50(9.01)  & 71.53(9.29)    \\
Meyer\_4   & 75.97(7.17)  & 80.97(8.86)  & 80.48(8.37)  & 76.01(8.29)    \\ \midrule
\multicolumn{5}{c}{Discrete Wavelet Transform (DWT)} \\ \hline
Coif1\_2 & 68.70(7.60)    & 72.05(9.70)  & 73.19(9.61)  & 59.53(8.87)    \\
Coif1\_3 & 70.84(7.13)    & 78.31(7.94)  & 76.17(8.60)  & 68.37(8.48)    \\
Coif1\_4 & 68.87(7.68)    & 77.15(8.05)  & 75.85(8.26)  & 68.67(8.20)    \\
Coif2\_2 & 69.33(7.40)    & 72.48(9.73)  & 73.38(9.69)  & 59.42(8.98)    \\
Coif2\_3 & 71.54(7.05)    & 78.09(8.17)  & 76.24(8.78)  & 68.22(8.34)    \\
Coif2\_4 & 69.57(7.61)    & 77.50(7.82)  & 76.54(8.43)  & 69.29(8.07)    \\
Daub4\_2 & 68.80(7.56)    & 73.98(9.89)  & 74.51(9.74)  & 60.23(9.05)    \\
Daub4\_3 & 72.36(7.90)    & 80.48(8.74)  & 77.68(8.80)  & 71.76(9.60)    \\
Daub4\_4 & 72.52(7.68)    & 80.28(7.86)  & 77.35(8.73)  & 75.62(8.28)    \\
Daub6\_2 & 69.59(7.34)    & 73.77(10.00) & 74.54(9.89)  & 60.57(9.18)    \\
Daub6\_3 & 73.08(7.34)    & 80.96(8.41)  & 77.32(9.11)  & 71.80(9.26)    \\
Daub6\_4 & 73.57(7.14)    & 80.12(7.96)  & 77.41(8.50)  & 76.43(8.03)    \\
Sym4\_2  & 70.00(7.62)    & 73.33(9.89)  & 74.38(9.66)  & 60.33(9.08)    \\
Sym4\_3  & 72.72(7.09)    & 80.06(8.40)  & 77.53(8.83)  & 70.52(8.83)    \\
Sym4\_4  & 71.40(7.33)    & 79.53(7.88)  & 77.22(8.43)  & 72.81(8.08)    \\
Sym5\_2  & 69.60(7.12)    & 72.37(9.58)  & 73.63(9.49)  & 59.26(8.74)    \\
Sym5\_3  & 71.58(7.06)    & 77.80(8.06)  & 75.85(8.58)  & 67.77(8.14)    \\
Sym5\_4  & 69.39(7.99)    & 76.82(8.10)  & 76.03(8.43)  & 69.23(8.20)    \\ \midrule
\multicolumn{5}{c}{Undecimated Wavelet Transform} \\ \hline
Coif1\_2   & 63.82(7.78)    & 72.33(9.67)  & 72.66(10.30) & 58.37(8.48)    \\
Coif1\_3   & 60.88(9.31)    & 69.84(9.30)  & 74.00(9.87)  & 60.22(8.63)    \\
Coif1\_4   & 59.91(11.89)   & 68.75(9.38)  & 73.55(10.23) & 60.48(8.63)    \\
Coif2\_2   & 64.05(7.91)    & 71.92(9.71)  & 72.66(10.22) & 58.12(8.44)    \\
Coif2\_3   & 61.52(9.57)    & 69.64(9.31)  & 73.57(9.42)  & 59.70(8.32)    \\
Coif2\_4   & 60.22(12.04)   & 68.73(9.74)  & 72.05(9.95)  & 60.77(8.60)    \\
Daub4\_2   & 62.81(8.32)    & 73.50(10.20) & 74.02(9.95)  & 58.96(8.97)    \\
Daub4\_3   & 61.57(9.45)    & 72.12(9.63)  & 75.19(9.77)  & 64.38(9.36)    \\
Daub4\_4   & 63.11(9.89)    & 71.95(9.34)  & 75.29(8.95)  & 67.84(8.72)    \\
Daub6\_2   & 64.47(8.46)    & 73.79(10.18) & 74.67(9.93)  & 59.16(8.82)    \\
Daub6\_3   & 63.39(9.12)    & 72.43(9.76)  & 75.99(9.57)  & 64.20(9.01)    \\
Daub6\_4   & 64.34(9.88)    & 72.26(9.23)  & 75.39(9.15)  & 67.72(8.52)    \\
Sym4\_2    & 64.95(8.17)    & 72.87(10.00) & 73.97(9.84)  & 58.93(8.76)    \\
Sym4\_3    & 62.87(9.47)    & 71.12(9.50)  & 74.56(9.45)  & 61.77(8.69)    \\
Sym4\_4    & 61.83(11.55)   & 69.82(9.35)  & 73.49(9.53)  & 63.37(8.61)    \\
Sym5\_2    & 63.64(7.66)    & 71.58(9.74)  & 73.58(9.50)  & 58.06(8.61)    \\
Sym5\_3    & 61.51(9.57)    & 68.93(9.17)  & 73.21(9.31)  & 59.08(8.16)    \\
Sym5\_4    & 59.72(12.00)   & 68.47(9.55)  & 72.09(9.94)  & 60.26(8.53)    \\ \midrule
\multicolumn{5}{c}{Wavelet Packet Transform} \\ \hline
Coif1\_2 & 70.97(7.63)    & 74.50(10.22) & 69.36(9.89)  & 57.28(8.60)    \\
Coif1\_3 & 74.19(5.93)    & 75.21(9.18)  & 71.99(9.44)  & 62.75(8.11)    \\
Coif1\_4 & 68.47(6.52)    & 69.04(8.64)  & 66.12(7.94)  & 62.65(8.42)    \\
Coif2\_2 & 71.75(7.64)    & 74.19(10.11) & 69.50(9.96)  & 57.36(8.83)    \\
Coif2\_3 & 74.76(5.91)    & 75.35(9.09)  & 71.65(9.27)  & 63.05(8.28)    \\
Coif2\_4 & 66.39(7.08)    & 67.80(8.84)  & 64.83(7.99)  & 61.61(8.48)    \\
Daub4\_2 & 70.50(8.21)    & 74.98(10.46) & 69.64(10.61) & 57.12(8.75)    \\
Daub4\_3 & 78.23(7.21)    & 77.24(10.08) & 74.73(10.85) & 64.53(8.93)    \\
Daub4\_4 & 80.74(7.22)    & 80.63(9.57)  & 76.87(10.89) & 71.41(9.06)    \\
Daub6\_2 & 72.42(8.09)    & 74.76(10.83) & 70.44(10.25) & 57.67(8.87)    \\
Daub6\_3 & 79.57(6.89)    & 78.25(10.30) & 75.18(10.70) & 65.25(8.77)    \\
Daub6\_4 & 82.21(6.43)    & 81.16(9.74)  & 76.42(10.54) & 72.73(9.06)    \\
Sym4\_2  & 72.86(7.88)    & 75.09(10.51) & 70.66(10.08) & 57.90(8.92)    \\
Sym4\_3  & 77.67(6.02)    & 77.70(9.58)  & 73.88(9.86)  & 64.63(8.82)    \\
Sym4\_4  & 74.35(6.62)    & 74.26(8.51)  & 70.80(8.70)  & 66.81(8.42)    \\
Sym5\_2  & 71.66(7.27)    & 74.11(10.12) & 69.38(9.94)  & 57.24(8.51)    \\
Sym5\_3  & 73.55(5.95)    & 74.80(9.09)  & 70.59(8.90)  & 62.88(8.16)    \\
Sym5\_4  & 61.98(6.99)    & 65.56(8.65)  & 62.55(7.57)  & 60.81(8.41)    \\ \hline
\end{tabular*}}{}
\end{table}

\begin{figure}[!t]
\begin{center}
\begin{tabular}{ccc}
 & Outex & Brodatz\\
\rotatebox{90}{test image} &  \includegraphics[width=0.17\textwidth]{O26.pdf} &
\includegraphics[width=0.17\textwidth]{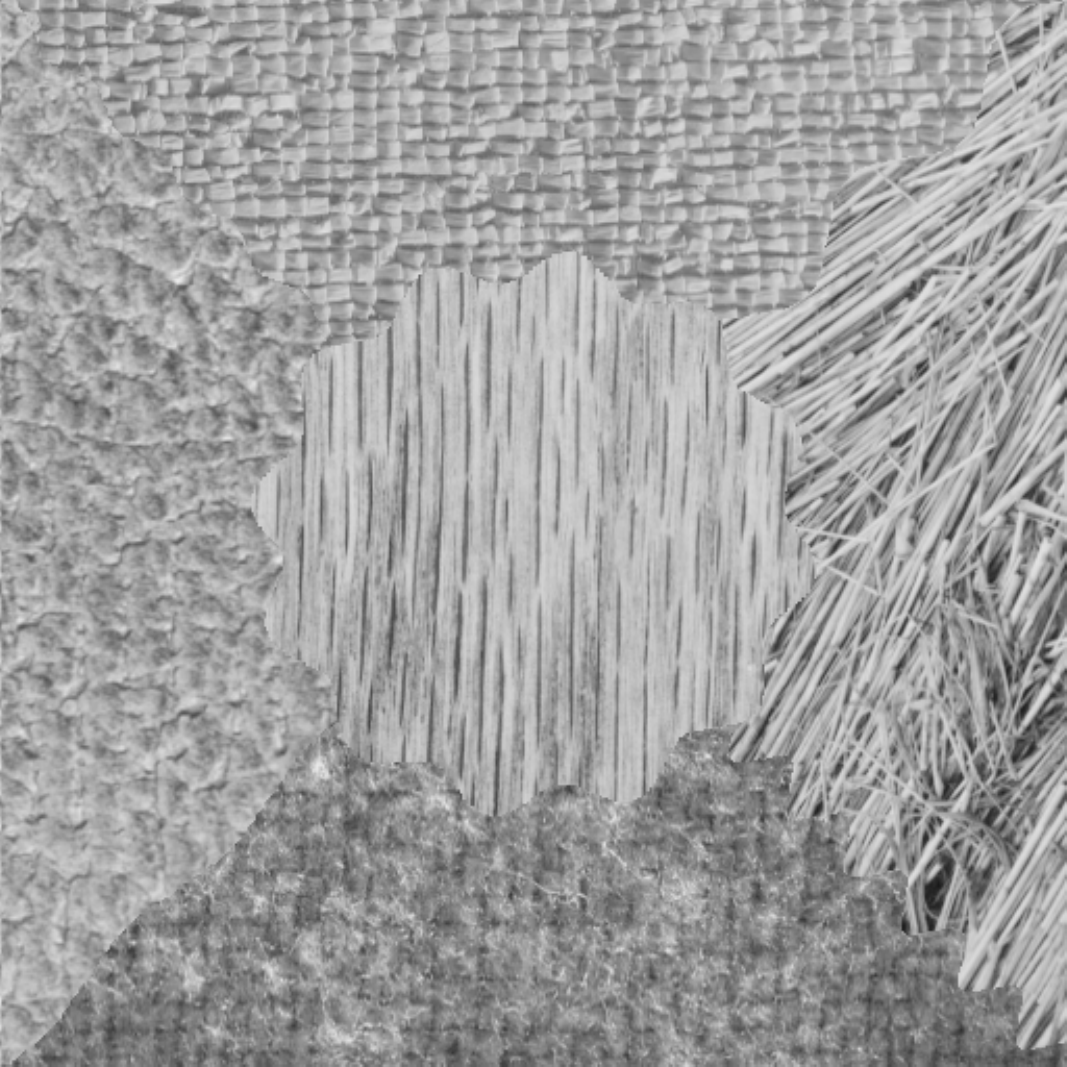} \\ 
\rotatebox{90}{Curvelet} &\includegraphics[width=0.17\textwidth]{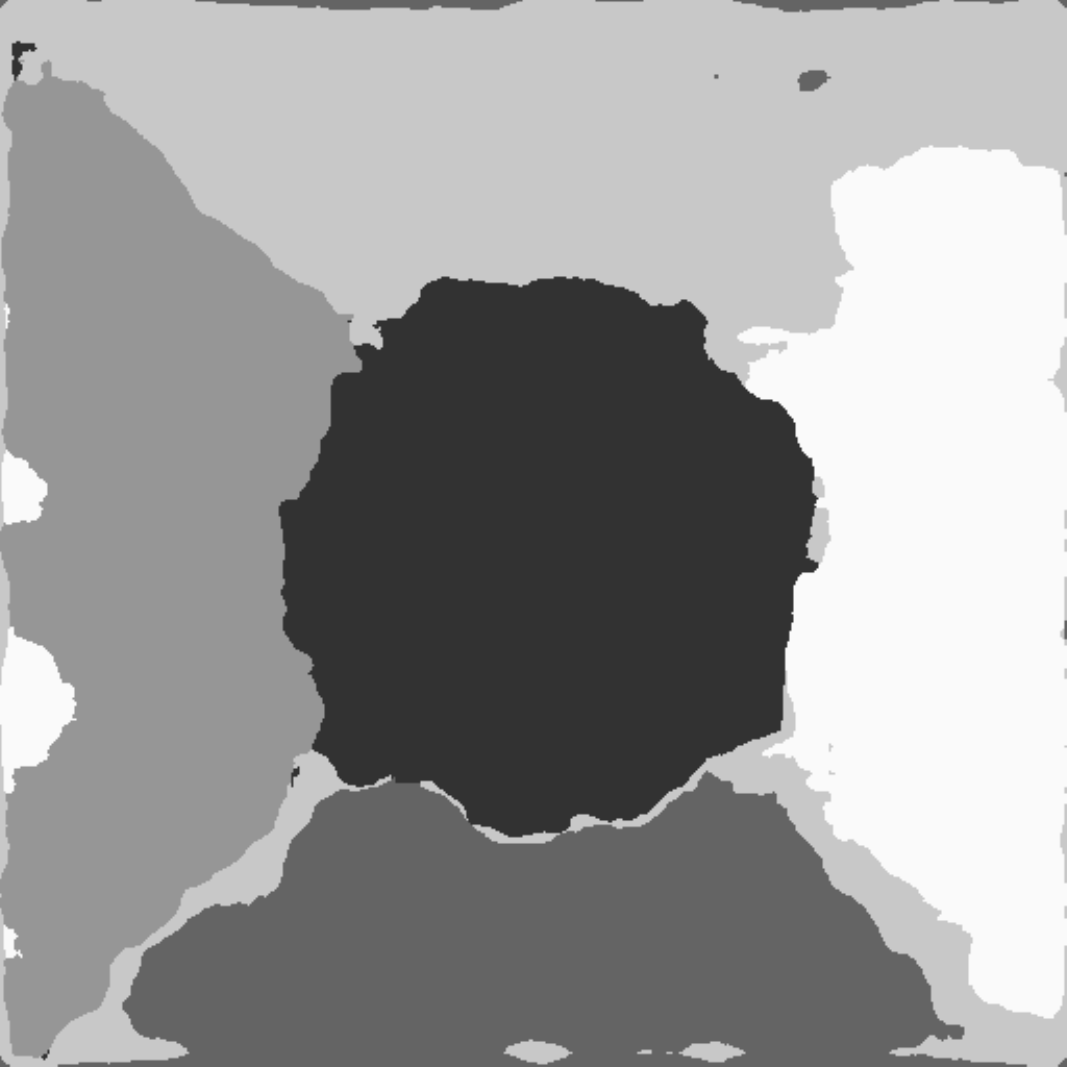} &
\includegraphics[width=0.17\textwidth]{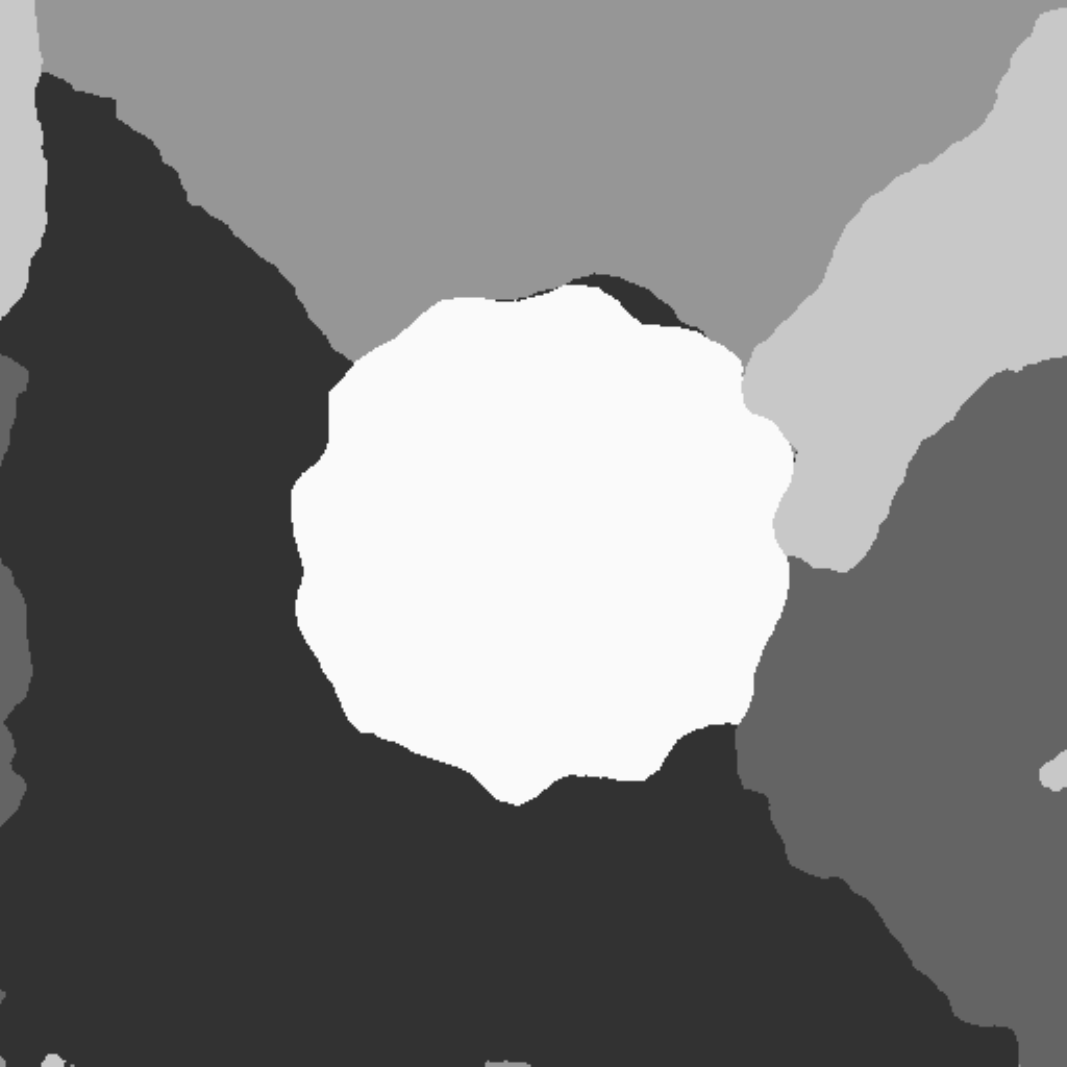} \\ 
\rotatebox{90}{EWTC1} &\includegraphics[width=0.17\textwidth]{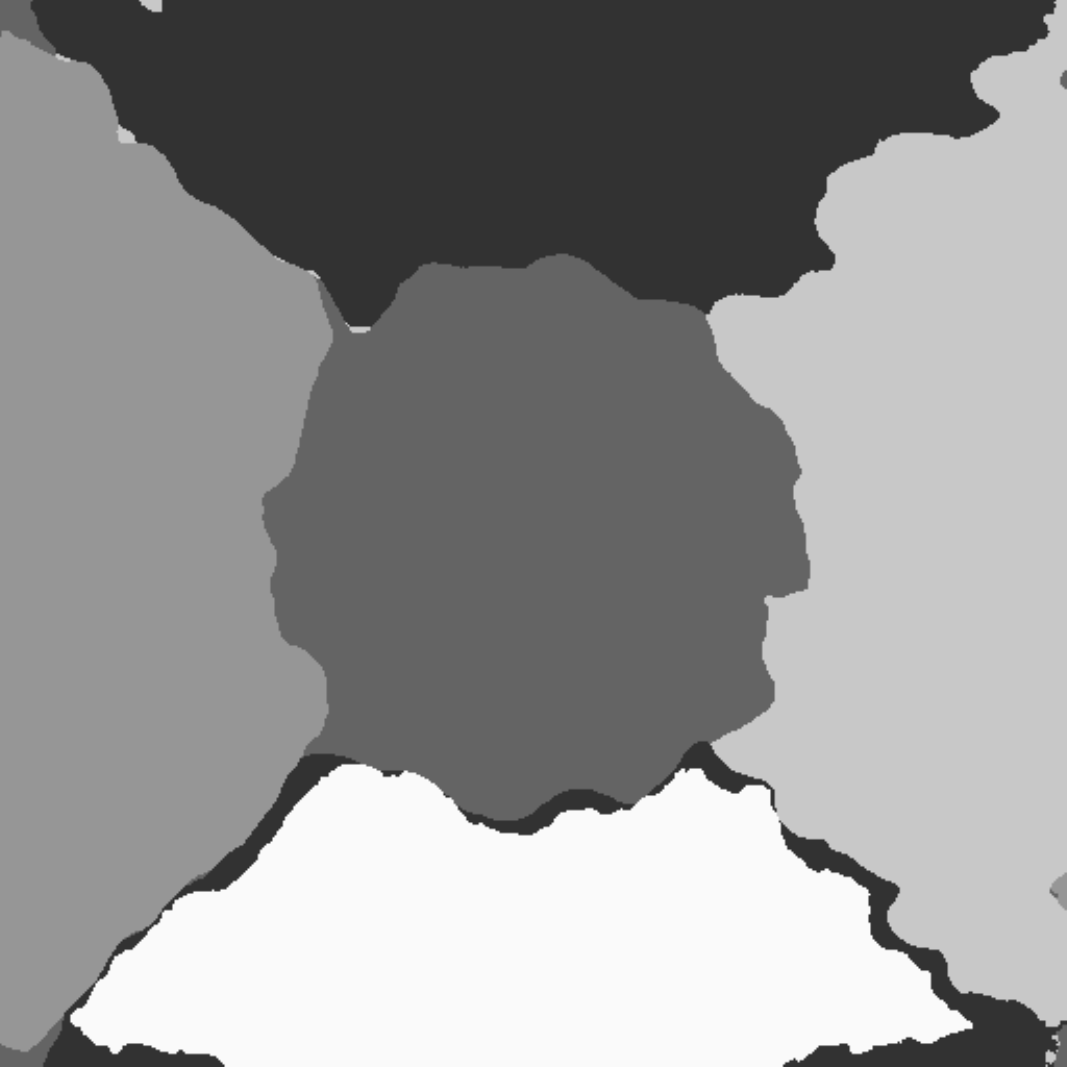} &
\includegraphics[width=0.17\textwidth]{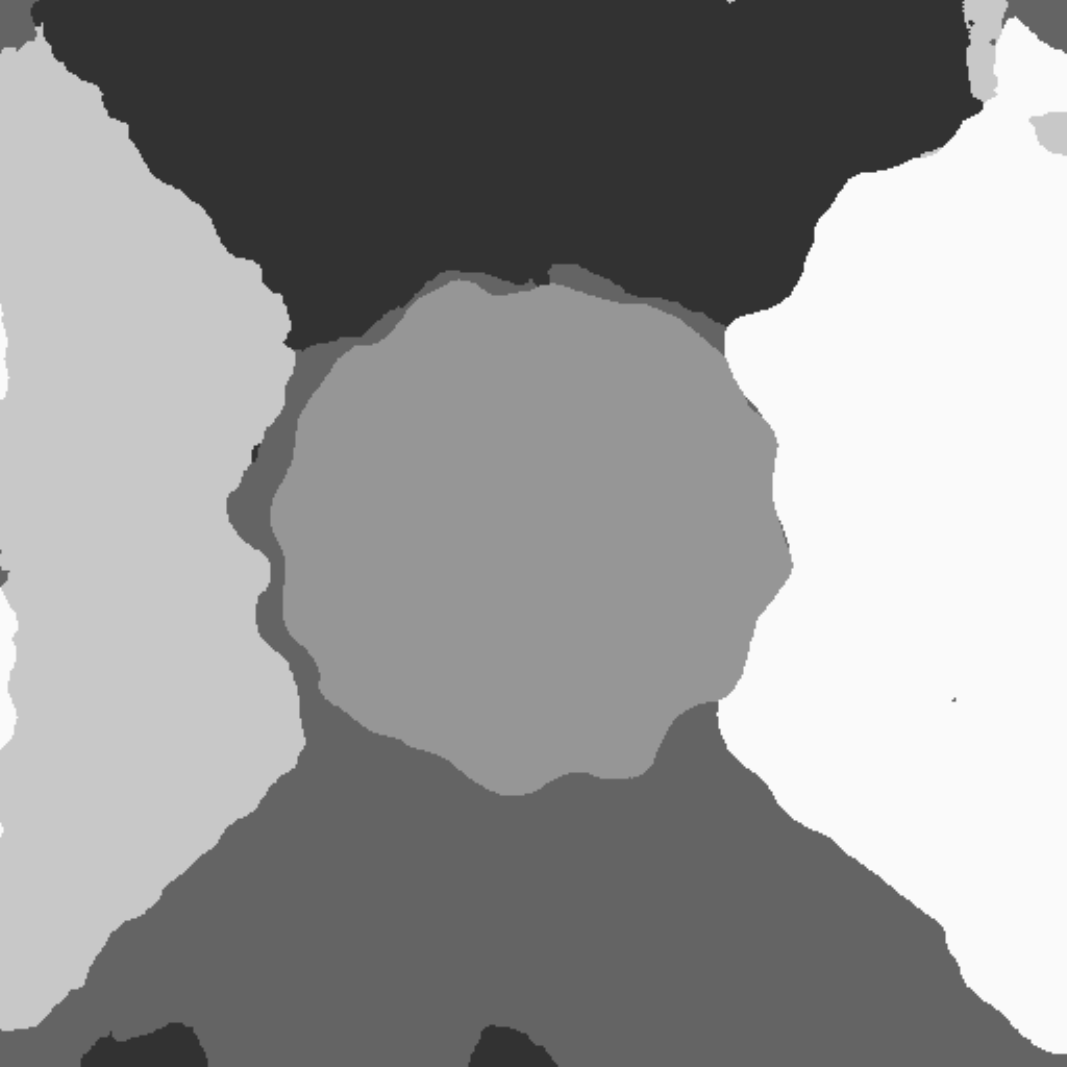} \\ 
\rotatebox{90}{EWTC2} &\includegraphics[width=0.17\textwidth]{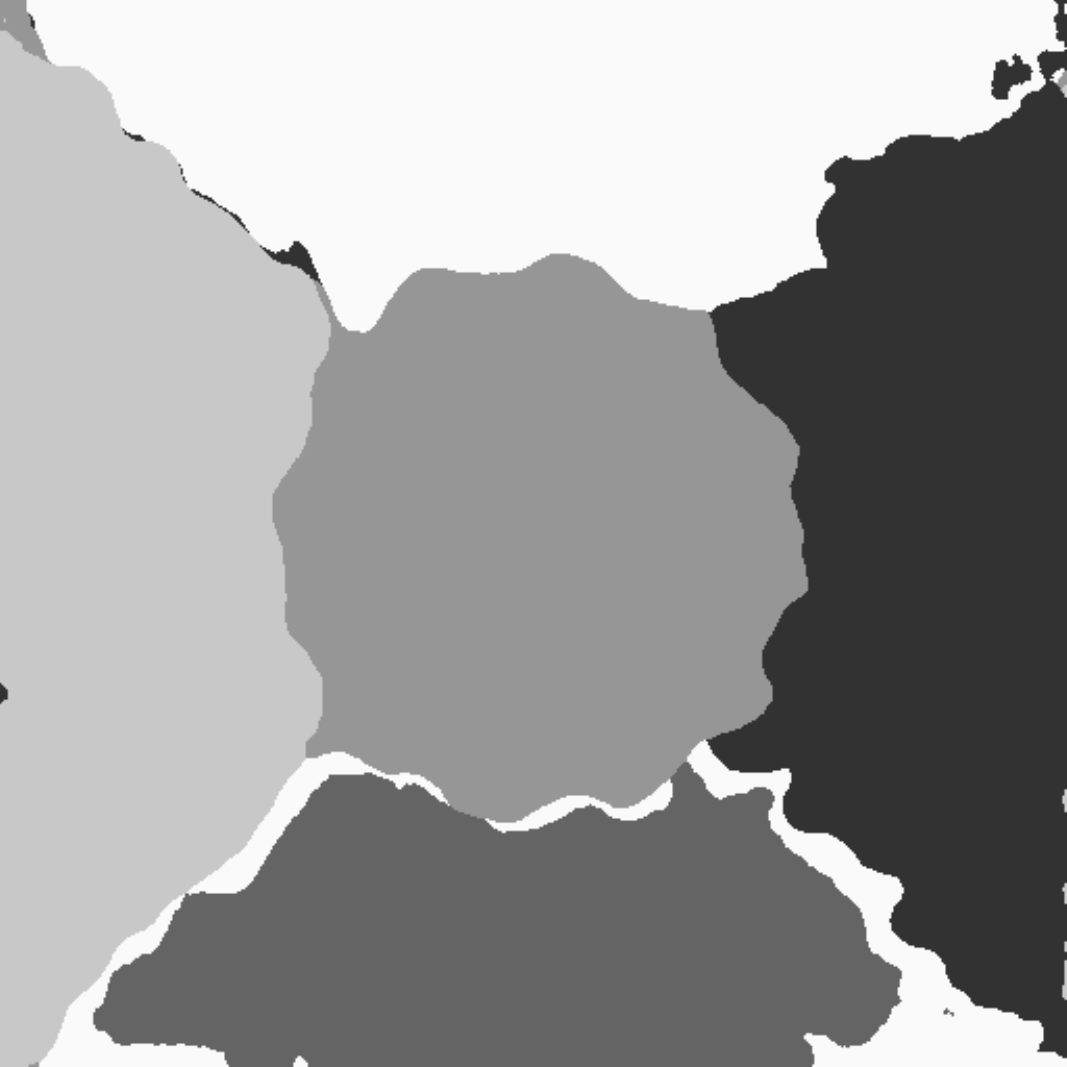} &
\includegraphics[width=0.17\textwidth]{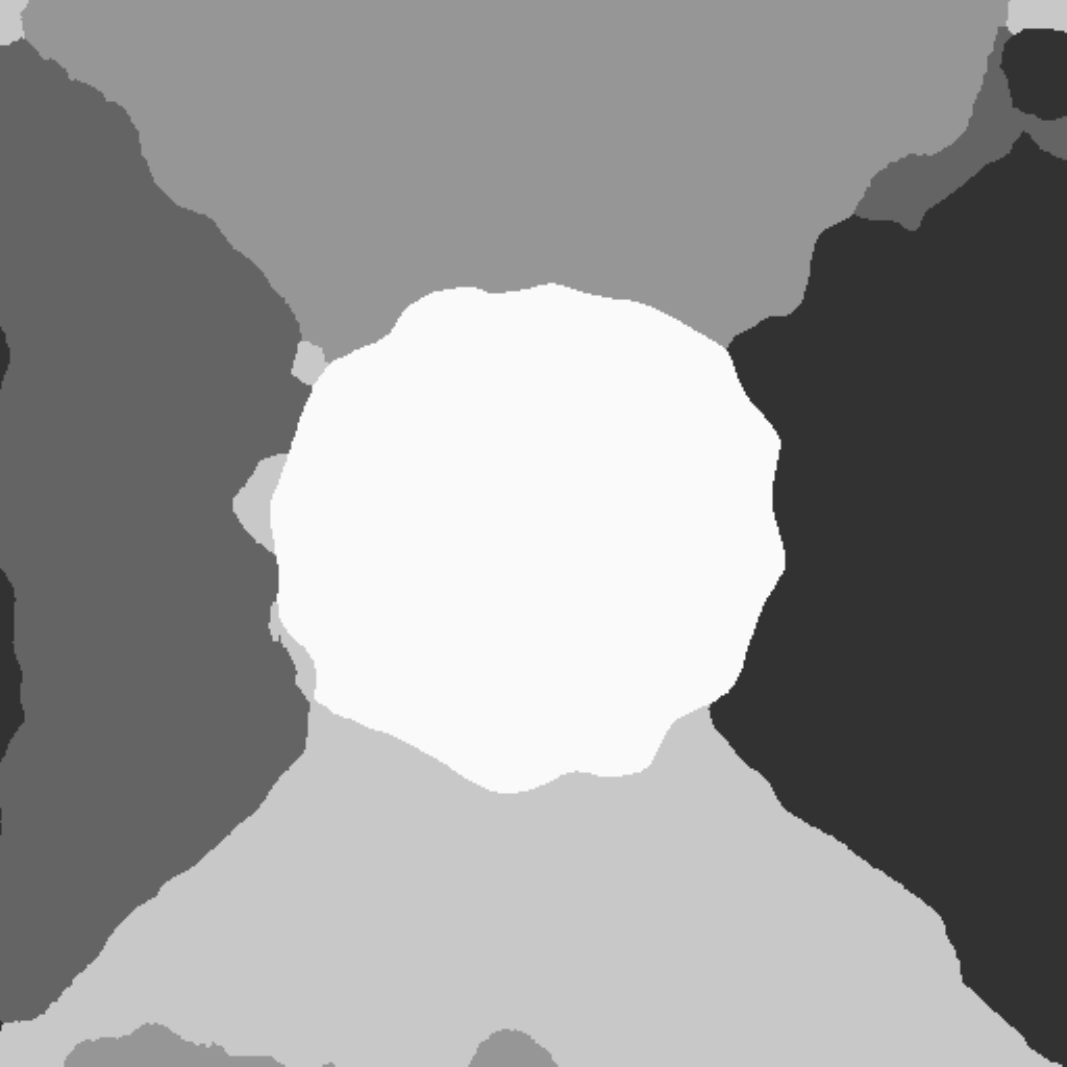} \\ 
\rotatebox{90}{EWTC3} &\includegraphics[width=0.17\textwidth]{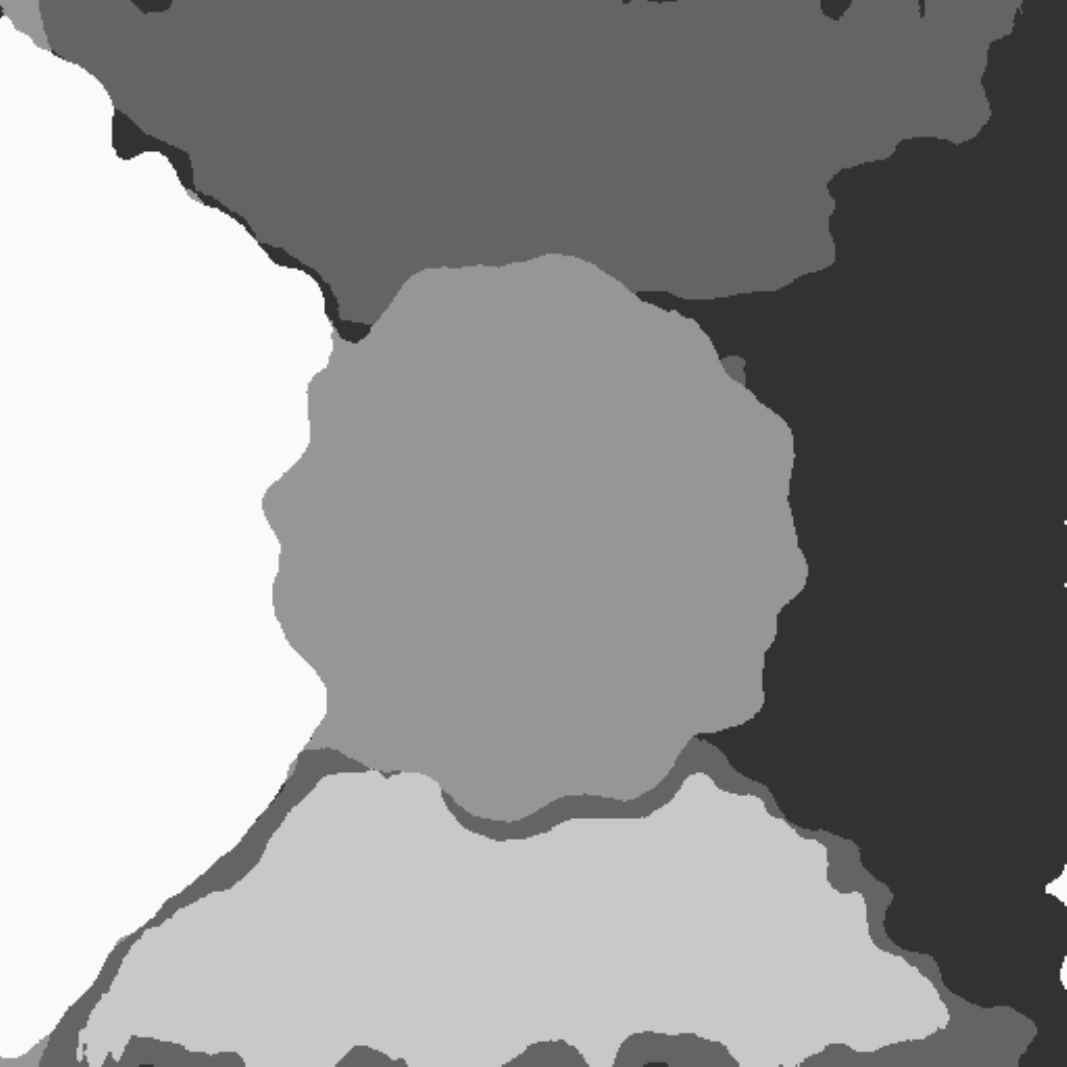} &
\includegraphics[width=0.17\textwidth]{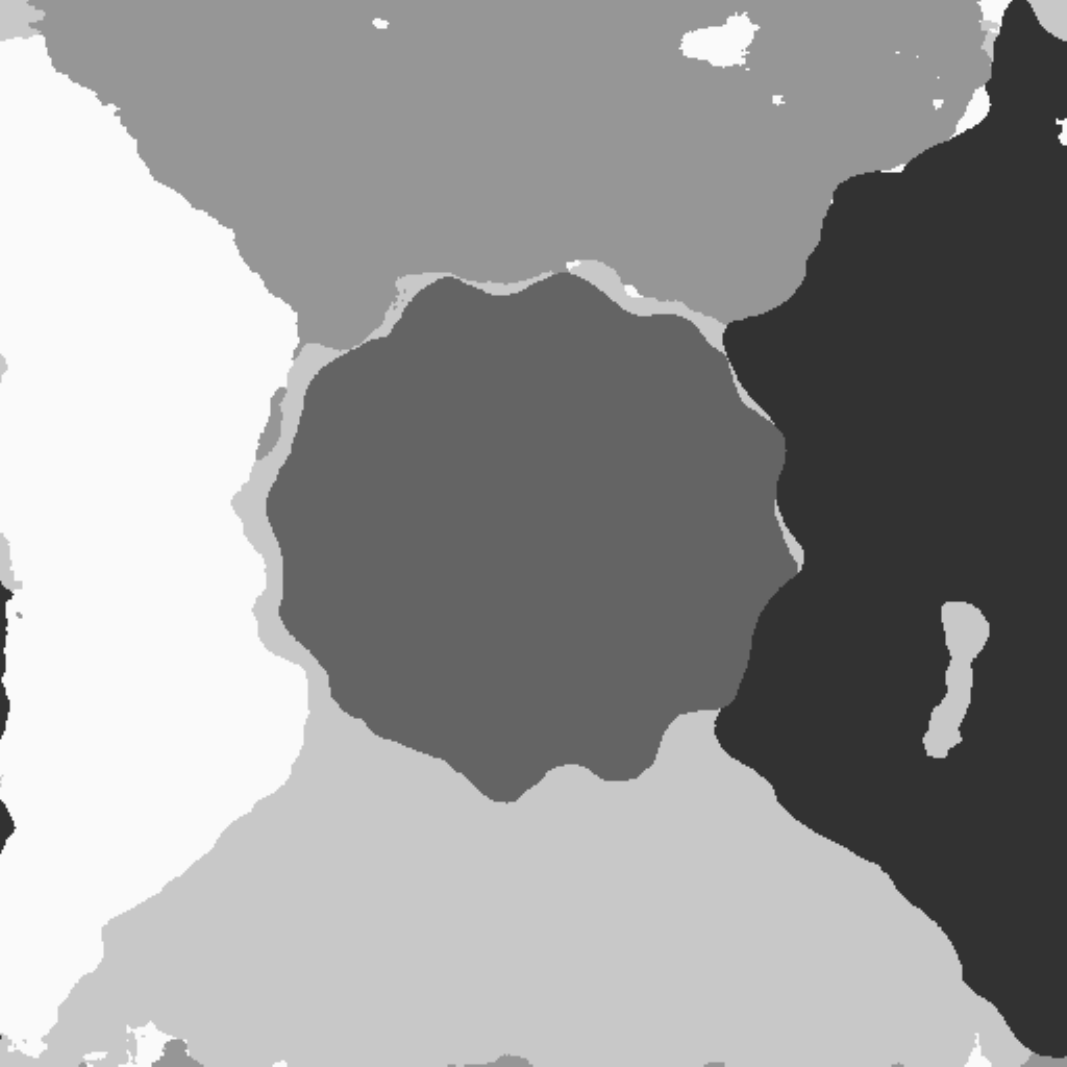} \\ 
\rotatebox{90}{EWT2DT} &\includegraphics[width=0.17\textwidth]{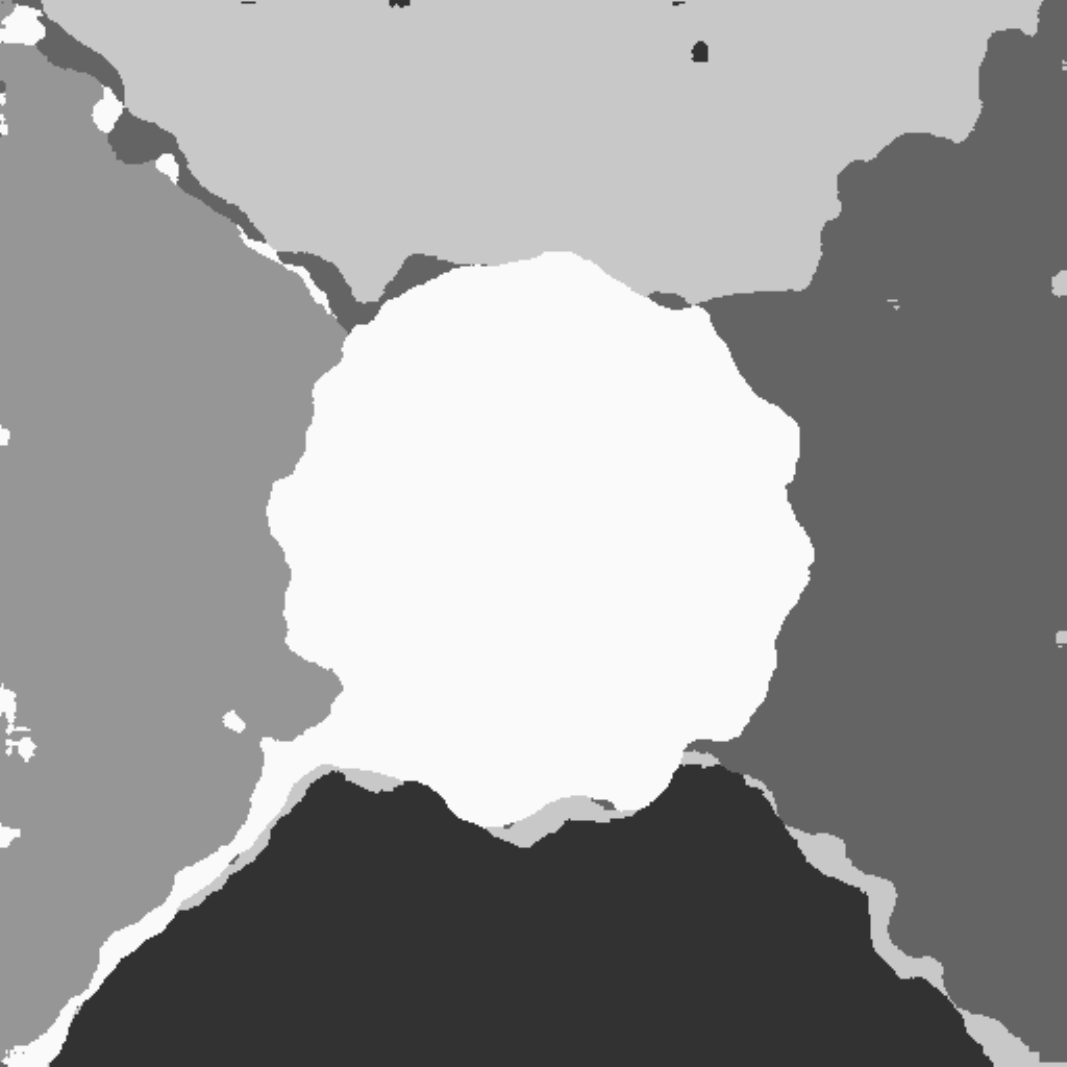} &
\includegraphics[width=0.17\textwidth]{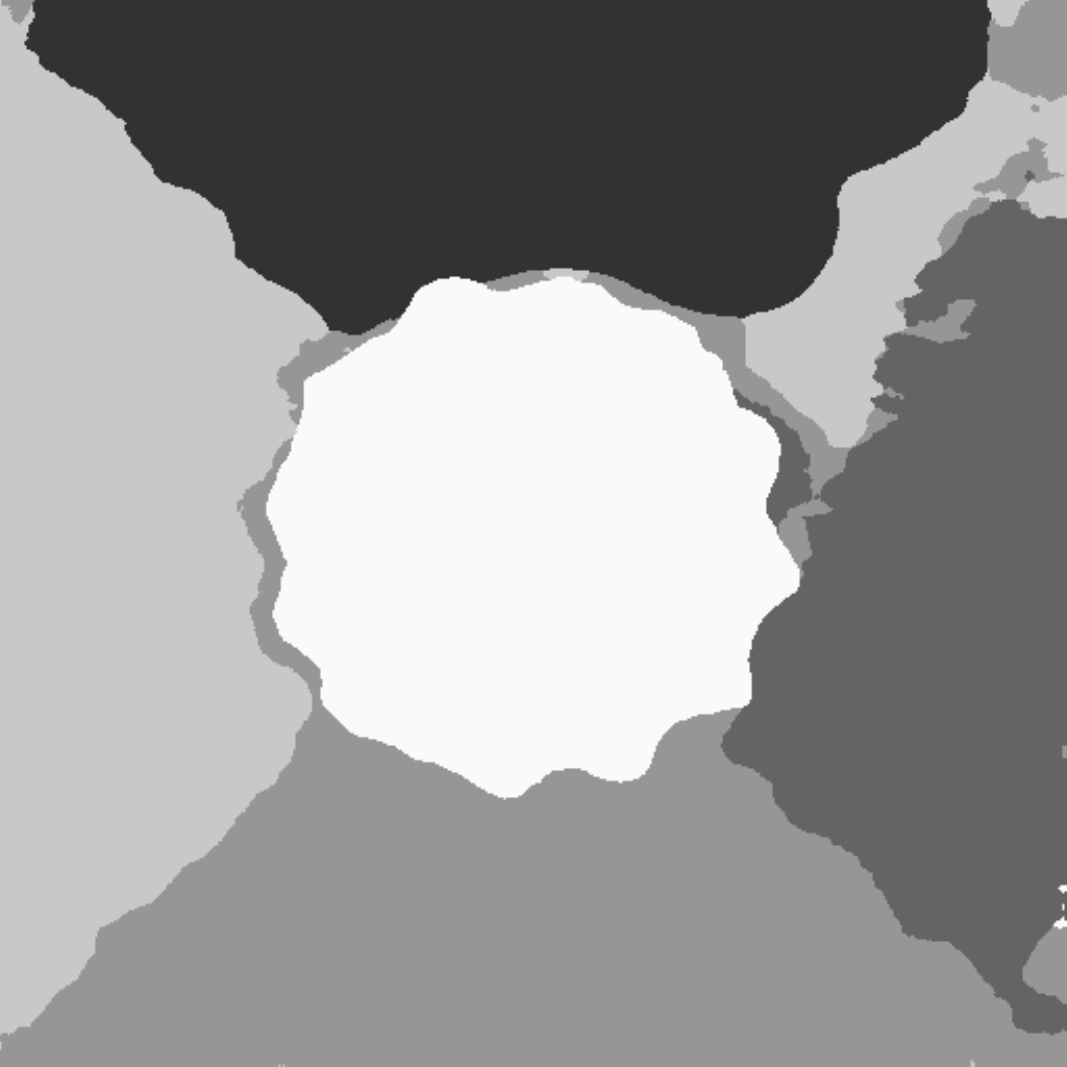} \\ 
\rotatebox{90}{Gabor} &\includegraphics[width=0.17\textwidth]{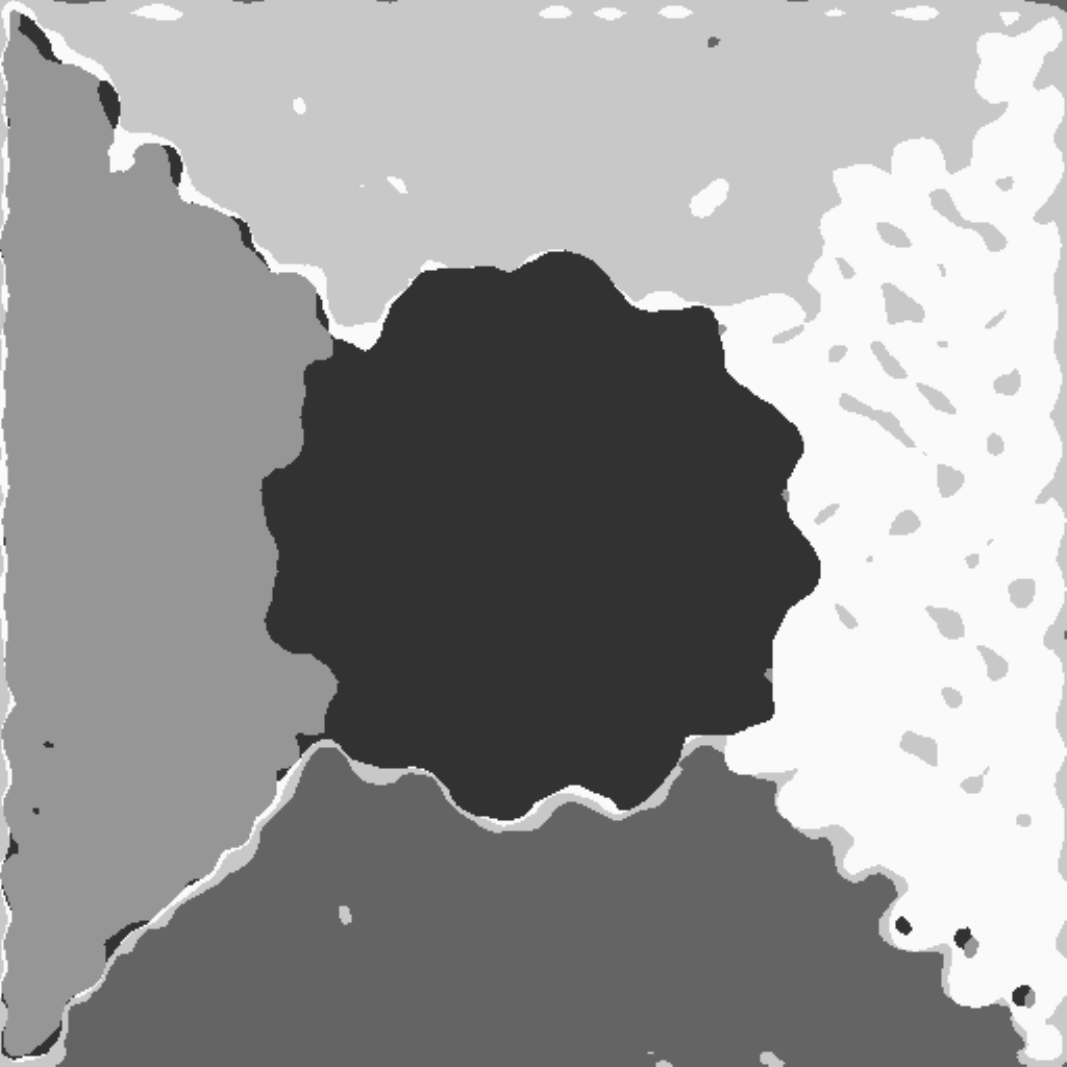} &
\includegraphics[width=0.17\textwidth]{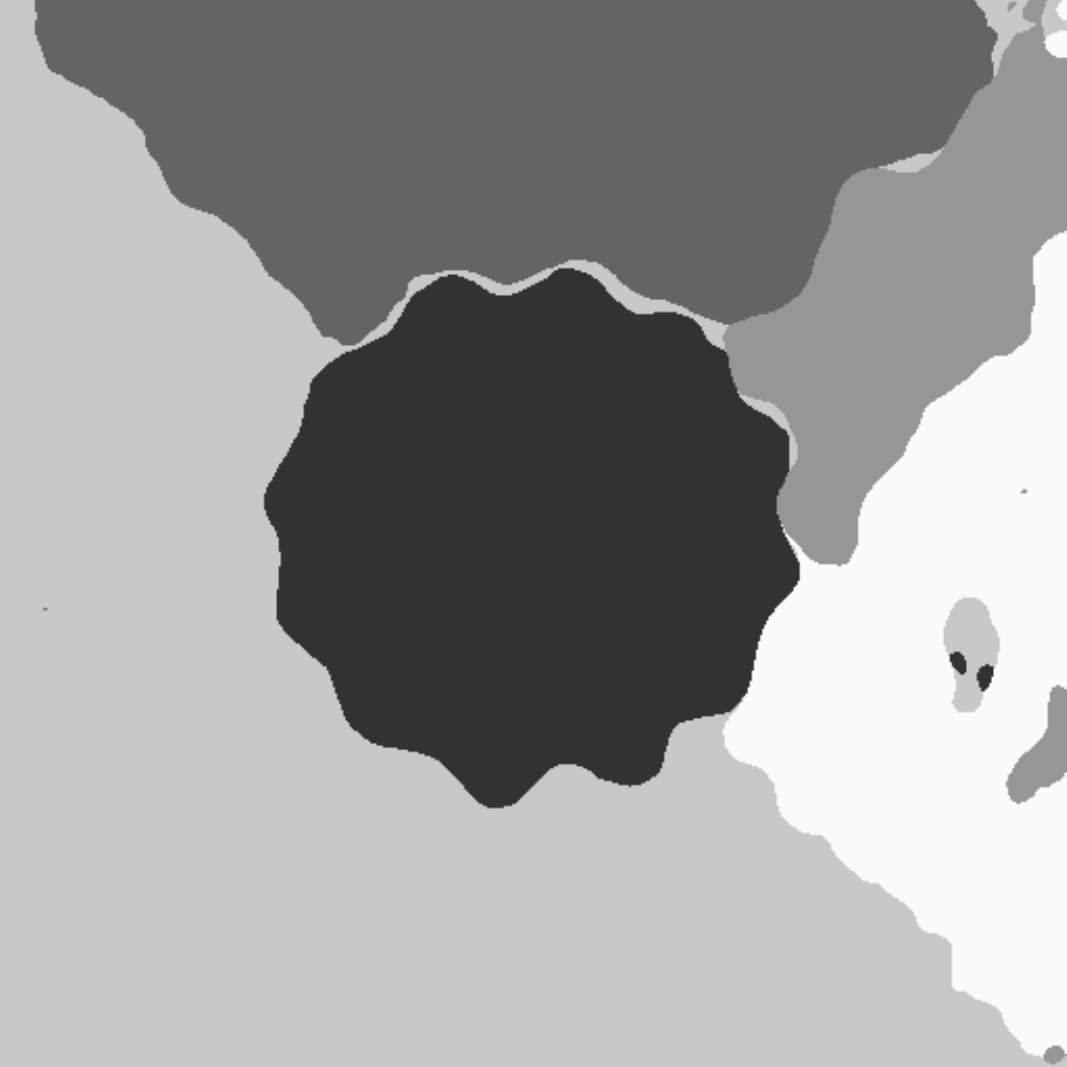} 
\end{tabular}
\end{center}
\caption{Visual comparisons of segmentation results for the best wavelet families on one image from the Outex and Brodatz datasets.}
\label{fig:visucompare}
\end{figure}

\begin{figure}[!ht]
\begin{center}
\begin{tabular}{ccc}
& ALOT & UIUC\\
\rotatebox{90}{test image} &  \includegraphics[width=0.17\textwidth]{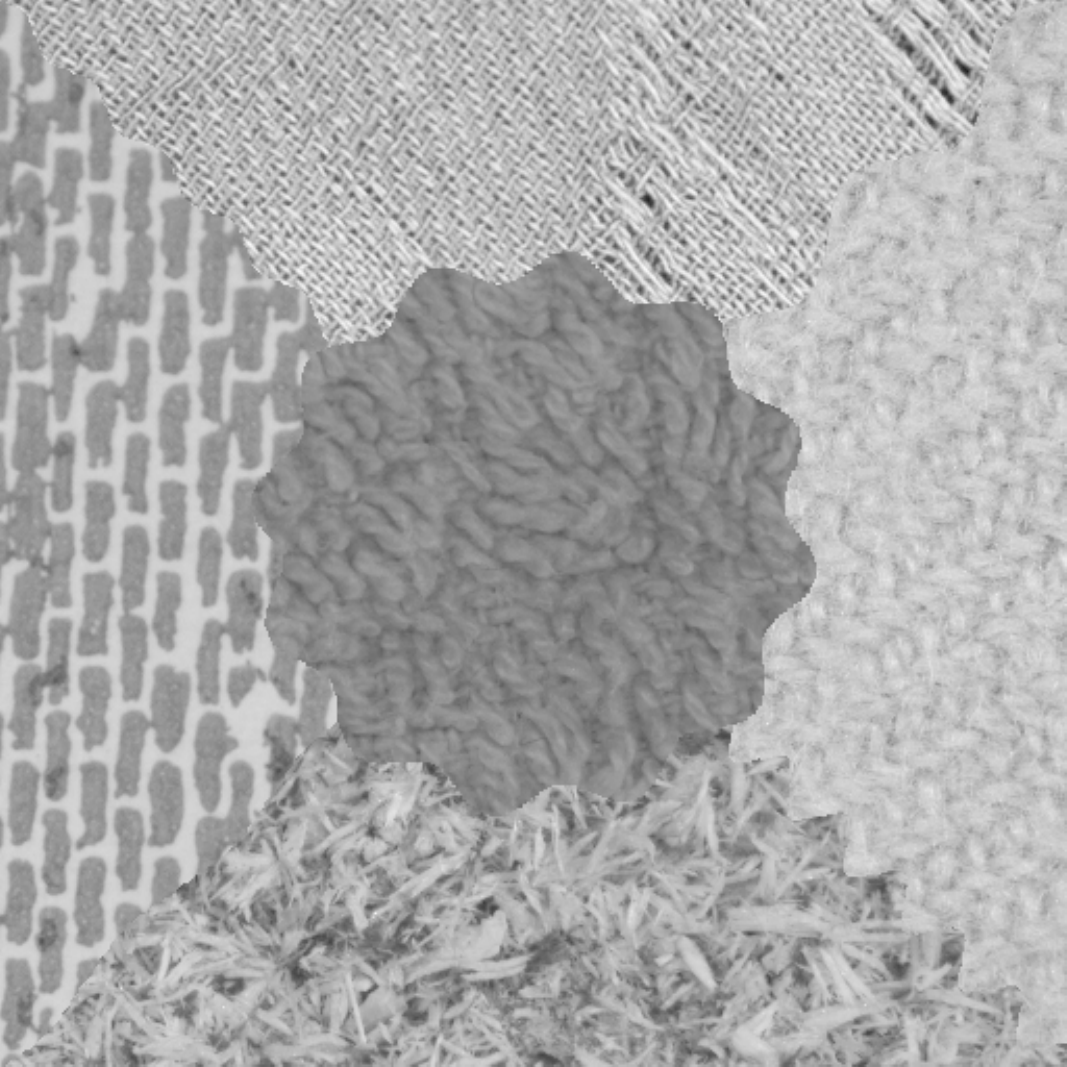} &
\includegraphics[width=0.17\textwidth]{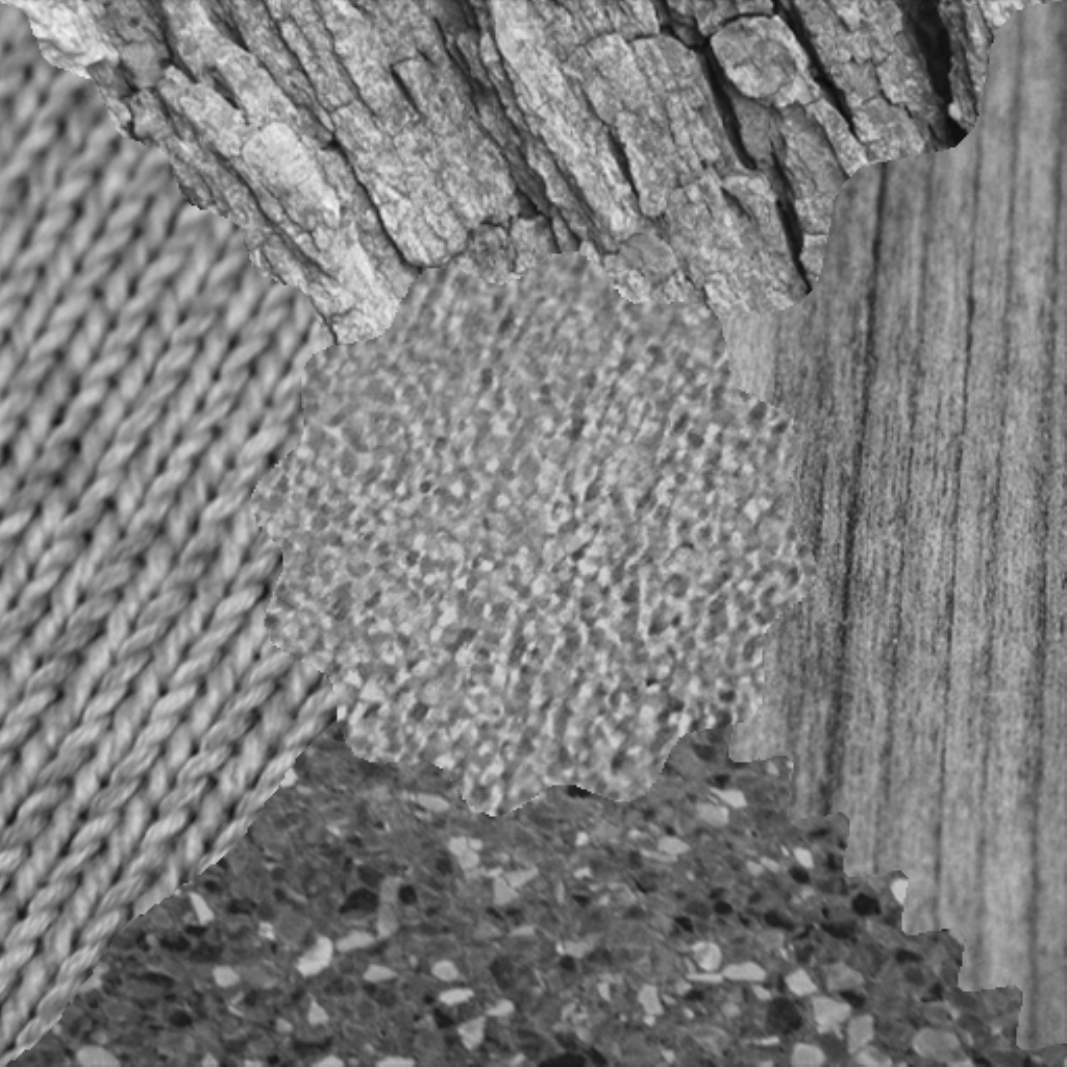}  \\ 
\rotatebox{90}{Curvelet} & \includegraphics[width=0.17\textwidth]{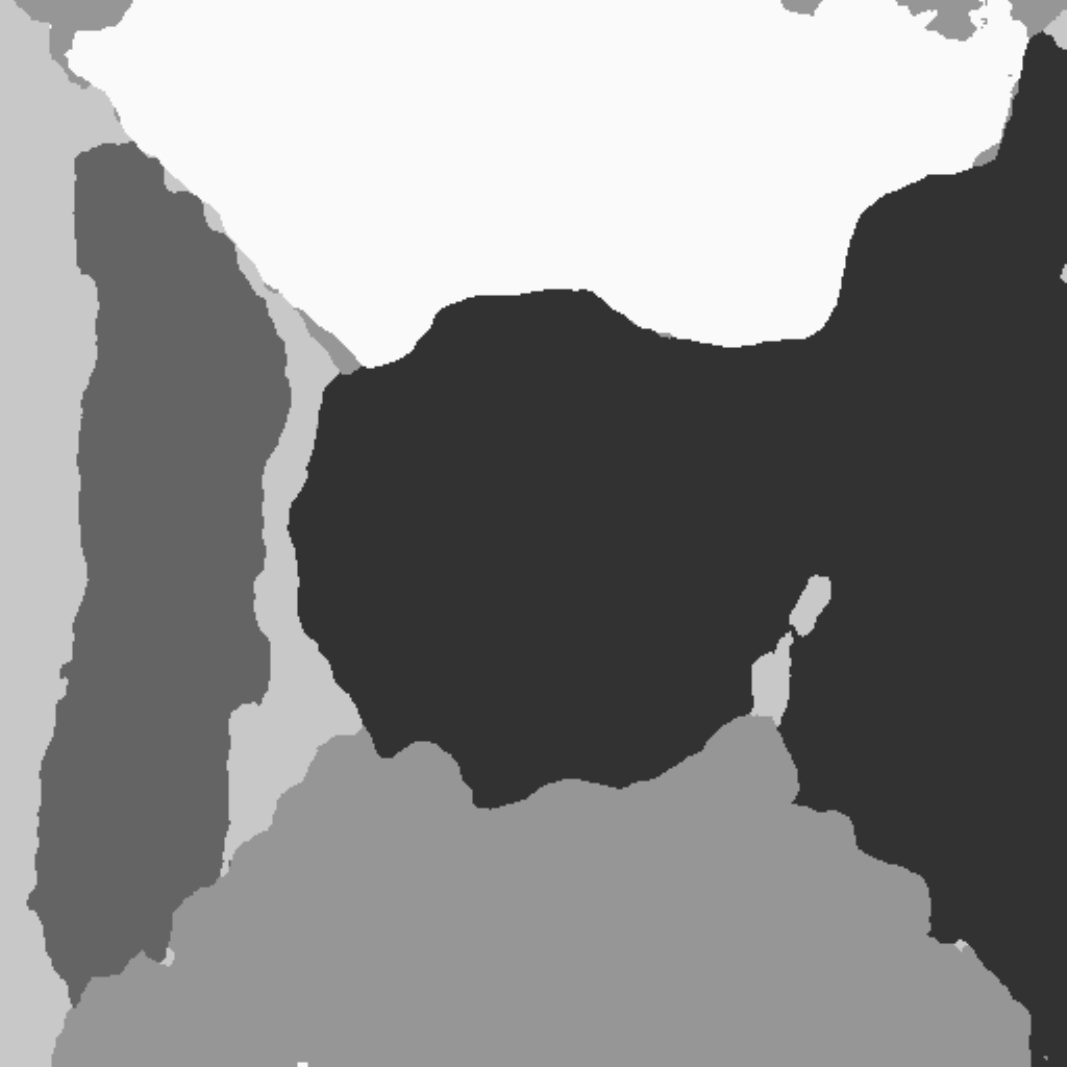}&
\includegraphics[width=0.17\textwidth]{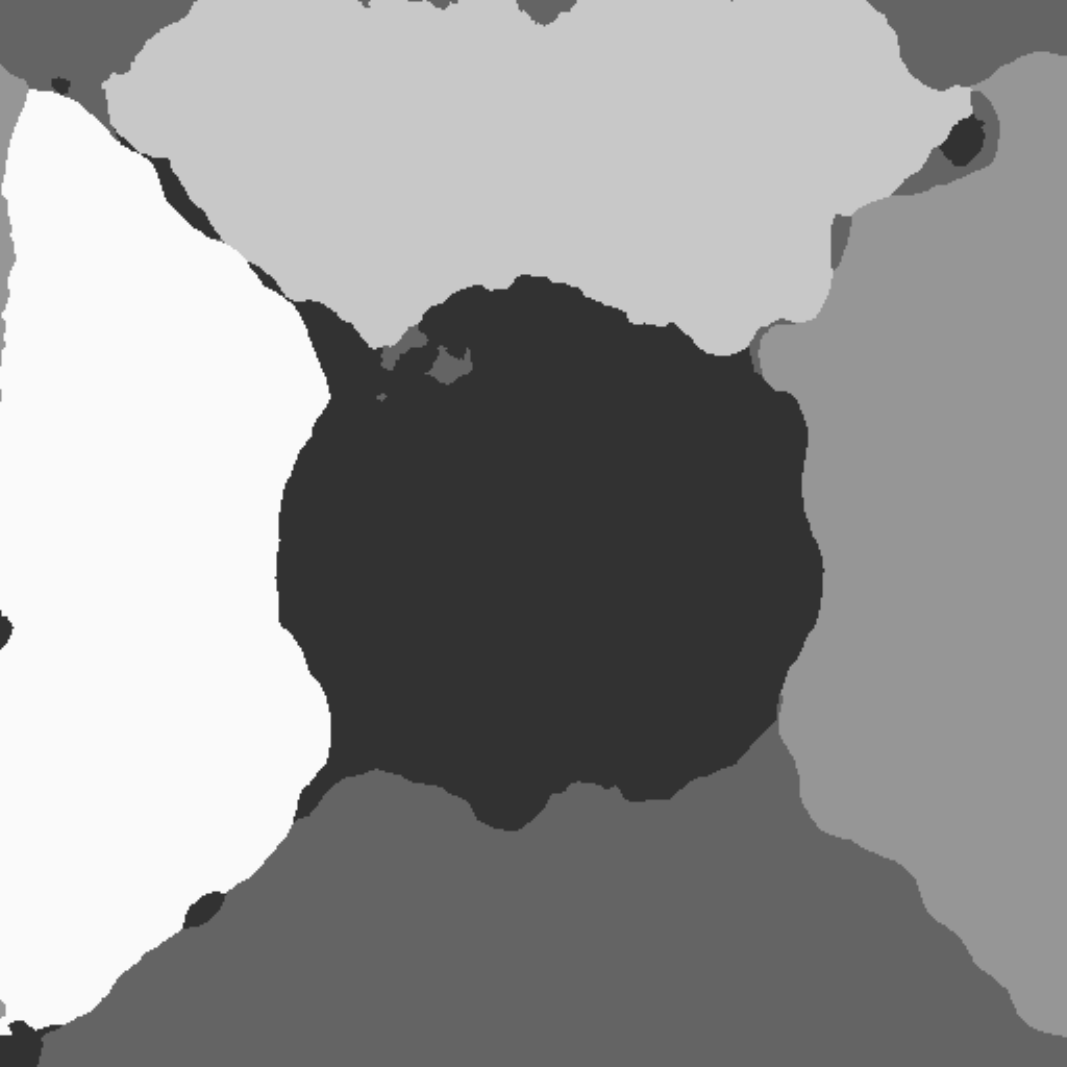} \\ 
\rotatebox{90}{EWTC1} & \includegraphics[width=0.17\textwidth]{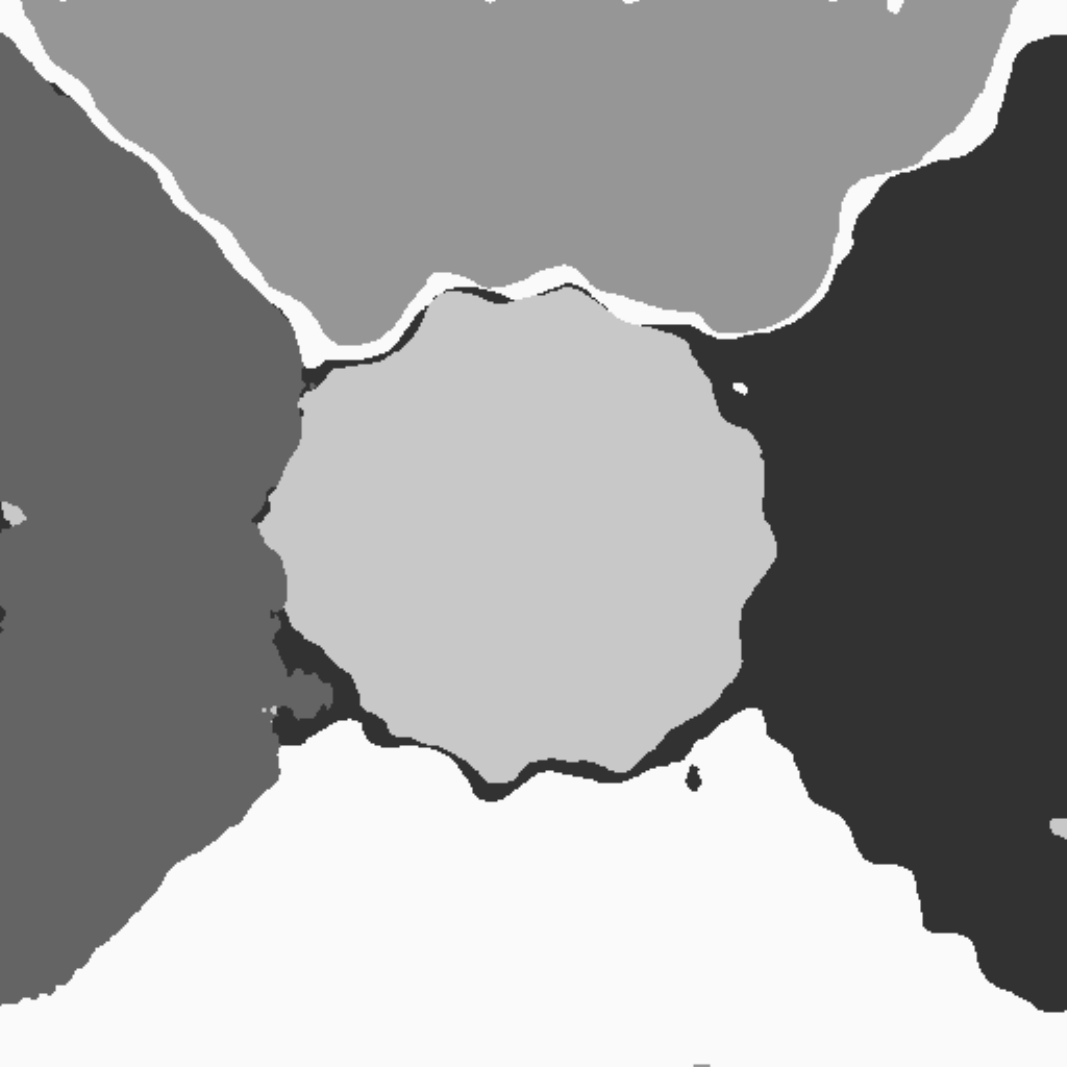} &
\includegraphics[width=0.17\textwidth]{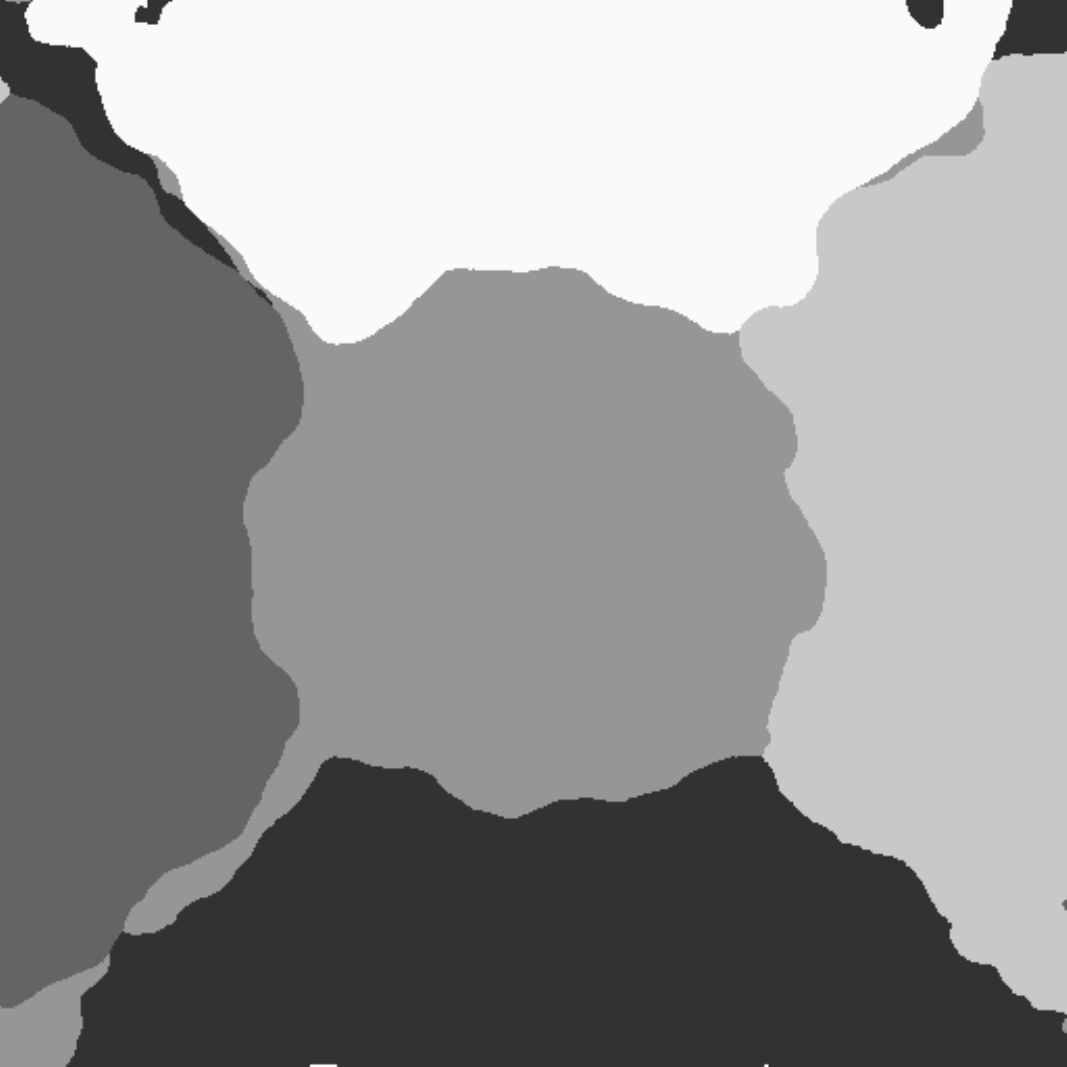} \\ 
\rotatebox{90}{EWTC2} & \includegraphics[width=0.17\textwidth]{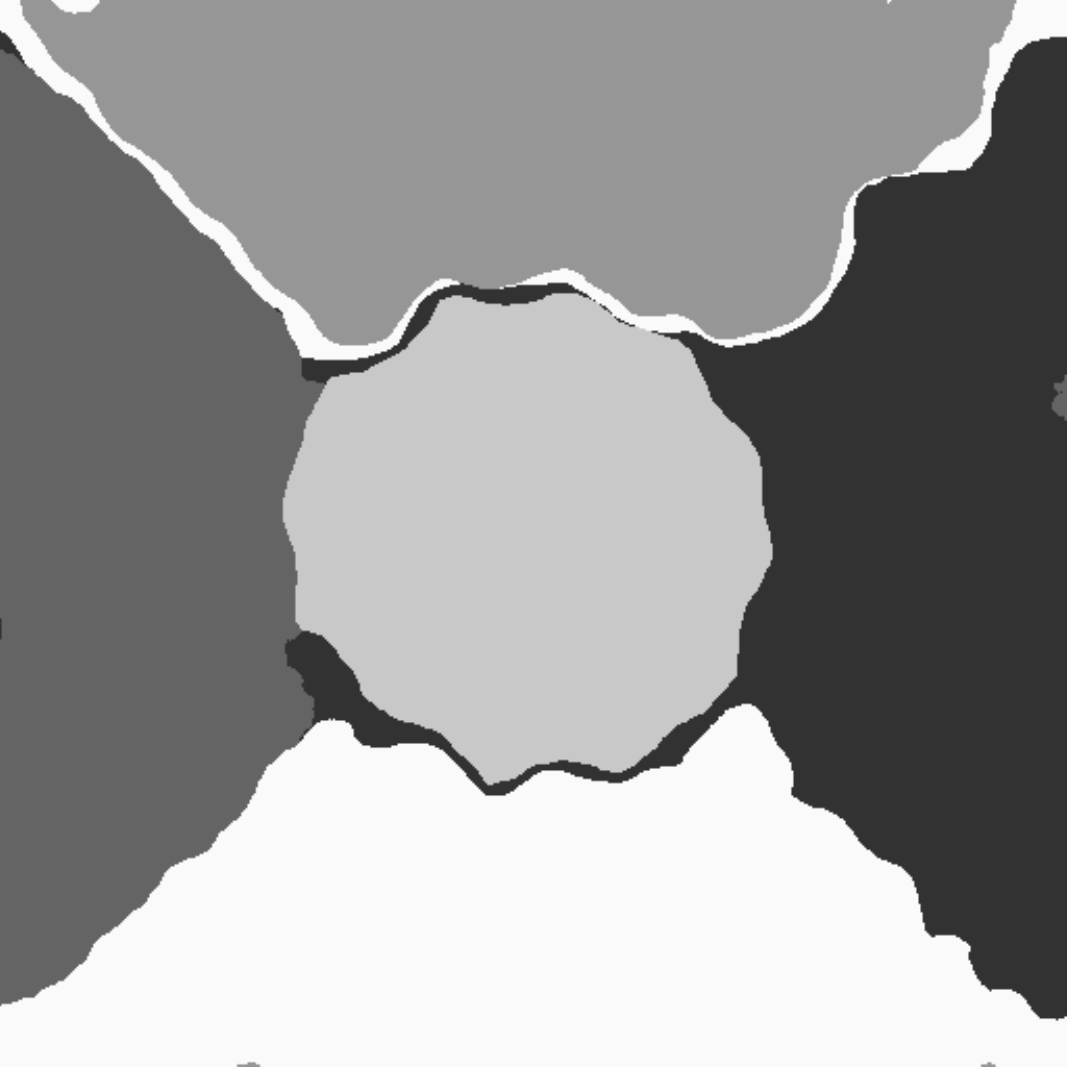} &
\includegraphics[width=0.17\textwidth]{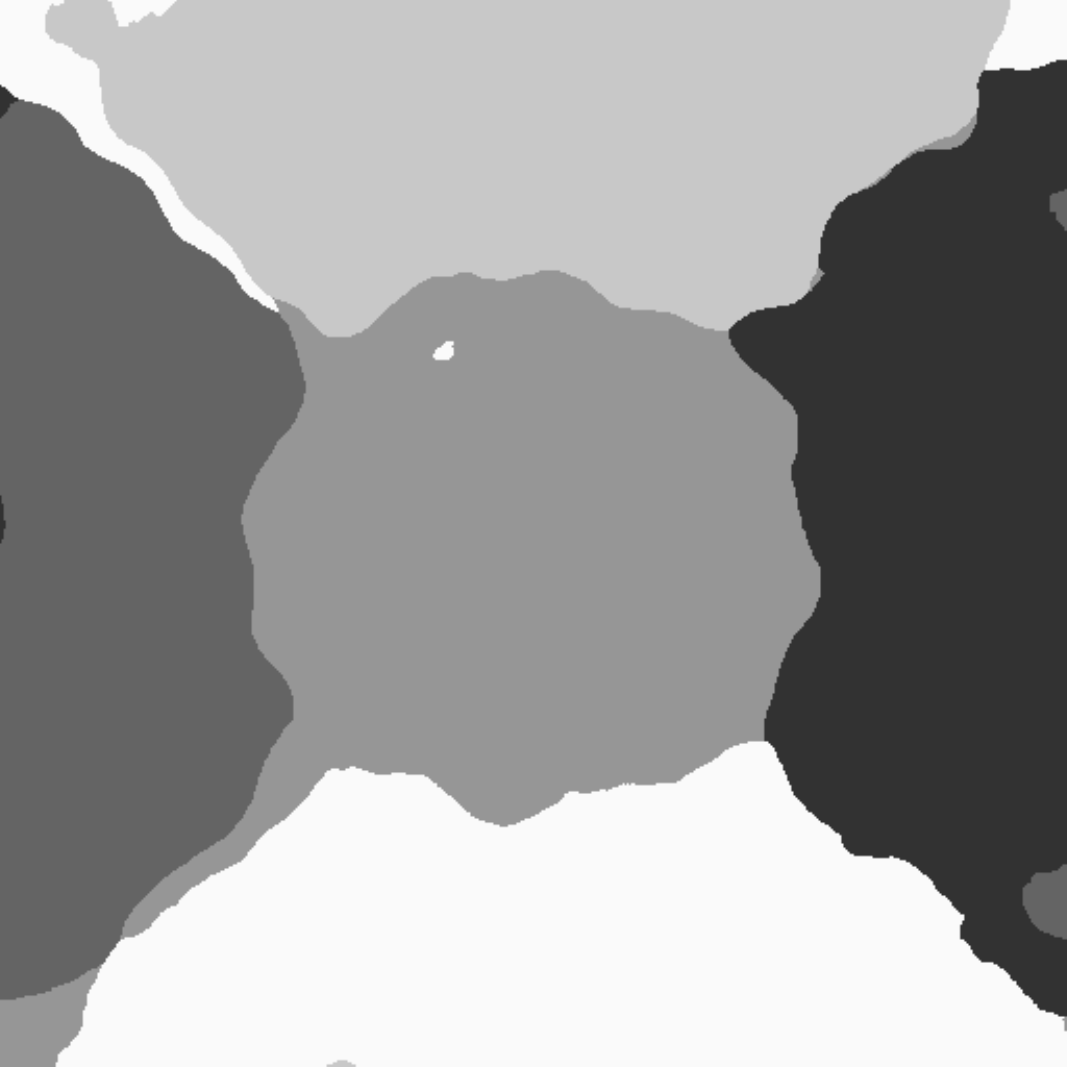} \\ 
\rotatebox{90}{EWTC3} & \includegraphics[width=0.17\textwidth]{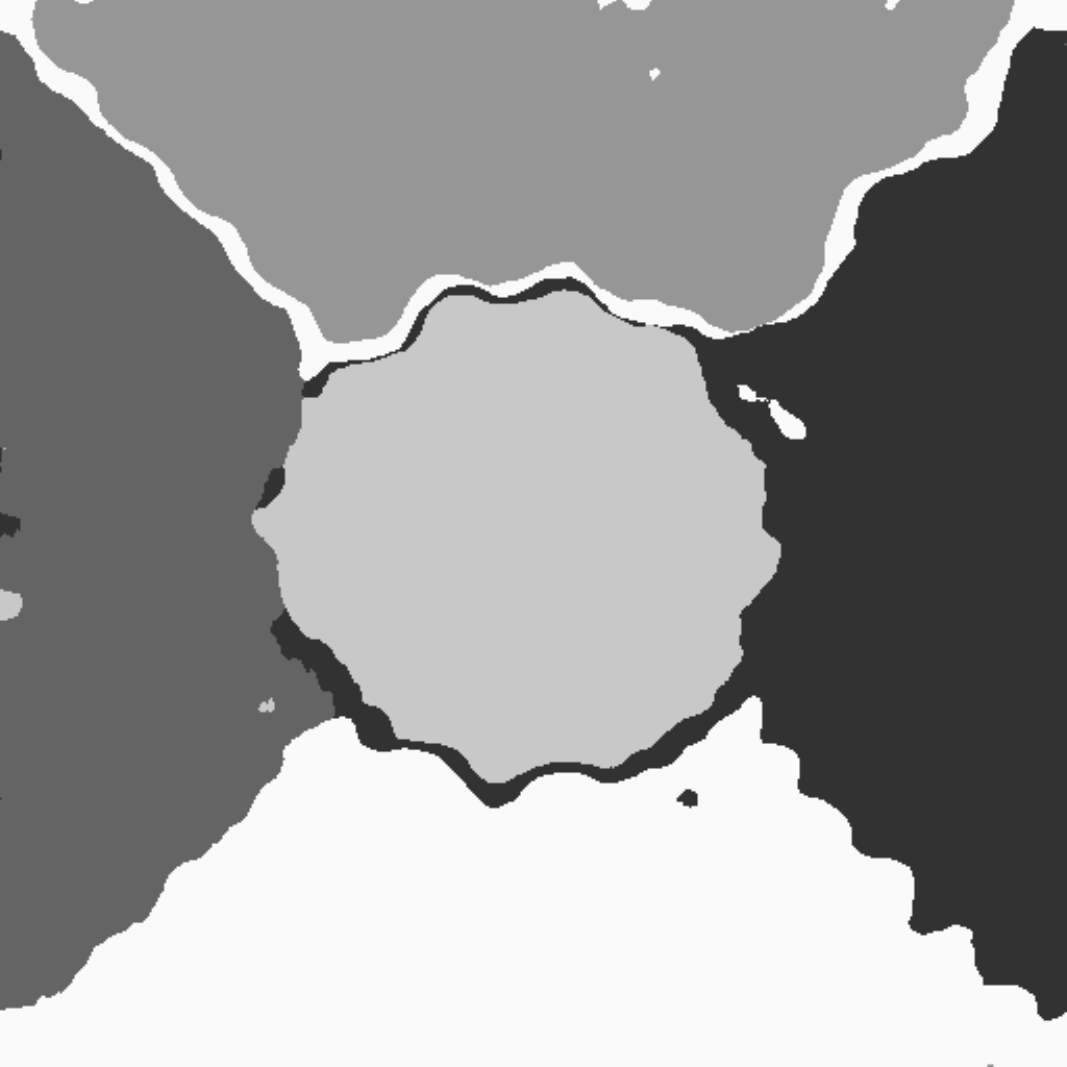} &
\includegraphics[width=0.17\textwidth]{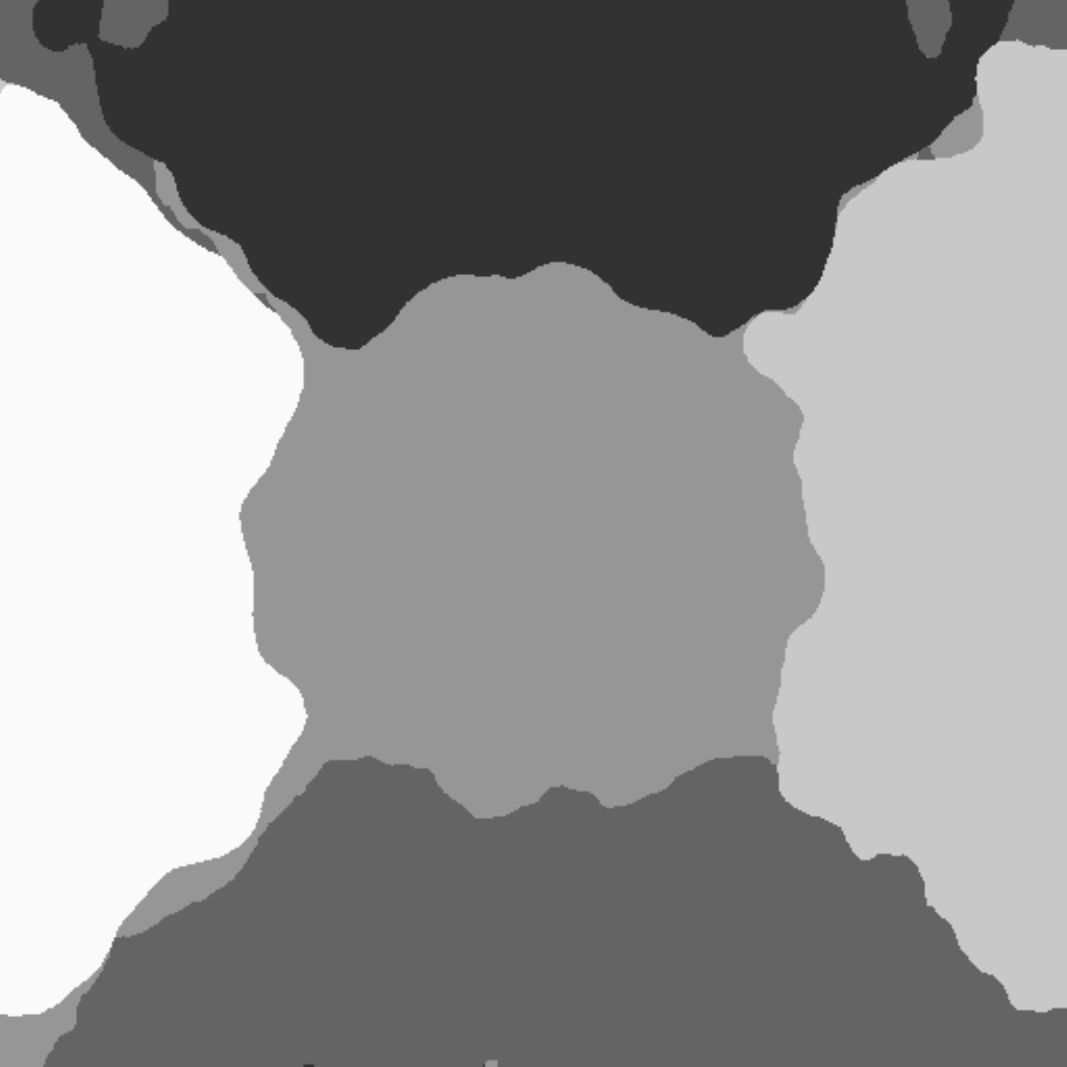} \\ 
\rotatebox{90}{EWT2DT} & \includegraphics[width=0.17\textwidth]{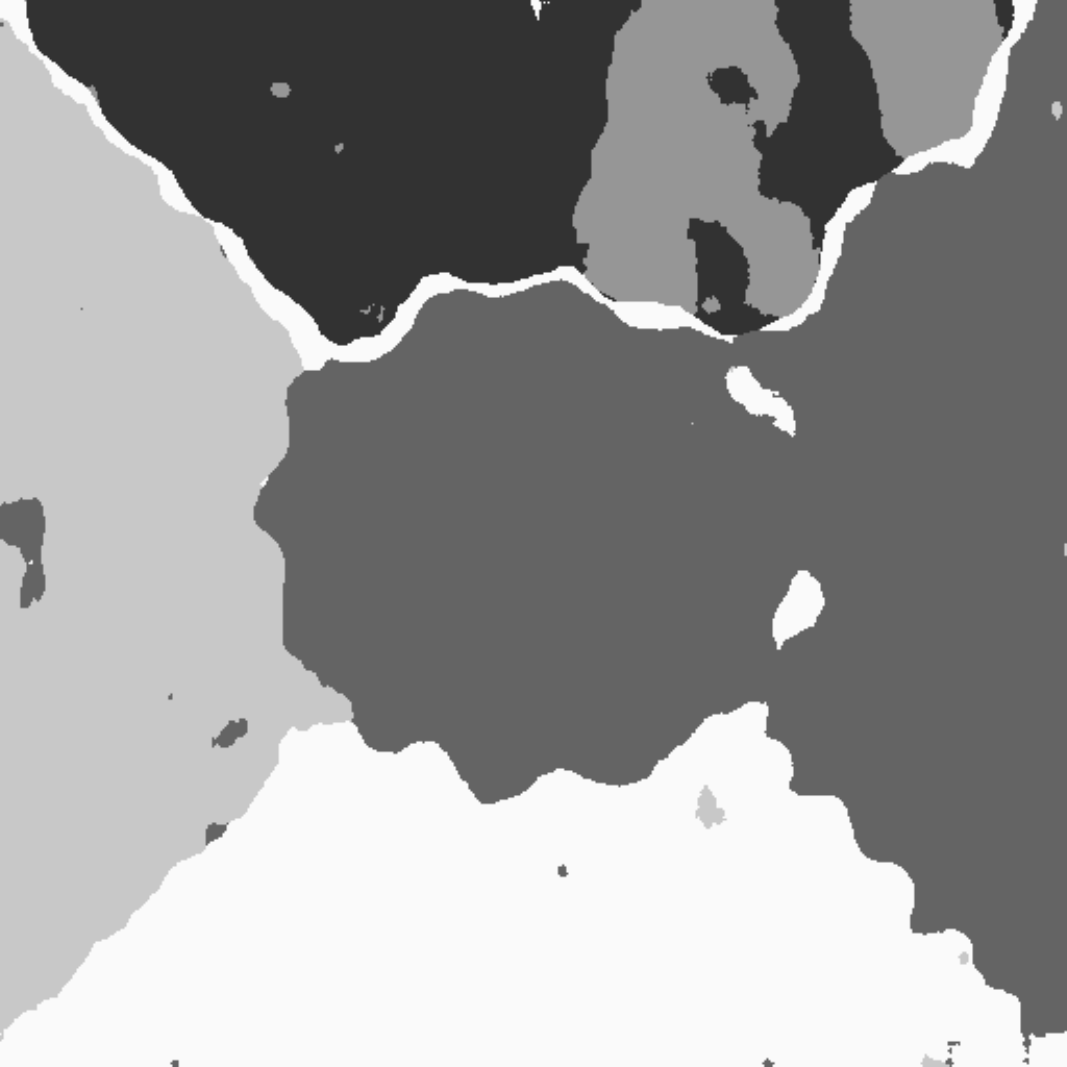} &
\includegraphics[width=0.17\textwidth]{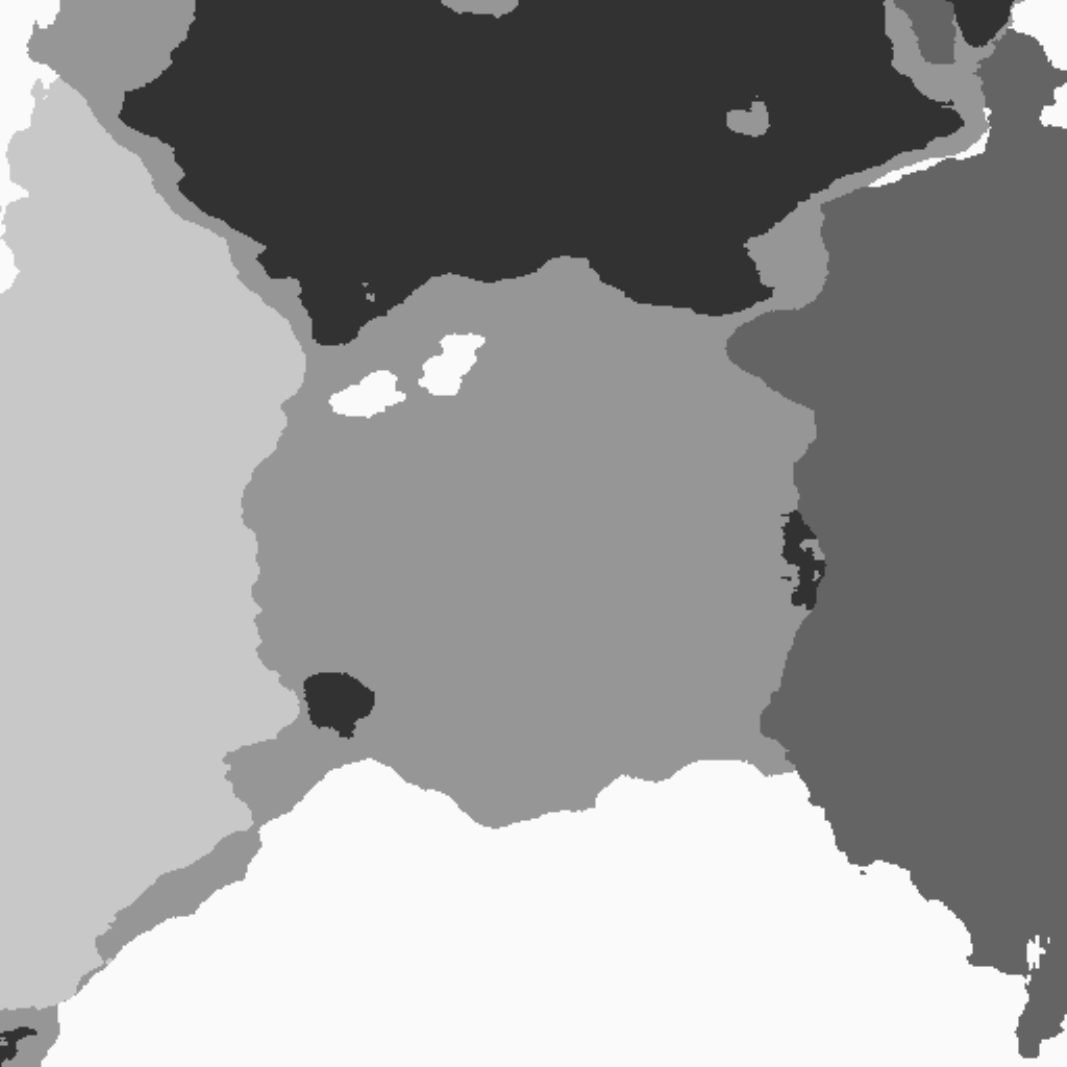} \\ 
\rotatebox{90}{Gabor} & \includegraphics[width=0.17\textwidth]{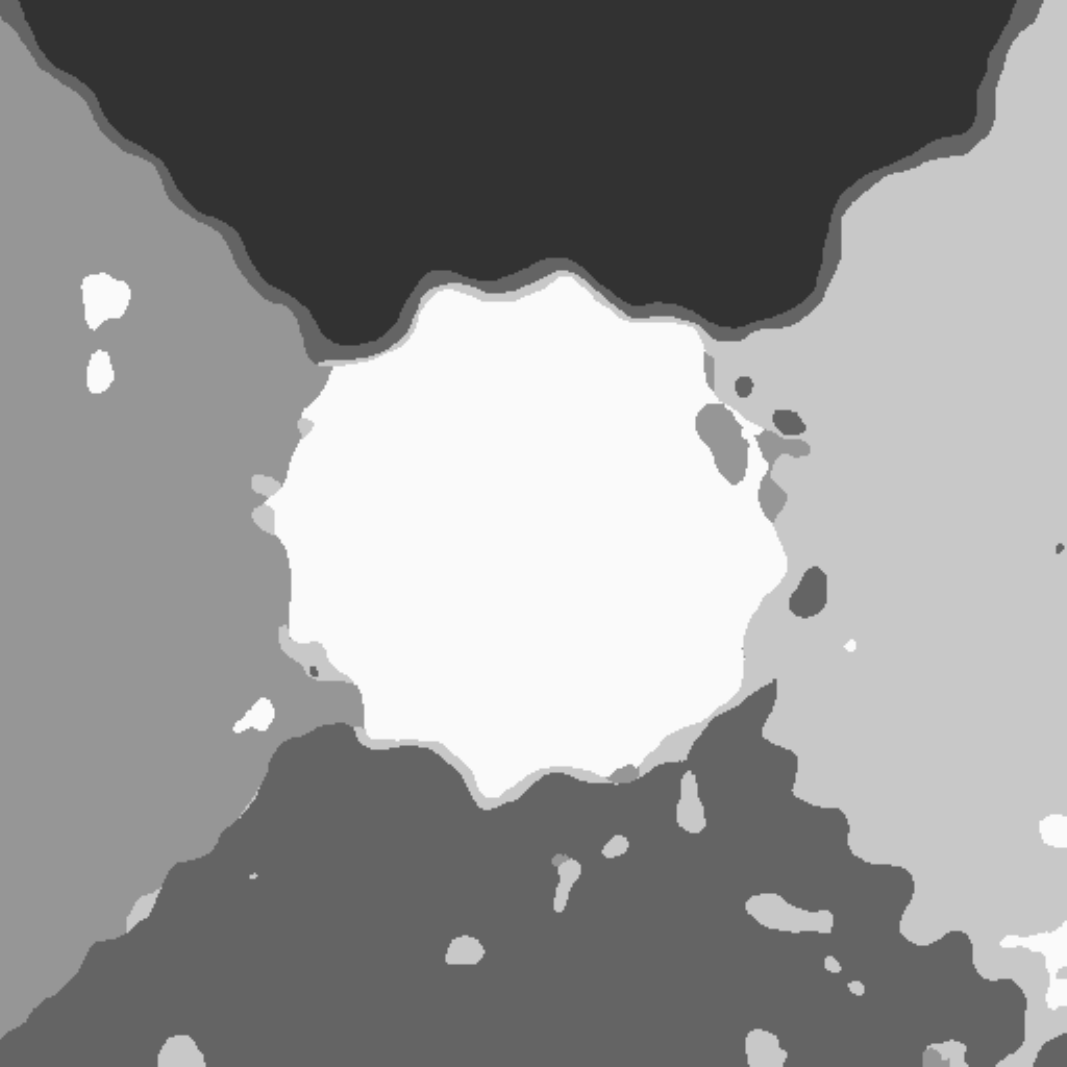} &
\includegraphics[width=0.17\textwidth]{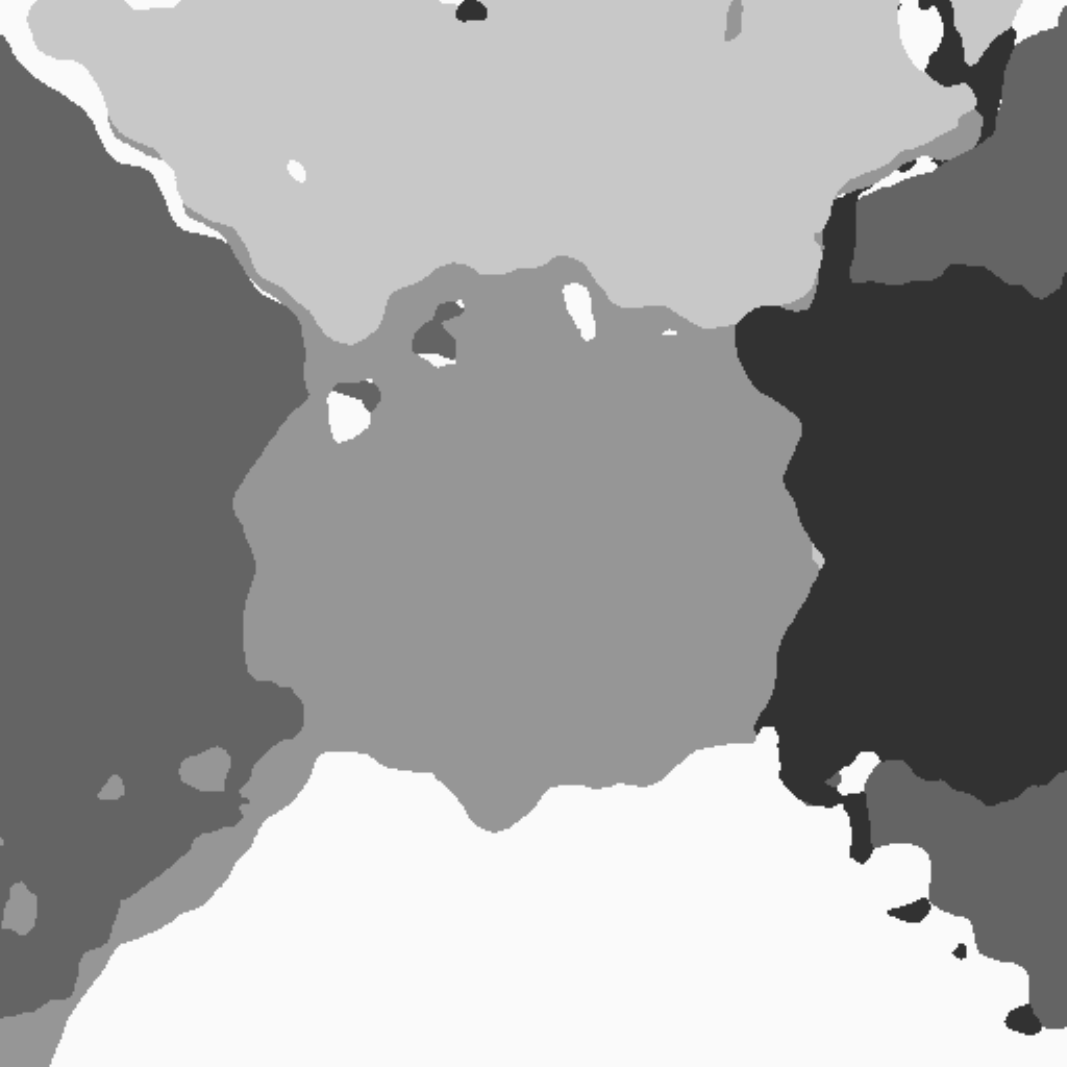} \\
\end{tabular}
\end{center}
\caption{Visual comparisons of segmentation results for the best wavelet families on one image from the ALOT and UIUC datasets.}
\label{fig:visucompare2}
\end{figure}

In Figures~\ref{fig:visucompare} and \ref{fig:visucompare2}, we illustrate some segmentation results on one image from each dataset for Gabor wavelets, classic curvelets and the 
best empirical wavelets. We can see that both the EWT2DC1 and EWT2DC2 give appealing segmentations in the sense that region boundaries are well localized. We can also observe that 
the original curvelets can have serious drawbacks in their result (see especially for texture from the ALOT dataset).

These experimental results show that for all datasets, the empirical wavelet family gives the best and most consistent segmentation results confirming the initial intuition that 
adaptive wavelet representations will better characterize textures.

%========================================================================================================================================
\section{Conclusion}\label{conclusion}
In this paper, we have assessed unsupervised wavelet-based texture segmentation. In order to focus on textures present in the image, we first decomposed the image 
into its cartoon and texture components and then processed only the texture part. In particular, we investigated the impact of considering an adaptive framework using 
the recently introduced empirical wavelets. We used four standard texture datasets to experiment the influence of the window's size used when computing the 
``wavelet energy'', the type of used wavelet energy, and the difference between a standard and a more advanced clustering technique. Finally, we ran the experiments based on a very 
broad choice of wavelets. As expected, wavelet families taking into account different directions are more efficient to characterize textures. Moreover, in almost all cases, the 
empirical wavelets (and in particular, the empirical curvelets) outperform the other types of wavelets. Their adaptability turns out to be a very important property 
to characterize textures.\\
In terms of future research, we did not take into account the information contained in the cartoon part of the original image in any of our experiments. Thus, it will be interesting 
to use a specific algorithm to segment the cartoon part on its own and finally find an ``optimal'' strategy to fusion both segmentations. Automatically detecting the 
expected number of classes in the segmentation is another open and difficult problem. If some approaches can be designed for particular type of images where some a priori 
information is available, the general case remains a challenge and further investigations must be conducted. We also started some investigation on how to combine the empirical 
transforms with deep learning techniques for the purpose of supervised texture segmentation.

\section*{Acknowledgment}
Yuan Huang stay at San Diego State University was supported by the Chinese Scholarship Council. The authors also want to thanks the anonymous referees for their helpful comments 
permitting to improve the quality of this paper.

%====================================================================

\end{document}